\documentclass[acmsmall]{acmart}

\AtBeginDocument{  \providecommand\BibTeX{{    \normalfont B\kern-0.5em{\scshape i\kern-0.25em b}\kern-0.8em\TeX}}}

\setcopyright{rightsretained}
\acmJournal{CSUR}
\acmYear{2021} \acmVolume{1} \acmNumber{1} \acmArticle{1} \acmMonth{1} \acmPrice{}\acmDOI{10.1145/3466819}

\usepackage{soul}
\usepackage{lscape}
\usepackage{longtable}
\usepackage[normalem]{ulem}
\useunder{\uline}{\ul}{}

\begin{document}

\title{A Survey on End-User Robot Programming}

\author{Gopika Ajaykumar}
\email{gopika@cs.jhu.edu}
\author{Maureen Steele}
\email{msteel21@jhu.edu}
\author{Chien-Ming Huang}
\email{cmhuang@cs.jhu.edu}
\orcid{0000-0002-6838-3701}
\affiliation{
  \institution{Department of Computer Science, Johns Hopkins University}
  \streetaddress{3400 N. Charles St.}
  \city{Baltimore}
  \state{Maryland}
  \postcode{21218}
}

\thanks{This work was supported in part by the National Science Foundation Graduate Research Fellowship Program under Grant No. DGE-1746891 and the Nursing/Engineering joint fellowship from the Johns Hopkins University.}

\renewcommand{\shortauthors}{Ajaykumar, Steele, and Huang}

\begin{abstract}
As robots interact with a broader range of end-users, end-user robot programming has helped democratize robot programming by empowering end-users who may not have experience in robot programming to customize robots to meet their individual contextual needs. This article surveys work on end-user robot programming, with a focus on end-user program specification. It describes the primary domains, programming phases, and design choices represented by the end-user robot programming literature. The survey concludes by highlighting open directions for further investigation to enhance and widen the reach of end-user robot programming systems.
\end{abstract}

\ccsdesc[500]{Computer systems organization~External interfaces for robotics}
\ccsdesc[300]{Human-centered computing~User studies}
\ccsdesc[300]{Software and its engineering~Development frameworks and environments}

\keywords{End-User Robot Programming, Human-Robot Interaction, End-User Programming}

\maketitle

\section{Introduction}
As computing technologies have been adopted in a variety of domains, the needs of their end-users in quickly customizing the operation of these technologies to meet domain-specific goals has resulted in increasing development of \emph{end-user programming} systems for users who are not professional software developers, such as children, accountants, and scientists. End-user programming technologies, such as spreadsheets, have reached such pervasiveness that it is estimated that end-user programmers substantially outnumber professional programmers today \cite{scaffidi2005estimating}. In this way, end-user programming has helped make computer programming widely accessible for those who do not practice software development as a career and has fostered the financial growth of individuals \cite{scaffidi2016potential} and the productivity of organizations \cite{wulf2004economics}. Analogous to how end-user programming has increased the accessibility of computer programming for end-users of computing technologies without extensive training in software development, \emph{end-user robot programming} is an emerging research area that seeks to enable end-users of general-purpose robotic technologies who are not robotics engineers to re-task and customize robots according to their needs.

End-users of robots are working closely with robots in manufacturing plants, warehouses, offices, shops, homes, and even on the road to a growing degree \cite{christensen2009roadmap}. As of 2020, there is a record of 2.7 million industrial robots being used in factories, with a growing share of these industrial robots being collaborative robots that work closely with end-users \cite{international2020world}. The variety of domains in which robots are adopted involve different user needs, environments, and task requirements. Rather than requiring robots to be pre-programmed by robotics engineers for specific application domains, which reduces flexibility, or for all potential task scenarios, end-user robot programming allows the end-users of robots to modify the operation of robots to work within their contexts. However, enabling end-user robot programming is a challenging problem. Robot programming is characterized by the embodied nature of the technology being programmed and the situated nature of its programs, introducing unique challenges such as the need for programs to reference and interact with the surrounding environment, which may potentially include task objects, obstacles, and other agents, including people. Effective robot programming not only requires mastery of the technical concepts involved in computer programming in general, such as data structures and algorithms, but also of a variety of multidisciplinary topics, from control engineering to mechatronics. Robotics software engineers must also have an in-depth knowledge of advanced topics related to robotics including path planning, localization, mapping, kinematics, control theory, computer vision, and machine learning. Furthermore, unlike the average robotics engineer, end-users vary widely in their backgrounds and technological literacy and may not have the time or capacity to learn how to program robotic technologies. The primary goal of end-user robot programming is to distill the complexity of robot programming into programming methods that are approachable for users without extensive experience in robotics. The challenge of enabling end-users to effectively develop robot programs that can safely be run in real-world environments without requiring expertise in software development practices and in the underlying hardware, mechanics, and control of robots has inspired a variety of methods for easily re-tasking robots from both industry and academia. 

Methods to enable end-users to re-task robots and author robot behaviors largely involve two different approaches---\emph{demonstration of skills} and \emph{specification of programs}. Robot \emph{Learning from Demonstration (LfD)}, known variously as \emph{Programming by Demonstration (PbD)} and \emph{imitation learning}, seeks to eliminate the complexities involved in having end-users manually specify robot programs of their desired robot skills by instead having users \emph{demonstrate} how to perform the desired robot skills. Under the LfD paradigm, robots develop new skills not by executing pre-specified programs, but by learning how to generalize multiple user-provided demonstrations of a task into a robot skill that can be applied in a variety of environmental and task variations. While LfD seeks to simplify end-user customization of robot behavior by eliminating the act of program specification completely, an alternative approach is to simplify robot programming enough such that end-users are capable of program specification. In this article, we survey existing work targeted towards this approach, which, unlike LfD, seeks to enable end-user programming of the structure, logic, and characteristics of the desired robot behavior rather than having end-users demonstrate several instantiations of it. Our survey of end-user robot programming may subsume works on end-user programming using demonstrations, but we focus on demonstrations as a tool for program specification rather than on the techniques to enable robot learning from end-user demonstrations, which have been covered by numerous surveys (e.g., \cite{argall2009survey, billard2008survey, schaal2003computational, breazeal2002robots, zhu2018robot, calinon2009robot, hussein2017imitation, chernova2014robot, liu2020skill}). Furthermore, we only survey papers on end-user program specification, as opposed to papers on direct control or instruction of robots (e.g., \cite{han2020structuring}).

We organize our paper around the domains, processes, and designs involved in end-user robot programming. In Section \ref{sec:background}, we give an overview of how end-user robot programming fits into the larger context of end-user programming and robot programming. Section \ref{sec:methodology} details the methodology we used to collect papers for the survey. In Section \ref{sec:domains}, we begin our survey, focusing specifically on the domains covered by end-user robot programming systems. Section \ref{sec:phases} surveys techniques used by end-user robot programming systems to enable end-users to participate in different phases of the robot programming process. In Section \ref{sec:design}, we review trends and goals in the design of end-user robot programming systems. Section \ref{sec:eval} provides an overview of user evaluation methods in end-user robot programming. Finally, we discuss open challenges and opportunities for future research in Section \ref{sec:discussion} and provide concluding remarks in Section \ref{sec:conclusion}.

\section{Background}\label{sec:background}
Although end-user robot programming may differ from other forms of programming in terms of its goals and constraints, it draws parallels and inspiration from various related fields. 
We describe two related fields, \textit{end-user programming} and \textit{robot programming}.

\subsection{End-User Programming}
End-user robot programming is just one direction within the overall area of end-user programming, which seeks to empower users to program their computing technologies themselves even without experience in programming. In addition to robotics, end-user programming has found use in a variety of application domains, including animation, e-mail, and gaming. End-user programming technologies have reached widespread use, with some estimates indicating that there were more than 55 million end-user programmers in the United States by 2012 \cite{scaffidi2005estimating}. End-user programming has also seen increasing research interest in the last decade, especially within the computer science and human-computer interaction communities, with several surveys being published on the topic over the past decade (e.g., \cite{barricelli2019end, paterno2013end, spahn2008end, maceli2017tools}). Common end-user programming approaches include spreadsheets, natural language programming, rule-based programming, visual programming, wizard-based programming, and programming by demonstration or example, many of which are used to enable end-user robot programming as well (Section \ref{programming-methods}). 

Using end-user programming tools, users may perform operations such as automating processes or manipulating data objects without extensive experience in using lower-level programming languages.  Unlike traditional programming, end-user programming is generally intended for personal, rather than public, use \cite{ko2011state}. Although end-users may program to meet different goals compared to professional developers, they face similar software engineering challenges, such as the need to test, debug, and secure their software \cite{ko2011state}, which has inspired focus on enabling not just end-user programming, but \emph{end-user software engineering}. A primary challenge in end-user software engineering has been to support the development of high-quality software in the face of extensive errors in software developed by end-users \cite{burnett2006next} and despite the tendency of end-users who are novice programmers to prioritize efficiency and ease over maintainability and robustness during programming \cite{brandt2008opportunistic}. End-user software engineering research in the past decade has focused on discovery of new programming methods and platforms, software engineering concepts, and application domains \cite{ko2011state}, a trend seen in end-user robot programming research as well.

Rooted in the increase in the number and diversity of computer users, end-user programming has helped transform programming by enabling shorter development cycles and empowering end-users to realize their real-time contextual needs through computing. Similarly, as the user base of robotic technologies grows in size and variety, end-user robot programming has emerged as a research area in its own right. Although it shares similarities and builds off of end-user programming approaches, end-user robot programming researchers must also contend with challenges and user needs specific to robot programming and human-robot interaction (HRI), motivating the need to consider end-user robot programming not only as a subcategory within end-user programming, but as an independent field involving unique user needs, goals, and constraints. 

\subsection{Robot Programming}
\label{robot-programming}
A central force driving end-user adoption of robotic technologies lies in their programmability, which enables them to be used for a variety of tasks without requiring substantial modification of their hardware or control. The availability of adequate robot programming tools to leverage robots' programmability is a key factor in determining the tasks a robot can be used for, as well as the total cost of applications involving robots \cite{lozano1983robot}. Thus, robot programming systems have played a central role in determining the adoption and evolution of robotics. Some of the earliest instances of robot programming took the form of robot-specific languages, which enabled programmers to specify robot motions and sensing operations for particular robots \cite{biggs2003survey}. These languages, which are commonly used to program industrial robots (e.g., Universal Robots' URScript and KUKA's KUKA Robot Language) and are still popular today, provide programmers with application programming interfaces and scripting languages for programming a specific robot or multiple robots from the same company. Although robot-specific languages tend to involve relatively simple syntax and programming commands, they still require programmers to have experience in programming languages and robotics. Furthermore, programmers need to learn a new robot-specific language every time they program a different robot due to the specificity of robot-specific languages. 
The limitation of robot-specific languages has resulted in a drive towards more abstracted robot programming systems that can be used to program different robots \cite{biggs2003survey}. Among these more abstracted programming methods, Robot Operating System (ROS) \cite{quigley2009ros} has emerged as the foremost framework for programming a variety of robotic technologies. As one of the most widely used tools for robot programming today, especially in robotics research labs, ROS has helped standardize the procedure for programming different types of robots and provides numerous libraries and simulation tools to simplify specification of robot operations such as navigation and path planning. While ROS provides some GUI-based tools to specify aspects of robot behavior, such as motion, without the need to program using textual program languages, it is primarily intended for use by expert robot programmers, which is seen through its largely command line-based interaction style. At a bare minimum, programmers need experience with a textual programming language, such as C++ or Python, and an understanding of concepts such as reference frames, kinematics, and motion planning to effectively program robots using ROS. Therefore, while ROS and similar frameworks have helped address the overspecificity of robot-specific languages through their applicability to many different robotic technologies and by driving robot programming towards standardization, they require a similarly high level of programming and robotics expertise from robot programmers, limiting their accessibility for end-users who may need to customize robots but are not robotics experts, such as workers on factory floors. The technical barriers preventing end-users of robots from using both robot-specific and more abstracted programming languages and frameworks have motivated the need for end-user robot programming systems.

The confluence of trends in end-user programming and robot programming has helped propel research on end-user robot programming. Several surveys have reviewed end-user robot programming research literature in the context of specific methods and application domains (e.g., \cite{coronado2020visual, villani2018survey}) or in the context of end-user programming (e.g., \cite{paterno2019end}) or robot programming (e.g., \cite{biggs2003survey}) in general. In this article, we aim to provide an updated and more comprehensive survey of research on enabling end-user specification of robot programs.

\section{Methodology} \label{sec:methodology}
We conducted a multi-phased collection process to obtain relevant papers for our survey. We focused our search on the databases of four publishers that publish works from major conferences and journals on human-computer interaction, human-robot interaction, and robotics: ACM Digital Library (\textit{ACM}), IEEE Xplore (\textit{IEEE}), ScienceDirect (\textit{Elsevier}), and SpringerLink (\textit{Springer}). 

\subsection{Phase 1: Initial Paper Collection}

During Phase 1 of our collection process, we used the search string \textit{end-user AND (robot OR robots OR robotic OR robotics) AND (program OR programming OR development OR instruction)} to obtain an initial set of papers from each of the four databases. To determine whether these initial papers were relevant to our survey, we read each paper's title and abstract to determine whether the paper referenced end-user robot programming and then applied the following exclusion criteria:

\begin{itemize}
\item{The paper is written in a language other than English.}
\item{The paper is about end-user robot programming for children or students, with a focus on outreach or learning outcomes rather than on program specification methods and techniques (see \cite{bravo2017review} for a review on end-user robot programming in the context of non-technical education).}
\item{The paper was about robot learning rather than end-user robot program specification. Several existing surveys review this topic (e.g., \cite{argall2009survey, billard2008survey}).} 
\item{The paper focused on intuitive or user-friendly robot programming for programming or robotics experts, rather than end-users without significant experience in programming and robotics (e.g., \cite{diprose2017designing, neto2013direct}).}
\end{itemize}

If the title and abstract referenced end-user robot programming and did not meet the exclusion criteria, the paper was collected. In cases where it was ambiguous from the title and abstract whether the paper was related to end-user robot programming, we skimmed through the paper to determine whether it should be collected, with a special focus on whether the paper discussed end-users, experts, novices, experience, and user evaluations. Throughout this process, we only collected peer-reviewed papers from symposiums, conferences, and journals.

Since three of the four databases yielded from 3,000 to around 33,000 results using our search string, we set an upper limit to the collection for these databases, which we chose empirically for each database based on the relevancy of the search results. Some papers were listed on both the ACM Digital Library and on IEEE Xplore. We collected these papers from ACM Digital Library but not from IEEE Xplore to avoid duplicates.

\begin{itemize}
\item{ACM Digital Library: Yielded 32,548 results; read through the first 400 (after which papers were less relevant); collected 45 papers}
\item{IEEE Xplore: Yielded 258 results; collected 18 papers}
\item{ScienceDirect: Yielded 33,497 results; read through the first 375 (after which papers were less relevant); collected 19 papers}
\item{SpringerLink: Yielded 3,140 conference papers and 2,654 journal papers; read through the first 200 conference papers and the first 50 journal papers (after which papers were less relevant); collected 16 papers}
\end{itemize}

\subsection{Phase 2: Additional Paper Collection}
After collecting papers from the four databases, we performed a search on Google Scholar using the same search string as before, which yielded 128,000 results. We looked through the first 100 items in the search results to discover any peer-reviewed papers that we failed to collect in Phase 1. Through this process, we added nine additional papers to our initial paper collection, resulting in a total of 107 papers. Six of the papers collected from Google Scholar were published by \emph{IEEE}, and three of the papers were published by independent publishers, such as the \emph{Association for the Advancement of Artificial Intelligence (AAAI)}. Our collection process, including Phases 1 and 2, was concluded on June 2020. While we limited our paper collection to papers published by June 2020, we did not include any bounds on how early the papers we collected could be published. 

\subsection{Phase 3: Additional Paper Elimination}
Finally, we applied the additional elimination criteria listed below to the 107 papers that we collected:

\begin{itemize}
\item{The paper describes designs for end-user robot programming systems but does not describe a system implementation realizing the proposed designs (e.g., \cite{fernaeus2009comics}).}
\item{The paper is a short paper consisting of four pages or less (e.g., \cite{oishi2017toward}).}
\item{The paper does not describe the results from any form of user evaluation of the end-user robot programming system, including case studies (e.g., \cite{gorostiza2010natural}). We included this elimination criterion because of the focus of our paper on user experiences in end-user program specification, rather than on isolated programming methods or system performance metrics.} 
\end{itemize}

The application of the additional criteria resulted in the elimination of 62 papers (two of which described no system implementation, 28 of which were short papers, and 32 of which did not include results from a user evaluation). This left a total of 45 pages for our survey, the majority of which were from the United States ($n=18$). 

\section{Domains of End-User Robot Programming} \label{sec:domains}
End-user robot programming systems are designed for the common goal of making robot programming more accessible for a wider range of people than only robotics experts. However, they may differ in terms of the categories of users and robots that they are meant for. Below, we describe the diversity of end-users and robots covered by the end-user robot programming research literature.

\subsection{Types of End-Users}
End-user robot programming systems cover a spectrum of end-user expertise levels and backgrounds (Table \ref{tab:table1}). While the majority of the systems described in the surveyed papers are geared for general users from any background or expertise level, some systems are targeted towards specific groups of users. These include domain specialists such as healthcare workers (e.g., \cite{kubota2020jessie}) 
and industrial workers (e.g., \cite{ong2020augmented}).
These domain specialists may have experience in Science, Technology, Engineeering, and Mathematics (STEM) \cite{barakova2013end, paxton2018evaluating} but tend to have little to no background in robotics or computer programming and no prior experience with robot programming. On the other hand, some systems are meant for everyday groups of users who may not be specialized in a particular field, including children (e.g., \cite{barivsic2018leveraging, ryokai2009children, sapounidis2013tangible}), elderly adults (e.g., \cite{datta2011end}), caregivers (e.g., \cite{barakova2013end, datta2011end}), store workers (e.g., \cite{liang2018simultaneous}), and non-programmer adults (e.g., \cite{ramouglu2017programming, weintrop2018evaluating}).

Due to the diversity of expertise levels that users have, most end-user robot programming systems are designed to support programming for users who have no experience at all in STEM, programming, and robotics. However, some systems may include more advanced programming concepts, such as parallelization (e.g., \cite{barakova2013end, gorostiza2011end, guerin2015framework, kapinus2019spatially, leonardi2019trigger}), procedural abstraction (e.g., \cite{alexandrova2015roboflow}), and functions (e.g., \cite{huang2020vipo}). Furthermore, systems may support a variety of expertise levels, such as by including natural language-based high-level programming features for novice programmers and more expressive, block-based programming features for more advanced users (e.g., \cite{beschi2019capirci}). Such systems can support novice robot programmers in gradually learning to leverage the full expressivity of an end-user robot programming system and allow both inexperienced and experienced users to make use of the system for their programming needs. 

\subsection{Types of Robots}
As robots vary greatly in terms of form factors, capabilities, and use cases, end-user robot programming systems often focus on enabling intuitive customization of a specific type or make of robot (Table \ref{tab:table1}). The types of robots used in end-user robot programming systems today can be broadly categorized into two categories: robots primarily intended for manipulation tasks and robots primarily intended for non-manipulation tasks. We describe the robots that are commonly programmed by end-users within these two classes below.

\subsubsection{Robots Primarily Intended for Manipulation Tasks}
The most common type of robot used for end-user robot programming in the surveyed literature is the industrial robot manipulator (arm), which is traditionally used for manufacturing tasks.  
These robots, which have six to seven movable joints, can be programmed to perform various manipulation tasks, ranging from simple pick-and-place tasks to high precision assembly, depending on the individual robot's capabilities and range of motion. 
Another common type of robot used in end-user robot programming systems exposes even more capabilities for programming through the use of two manipulators, paving the way for complex manipulation tasks that may require the simultaneous use of multiple forces, such as unscrewing the lid of a jar. Within the surveyed literature, three collaborative robots of this type are used in end-user robot programming systems: the PR2 robot, the Baxter robot, and the YuMi robot. Systems using the PR2 robot from Willow Garage may expose its navigation, manipulation, and sensing and perception capabilites for end-user programming, whereas systems meant for the Baxter robot from Rethink Robotics may expose its customizable facial appearance, its sensing and manipulation capabilities, and, depending on whether it is mounted on a stationary or mobile pedestal, its navigation capabilities for programming. Both types of robots have two 7-degrees-of-freedom arms and a 2-degrees-of-freedom head. On the other hand, the YuMi robot from ABB Robotics includes two 7-degrees-of-freedom arms but no manueverable head or base, which limits it to end-user robot programming of manipulation tasks. While the PR2 robot is a service robot meant for everyday human environments and daily living tasks that require dexterity, the Baxter robot is an industrial robot intended for simple industrial tasks, with a focus on pick-and-place operations, and the YuMi robot is an industrial robot designed for manufacturing environments and assembly tasks. 
\subsubsection{Robots Primarily Intended for Non-Manipulation Tasks}
In addition to robots intended for manipulation tasks, social robots, mobile service robots, home robots, and educational robots are also types of robots that are commonly programmed by end-users. Social robots include anthropomorphic and zoomorphic robots. The anthropomorphic robots used in end-user robot programming systems include the NAO (e.g., \cite{barakova2013end, buchina2016design, buchina2019natural, manohar2014programming, erich2017visual}) and the Pepper (e.g., \cite{leonardi2019trigger}) robots from SoftBank Robotics, the Kaspar robot from the University of Hertfordshire (e.g., \cite{moros2019programming}), the Maggie robot from the University Carlos III of Madrid (e.g., \cite{gorostiza2011end}), and the Kuri robot from Mayfield Robotics (e.g., \cite{kubota2020jessie}). The Pepper, Maggie, and Kuri robots are mobile, allowing them to be programmed for navigation tasks. The zoomorphic robots used in end-user robot programming systems include stuffed animal robots (e.g., \cite{young2014design}) and the Pleo robot from Innvo Labs (e.g., \cite{manohar2014programming, ryokai2009children}). End-user robot programming systems that use social robots allow users to program both low-level actions, such as sensing and motion, as well as high-level interactive behaviors, such as emotion expression. 

Other robots used in end-user robot programming systems include mobile service robots (e.g., \cite{datta2011end, huang2016design}), such as the Savioke Relay robot. End-user robot programming systems that work with these types of robots allow users to program various types of commands, from navigation to sound and visual display. In addition to robots found in service sectors, some end-user robot programming systems focus on programming of robots found in everyday environments. These include home robots such as the Roomba robot (e.g., \cite{young2014design}), robotic toys such as  robotic spheres from Sphero (e.g., \cite{ramouglu2017programming}), Arduino cars (e.g., \cite{barivsic2018leveraging}), and LEGO robots (e.g., \cite{sapounidis2013tangible}). Programming systems that work with lower-cost consumer robots often center on the programming of robot movement and navigation.
While end-user robot programming systems tend to focus on the programming of one specific robot, some systems provide users with the option to program multiple robots simultaneously (e.g., \cite{barakova2013end, riedl2019fast}), including robots of different makes (e.g., \cite{riedl2019fast}). Recent systems further extend end-user programming by enabling the simultaneous programming of robots together with the Internet of Things that is present in the robot's environment (e.g., \cite{huang2020vipo, leonardi2019trigger}).

\section{Phases of End-User Robot Programming} \label{sec:phases}
The process of developing a program using an end-user robot programming system involves various phases, such as setting up the system, program, or task environment; specifying program logic; fixing and checking the authored program for correctness; and executing the program. We describe the various ways in which programming systems enable users to participate in each of these phases.

\subsection{Initialization/Setup}
Similar to how computer programming may require initialization of variables or setup of a program editor, end-user robot programming may require initial setup before program authoring can begin. While a setup or initialization phase is not supported or required in all end-user robot programming systems, some systems may offer system-level setup features, such as hardware testing or calibration (e.g., \cite{cakmak2014teaching}) or specification of user information (e.g., \cite{gorostiza2011end}), and program-level setup features, such as specifying task objects (e.g., \cite{huang2017code3}) or the workspace (e.g., \cite{schou2018skill}) that must be represented in the programming system. Since robot programming systems, unlike most computer programming systems, often work with programs involving interaction with other objects and entities in the environment, the end-user must also set up the task environment to include necessary components, such as task objects. Setup may also involve preparing necessary hardware for the programming system itself, such as microphones (e.g., \cite{alexandrova2014robot, cakmak2014teaching, ryokai2009children}), wearable devices (e.g., \cite{bambuvssek2019combining, quintero2018robot}), or projectors (e.g., \cite{schou2018skill, sefidgar2018robotist, gao2019pati}). To minimize the overhead required on the part of the user, some end-user robot programming systems automate parts of the setup process, such as mapping  of the programming workspace and detection of programmable devices and robots (e.g., \cite{huang2020vipo}). On the other hand, supporting users in appropriately initializing programs, which is challenging for novice programmers \cite{franklin2016initialization}, has been less explored in the literature, though prior work has suggested its potential usefulness in preventing program errors by end-users (e.g., \cite{weintrop2017blockly}). 

\subsection{Authoring}
The literature on end-user robot programming systems includes a diversity of focal points, methods, and approaches towards enabling end-user authoring of robot programs. We highlight the variety of authoring techniques that end-users can leverage for robot programming.

\subsubsection{Programmable Robot Capabilities}
\label{programmable-robot-capabilities}  
End-user robot programming systems limit programmable robot capabilities to specific actions that can interlock with their respective robots' capabilities and meet the needs of their intended users and domains (Table \ref{tab:table1}). Programs relating to manipulation are commonly supported by end-user robot programming systems. 
Some end-user robot programming systems focus on lower-level programming of manipulation and motion paths through which users may specify robot end-effector poses (e.g., \cite{cakmak2014teaching, stenmark2017simplified, gadre2019end}) or continuous trajectories (e.g., \cite{jha2015application}). Others focus on higher-level programming of manipulation through specification of the objects and locations to be used for pick-and-place actions (e.g., \cite{gao2019pati, sefidgar2017situated, ong2020augmented}), or they may  provide additional user flexibility by allowing both low-level and high-level programming of manipulation (e.g., \cite{quintero2018robot}).

\begin{landscape}
\begin{longtable}[c]{cccccccccccccc}
\caption{Domains, Programming Methods, and Programmable Capabilities in End-User Robot Programming.
\\ \emph{\textbf{Programming Method} (Section \ref{programming-methods}): V = Visual, XR = Augmented/Mixed Reality, D = Demonstration, NL = Natural Language, T = Tangible\\ \textbf{Programmable Capabilities} (Section \ref{programmable-robot-capabilities}): Interactive Behavior (e.g., emotion, dialog, interaction style), Manipulation (e.g., any motion involving manipulation of objects via gripper), Motion (e.g., motion paths, gestures), Navigation (e.g., motion of mobile robots to specified locations in environment), Sensing (e.g., active perception, touch sensor readings), Audiovisual (e.g., audio output such as robot speech and visual output such as turning on an LED)}}
\label{tab:table1}\\
\cline{4-14}
\multicolumn{3}{c}{\textbf{}} &
  \multicolumn{5}{c|}{\textbf{\begin{tabular}[c]{@{}c@{}}Programming \\ Method\end{tabular}}} &
  \multicolumn{6}{c}{\textbf{\begin{tabular}[c]{@{}c@{}}Programmable \\ Capabilities\end{tabular}}} \\ \hline
\multicolumn{1}{c|}{\textbf{Basics}} &
  \multicolumn{1}{c|}{\textbf{End-User}} &
  \multicolumn{1}{c|}{\textbf{Robot}} &
  \multicolumn{1}{c|}{\textbf{V}} &
  \multicolumn{1}{c|}{\textbf{XR}} &
  \multicolumn{1}{c|}{\textbf{D}} &
  \multicolumn{1}{c|}{\textbf{NL}} &
  \multicolumn{1}{c|}{\textbf{T}} &
  \multicolumn{1}{c|}{\textbf{Interac.}} &
  \multicolumn{1}{c|}{\textbf{Manip.}} &
  \multicolumn{1}{c|}{\textbf{Motion}} &
  \multicolumn{1}{c|}{\textbf{Nav.}} &
  \multicolumn{1}{c|}{\textbf{Sensing}} &
  \textbf{Aud./Vis.} \\ \hline
\endfirsthead
\multicolumn{3}{c}        {Table \thetable\ Continued from Previous Page} \\
\cline{4-14}
\multicolumn{3}{c}{\textbf{}} &
  \multicolumn{5}{c|}{\textbf{\begin{tabular}[c]{@{}c@{}}Programming \\ Method\end{tabular}}} &
  \multicolumn{6}{c}{\textbf{\begin{tabular}[c]{@{}c@{}}Programmable \\ Capabilities\end{tabular}}} \\ \hline
\multicolumn{1}{c|}{\textbf{Basics}} &
  \multicolumn{1}{c|}{\textbf{End-User}} &
  \multicolumn{1}{c|}{\textbf{Robot}} &
  \multicolumn{1}{c|}{\textbf{V}} &
  \multicolumn{1}{c|}{\textbf{XR}} &
  \multicolumn{1}{c|}{\textbf{D}} &
  \multicolumn{1}{c|}{\textbf{NL}} &
  \multicolumn{1}{c|}{\textbf{T}} &
  \multicolumn{1}{c|}{\textbf{Interac.}} &
  \multicolumn{1}{c|}{\textbf{Manip.}} &
  \multicolumn{1}{c|}{\textbf{Motion}} &
  \multicolumn{1}{c|}{\textbf{Nav.}} &
  \multicolumn{1}{c|}{\textbf{Sensing}} &
  \textbf{Aud./Vis.} \\ \hline
\endhead
\begin{tabular}[c]{@{}c@{}}Alexandrova et al. \\ (2014) \cite{alexandrova2014robot}\end{tabular} &
  General &
  PR2 &
  \checkmark &
   &
  \checkmark &
  \checkmark &
   &
   &
  \checkmark &
  \checkmark &
   &
   &
   \\ \hline
\begin{tabular}[c]{@{}c@{}}Alexandrova et al. \\ (2015) \cite{alexandrova2015roboflow}\end{tabular} &
  General &
  PR2 &
  \checkmark &
   &
  \checkmark &
   &
   &
   &
  \checkmark &
  \checkmark &
  \checkmark &
  \checkmark &
   \\ \hline
\begin{tabular}[c]{@{}c@{}}Bambu\v{s}ek et al. \\ (2019) \cite{bambuvssek2019combining}\end{tabular} &
  General &
  PR2 &
   &
  \checkmark &
   &
   &
   &
   &
  \checkmark &
  \checkmark &
   &
   &
   \\ \hline
\begin{tabular}[c]{@{}c@{}}Barakova et al. \\ (2013) \cite{barakova2013end}\end{tabular} &
  \begin{tabular}[c]{@{}c@{}}Healthcare\\ workers\end{tabular} &
  Social &
  \checkmark &
   &
   &
  \checkmark &
   &
   &
   &
  \checkmark &
   &
  \checkmark &
  \checkmark \\ \hline
\begin{tabular}[c]{@{}c@{}}Bari\v{s}i\'c et al.\\ (2018) \cite{barivsic2018leveraging}\end{tabular} &
  Children &
  \begin{tabular}[c]{@{}c@{}}Arduino\\ car\end{tabular} &
  \checkmark &
   &
   &
   &
   &
   &
   &
  \checkmark &
   &
  \checkmark &
   \\ \hline
\begin{tabular}[c]{@{}c@{}}Beschi et al.\\ (2019) \cite{beschi2019capirci}\end{tabular} &
  \begin{tabular}[c]{@{}c@{}}Industrial \\ workers\end{tabular} &
  Arm &
  \checkmark &
   &
   &
  \checkmark &
   &
   &
  \checkmark &
  \checkmark &
   &
  \checkmark &
   \\ \hline
\begin{tabular}[c]{@{}c@{}}Buchina et al.\\ (2016) \cite{buchina2016design}\end{tabular} &
  \begin{tabular}[c]{@{}c@{}}Healthcare\\ workers\end{tabular} &
  Social &
  \checkmark &
   &
   &
  \checkmark &
   &
   &
   &
  \checkmark &
   &
  \checkmark &
  \checkmark \\ \hline
\begin{tabular}[c]{@{}c@{}}Buchina et al.\\ (2019) \cite{buchina2019natural}\end{tabular} &
  \begin{tabular}[c]{@{}c@{}}Healthcare\\ workers\end{tabular} &
  Social &
  \checkmark &
   &
   &
  \checkmark &
   &
   &
   &
  \checkmark &
   &
  \checkmark &
  \checkmark \\ \hline
\begin{tabular}[c]{@{}c@{}}Cakmak and \\ Takayama \\ (2014) \cite{cakmak2014teaching}\end{tabular} &
  General &
  PR2 &
   &
   &
  \checkmark &
  \checkmark &
   &
   &
  \checkmark &
  \checkmark &
   &
   &
   \\ \hline
\begin{tabular}[c]{@{}c@{}}Datta et al.\\ (2011) \cite{datta2011end}\end{tabular} &
  \begin{tabular}[c]{@{}c@{}}Healthcare\\ workers,\\ Elderly\\ adults,\\ Caregivers\end{tabular} &
  \begin{tabular}[c]{@{}c@{}}Mobile\\ service\end{tabular} &
  \checkmark &
   &
   &
   &
   &
  \checkmark &
   &
   &
   &
   &
  \checkmark \\ \hline
\begin{tabular}[c]{@{}c@{}}Erich et al.\\ (2017) \cite{erich2017visual}\end{tabular} &
  \begin{tabular}[c]{@{}c@{}}Healthcare\\ workers\end{tabular} &
  Social &
  \checkmark &
   &
   &
   &
   &
   &
   &
   &
   &
  \checkmark &
  \checkmark \\ \hline
\begin{tabular}[c]{@{}c@{}}Forbes et al.\\ (2014) \cite{forbes2014robot}\end{tabular} &
  General &
  PR2 &
  \checkmark &
   &
  \checkmark &
   &
   &
   &
  \checkmark &
  \checkmark &
   &
   &
   \\ \hline
\begin{tabular}[c]{@{}c@{}}Gadre et al.\\ (2019) \cite{gadre2019end}\end{tabular} &
  General &
  Baxter &
   &
  \checkmark &
   &
   &
   &
   &
  \checkmark &
  \checkmark &
   &
   &
   \\ \hline
\begin{tabular}[c]{@{}c@{}}Gao and Huang\\ (2019) \cite{gao2019pati}\end{tabular} &
  General &
  Arm &
   &
  \checkmark &
   &
   &
   &
   &
  \checkmark &
   &
   &
   &
   \\ \hline
\begin{tabular}[c]{@{}c@{}}Gorostiza and\\ Salichs (2011) \cite{gorostiza2011end}\end{tabular} &
  General &
  Social &
   &
   &
   &
  \checkmark &
   &
  \checkmark &
   &
  \checkmark &
  \checkmark &
  \checkmark &
  \checkmark \\ \hline
\begin{tabular}[c]{@{}c@{}}Guerin et al.\\ (2015) \cite{guerin2015framework}\end{tabular} &
  \begin{tabular}[c]{@{}c@{}}Industrial\\ workers\end{tabular} &
  Arm &
  \checkmark &
   &
   &
   &
   &
   &
  \checkmark &
  \checkmark &
   &
  \checkmark &
   \\ \hline
\begin{tabular}[c]{@{}c@{}}Huang et al.\\ (2016) \cite{huang2016design}\end{tabular} &
  General &
  \begin{tabular}[c]{@{}c@{}}Mobile\\ service\end{tabular} &
  \checkmark &
   &
   &
   &
   &
  \checkmark &
   &
  \checkmark &
  \checkmark &
  \checkmark &
  \checkmark \\ \hline
\begin{tabular}[c]{@{}c@{}}Huang and \\ Cakmak (2017) \cite{huang2017code3}\end{tabular} &
  General &
  PR2 &
  \checkmark &
   &
  \checkmark &
  \checkmark &
   &
  \checkmark &
  \checkmark &
  \checkmark &
   &
  \checkmark &
  \checkmark \\ \hline
\begin{tabular}[c]{@{}c@{}}Huang et al.\\ (2020) \cite{huang2020vipo}\end{tabular} &
  \begin{tabular}[c]{@{}c@{}}Industrial\\ workers\end{tabular} &
  Mobile &
  \checkmark &
   &
   &
   &
   &
   &
  \checkmark &
  \checkmark &
  \checkmark &
   &
   \\ \hline
\begin{tabular}[c]{@{}c@{}}Jha et al.\\ (2015) \cite{jha2015application}\end{tabular} &
  General &
  Arm &
   &
   &
  \checkmark &
   &
   &
   &
   &
  \checkmark &
   &
   &
   \\ \hline
\begin{tabular}[c]{@{}c@{}}Kapinus et al.\\ (2019) \cite{kapinus2019spatially}\end{tabular} &
  \begin{tabular}[c]{@{}c@{}}Industrial\\ workers\end{tabular} &
  PR2 &
   &
  \checkmark &
   &
   &
   &
   &
  \checkmark &
   &
   &
   &
   \\ \hline
\begin{tabular}[c]{@{}c@{}}Kubota et al.\\ (2020) \cite{kubota2020jessie}\end{tabular} &
  \begin{tabular}[c]{@{}c@{}}Healthcare\\ workers\end{tabular} &
  Social &
   &
   &
   &
   &
  \checkmark &
  \checkmark &
   &
   &
   &
  \checkmark &
  \checkmark \\ \hline
\begin{tabular}[c]{@{}c@{}}Leonardi et al.\\ (2019) \cite{leonardi2019trigger}\end{tabular} &
  General &
  Social &
  \checkmark &
   &
   &
   &
   &
   &
   &
  \checkmark &
   &
  \checkmark &
  \checkmark \\ \hline
\begin{tabular}[c]{@{}c@{}}Liang et al.\\ (2017) \cite{liang2017framework}\end{tabular} &
  General &
  Baxter &
   &
   &
  \checkmark &
   &
   &
   &
  \checkmark &
   &
   &
   &
   \\ \hline
\begin{tabular}[c]{@{}c@{}}Liang et al.\\ (2018) \cite{liang2018simultaneous}\end{tabular} &
  \begin{tabular}[c]{@{}c@{}}Store\\ workers\end{tabular} &
  \begin{tabular}[c]{@{}c@{}}Mobile\\ arm\end{tabular} &
  \checkmark &
   &
  \checkmark &
   &
   &
   &
  \checkmark &
   &
   &
   &
   \\ \hline
\begin{tabular}[c]{@{}c@{}}Liang et al.\\ (2019) \cite{liang2019end}\end{tabular} &
  General &
  Baxter &
  \checkmark &
   &
  \checkmark &
   &
   &
   &
  \checkmark &
   &
   &
   &
   \\ \hline
\begin{tabular}[c]{@{}c@{}}Manohar and\\ Crandall (2014) \cite{manohar2014programming}\end{tabular} &
  General &
  Social &
   &
   &
  \checkmark &
   &
   &
  \checkmark &
   &
   &
   &
   &
   \\ \hline
\begin{tabular}[c]{@{}c@{}}Matthaiakis et al.\\ (2017) \cite{matthaiakis2017flexible}\end{tabular} &
  \begin{tabular}[c]{@{}c@{}}Industrial\\ workers\end{tabular} &
  Arm &
  \checkmark &
   &
   &
   &
   &
   &
  \checkmark &
   &
   &
   &
   \\ \hline
\begin{tabular}[c]{@{}c@{}}Moros et al.\\ (2019) \cite{moros2019programming}\end{tabular} &
  General &
  Social &
  \checkmark &
   &
   &
   &
   &
  \checkmark &
   &
  \checkmark &
   &
  \checkmark &
  \checkmark \\ \hline
\begin{tabular}[c]{@{}c@{}}Ong et al.\\ (2020) \cite{ong2020augmented}\end{tabular} &
  \begin{tabular}[c]{@{}c@{}}Industrial\\ workers\end{tabular} &
  Arm &
   &
  \checkmark &
   &
   &
   &
   &
  \checkmark &
  \checkmark &
   &
   &
   \\ \hline
\begin{tabular}[c]{@{}c@{}}Paxton et al.\\ (2018) \cite{paxton2018evaluating}\end{tabular} &
  \begin{tabular}[c]{@{}c@{}}Industrial\\ workers\end{tabular} &
  Arm &
  \checkmark &
   &
   &
   &
   &
   &
  \checkmark &
  \checkmark &
   &
  \checkmark &
   \\ \hline
\begin{tabular}[c]{@{}c@{}}Pedersen and\\ Kr\"{u}ger (2015) \cite{pedersen2015gesture}\end{tabular} &
  \begin{tabular}[c]{@{}c@{}}Industrial\\ workers\end{tabular} &
  \begin{tabular}[c]{@{}c@{}}Mobile\\ arm\end{tabular} &
  \checkmark &
   &
  \checkmark &
   &
   &
   &
  \checkmark &
   &
   &
   &
   \\ \hline
\begin{tabular}[c]{@{}c@{}}Quintero et al.\\ (2018) \cite{quintero2018robot}\end{tabular} &
  \begin{tabular}[c]{@{}c@{}}Industrial\\ workers\end{tabular} &
  Arm &
   &
  \checkmark &
   &
  \checkmark &
   &
   &
  \checkmark &
  \checkmark &
   &
   &
   \\ \hline
\begin{tabular}[c]{@{}c@{}}Racca et al.\\ (2020) \cite{racca2020interactive}\end{tabular} &
  General &
  Arm &
  \checkmark &
   &
   &
   &
   &
   &
  \checkmark &
  \checkmark &
   &
  \checkmark &
   \\ \hline
\begin{tabular}[c]{@{}c@{}}Ramo\u{g}lu et al.\\ (2017) \cite{ramouglu2017programming}\end{tabular} &
  Adults &
  Toy &
  \checkmark &
   &
   &
   &
   &
   &
   &
  \checkmark &
   &
  \checkmark &
  \checkmark \\ \hline
\begin{tabular}[c]{@{}c@{}}Riedl and\\ Henrich (2019) \cite{riedl2019fast}\end{tabular} &
  \begin{tabular}[c]{@{}c@{}}Industrial\\ workers\end{tabular} &
  Arm &
  \checkmark &
   &
  \checkmark &
   &
   &
   &
  \checkmark &
  \checkmark &
   &
   &
   \\ \hline
\begin{tabular}[c]{@{}c@{}}Ryokai et al.\\ (2009) \cite{ryokai2009children}\end{tabular} &
  Children &
  Social &
  \checkmark &
   &
   &
   &
  \checkmark &
  \checkmark &
   &
  \checkmark &
   &
  \checkmark &
  \checkmark \\ \hline
\begin{tabular}[c]{@{}c@{}}Sapounidis and\\ Demetriadis\\ (2013) \cite{sapounidis2013tangible}\end{tabular} &
  Children &
  LEGO &
  \checkmark &
   &
   &
   &
  \checkmark &
   &
   &
  \checkmark &
   &
  \checkmark &
  \checkmark \\ \hline
\begin{tabular}[c]{@{}c@{}}Schou et al.\\ (2018) \cite{schou2018skill}\end{tabular} &
  \begin{tabular}[c]{@{}c@{}}Industrial\\ workers\end{tabular} &
  Arm &
  \checkmark &
   &
  \checkmark &
   &
   &
   &
  \checkmark &
  \checkmark &
  \checkmark &
  \checkmark &
   \\ \hline
\begin{tabular}[c]{@{}c@{}}Sefidgar et al.\\ (2017) \cite{sefidgar2017situated}\end{tabular} &
  General &
  PR2 &
   &
   &
   &
   &
  \checkmark &
   &
  \checkmark &
   &
   &
   &
   \\ \hline
\begin{tabular}[c]{@{}c@{}}Sefidgar et al.\\ (2018) \cite{sefidgar2018robotist}\end{tabular} &
  General &
  PR2 &
   &
   &
   &
   &
  \checkmark &
   &
  \checkmark &
   &
   &
   &
   \\ \hline
\begin{tabular}[c]{@{}c@{}}Stenmark et al.\\ (2017) \cite{stenmark2017simplified}\end{tabular} &
  General &
  YuMi &
  \checkmark &
   &
  \checkmark &
   &
   &
   &
  \checkmark &
  \checkmark &
   &
  \checkmark &
   \\ \hline
\begin{tabular}[c]{@{}c@{}}Weintrop et al.\\ (2017) \cite{weintrop2017blockly}\end{tabular} &
  \begin{tabular}[c]{@{}c@{}}Industrial\\ workers\end{tabular} &
  Arm &
  \checkmark &
   &
   &
   &
   &
   &
  \checkmark &
  \checkmark &
   &
   &
   \\ \hline
\begin{tabular}[c]{@{}c@{}}Weintrop et al.\\ (2018) \cite{weintrop2018evaluating}\end{tabular} &
  Adults &
  Arm &
  \checkmark &
   &
   &
   &
   &
   &
  \checkmark &
  \checkmark &
   &
   &
   \\ \hline
\begin{tabular}[c]{@{}c@{}}Young et al.\\ (2014) \cite{young2014design}\end{tabular} &
  General &
  \begin{tabular}[c]{@{}c@{}}Roomba,\\ Toy\end{tabular} &
   &
   &
  \checkmark &
   &
   &
  \checkmark &
   &
  \checkmark &
   &
   &
   \\ \hline
\end{longtable}
\end{landscape}

For mobile robots, end-user robot programming systems allow users to author programs related to robot navigation (e.g., \cite{alexandrova2015roboflow, huang2016design, huang2020vipo}). When authoring navigation-related programs, end-users may program the robot to navigate to a known location, which may either be preprogrammed by the system developer, automatically detected by the system (e.g., \cite{huang2016design, huang2020vipo}), or specified by the end-user (e.g., \cite{pedersen2015gesture}). Beyond task-related movements, end-users may also specify motion for robot gestures (e.g., \cite{barakova2013end, buchina2019natural}).

In addition to motion-related programs, end-user robot programming systems may allow user authoring of robot sensing using sensors such as cameras, motion sensors, and touch sensors to enable the robot to react appropriately to its surroundings (e.g., \cite{barakova2013end, buchina2019natural, guerin2015framework, ramouglu2017programming, kubota2020jessie}). Including sensing capabilities can extend the flexibility of end-user robot programming systems by allowing programming in more dynamic environments where task objects or obstacles change locations frequently. Sensor-related actions are often used for event-driven programming or conditional constructs in programming (e.g., \cite{kubota2020jessie}). For robots with first-person cameras, end-user robot programming systems may allow user authoring of programs related to active perception (e.g., \cite{alexandrova2014robot}). Active perception enables the robot to maneuver itself or its sensors to obtain information about its environment, which can further minimize the responsibility of the user in specifying object locations during robot programming. In addition to supporting general robot actions related to sensing and audiovisual capabilities such as speech, sound, and lighting (e.g., \cite{barakova2013end, buchina2019natural}), end-user robot programming systems may also support programming of higher-level actions, such as human-robot interaction and collaboration (e.g., \cite{alexandrova2015roboflow, guerin2015framework, huang2016design, ryokai2009children, young2014design}), or task-level commands, such as common therapeutic exercise commands (e.g., \cite{kubota2020jessie}) or object assembly commands (e.g., \cite{ schou2018skill}). 
\subsubsection{Authoring Scope}
End-user robot programming systems vary in their authoring scope. Many systems focus on enabling users to specify the structure and logic of a program using pre-specified robot primitives, such as grasps or spinning, that are often developed by expert robot programmers (e.g., \cite{barivsic2018leveraging, huang2016design, huang2020vipo, kubota2020jessie, leonardi2019trigger, ryokai2009children, sapounidis2013tangible, stenmark2017simplified, weintrop2018evaluating, barakova2013end, gorostiza2011end, weintrop2017blockly, alexandrova2015roboflow, buchina2016design, buchina2019natural, guerin2015framework, kapinus2019spatially, beschi2019capirci, moros2019programming, pedersen2015gesture, ramouglu2017programming}). Some end-user robot programs give the user more granular access by allowing end-users to specify the primitives used for subsequent programming (e.g., \cite{huang2016design, liang2017framework, schou2018skill, erich2017visual, liang2019end, stenmark2017simplified}) or by taking an end-to-end approach in which end-users are fully responsible for specifying every aspect of the program, as in imitation learning (e.g., \cite{jha2015application, manohar2014programming}). On the other hand, other end-user robot programming systems handle the bulk of the program specification using techniques such as automated planning (e.g., \cite{liang2017framework, liang2019end}). In such systems, end-users may only need to specify high-level aspects of the system, such as end goals or style, or make minor refinements to the system-generated program (e.g., \cite{sefidgar2018robotist, bambuvssek2019combining, liang2018simultaneous}). Similarly, some systems may focus on enabling rapid prototyping of robot programs rather than longer development cycles (e.g., \cite{huang2016design, huang2017code3}). In addition to enabling specification of the overall logic and flow of a program, some systems may enable end-users to specify the parameters for commands or functions (e.g., \cite{huang2020vipo, paxton2018evaluating, weintrop2017blockly}). In order to ease the process of specifying continuous parameters or parameters in 3-D space, systems may provide intuitive methods for parametrization, such as directional (e.g., \cite{racca2020interactive}) or gesture-based specification (e.g., \cite{pedersen2015gesture}).

\subsubsection{Programming Methods}
\label{programming-methods}
\begin{figure} 
  \includegraphics[width=\linewidth]{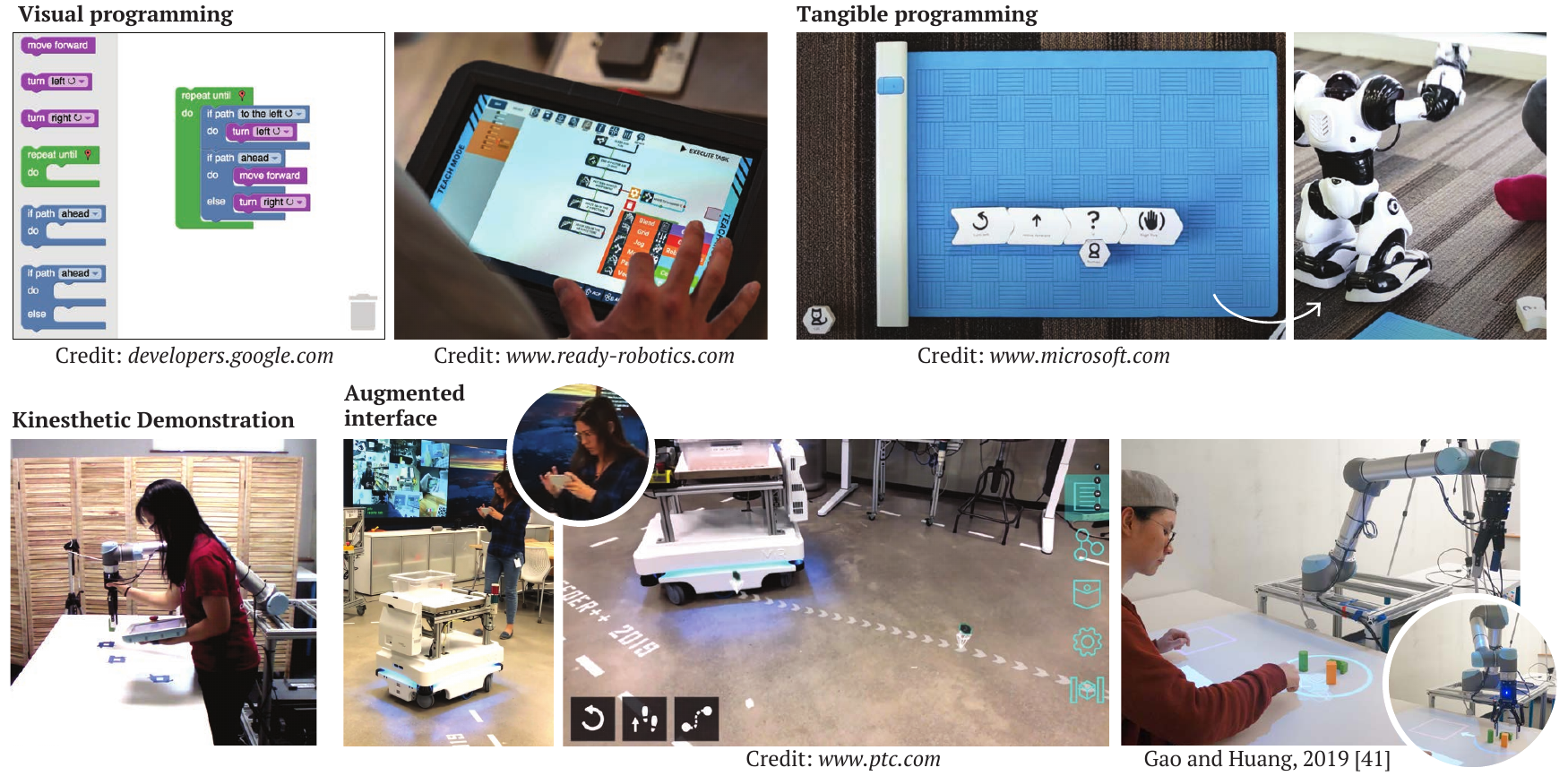}
  \caption{Examples of End-User Robot Programming Methods. \\\emph{(Top)} Visual, Tangible
  \\ \emph{(Bottom)} Demonstration, Augmented Reality
    }
  \label{fig:methods}
\end{figure}

End-user robot programming systems offer a variety of methods for users to program robot actions and behaviors (Table \ref{tab:table1}), each of which has its own advantages and limitations. We describe the most common methods used to date (see Fig. \ref{fig:methods} for examples). 

\emph{Visual.} Visual programming is a common end-user programming method that is widely recognized for its success in making programming more accessible for users without technical experience \cite{cypher1995kidsim, resnick2009scratch}. Using visual programming interfaces, end-users are able to author robot programs by manipulating a graphical representation of the program. In the end-user robot programming literature, visual programming interfaces commonly use flow diagrams (e.g., \cite{barakova2013end, erich2017visual}), behavior trees (e.g., \cite{barivsic2018leveraging, guerin2015framework, paxton2018evaluating}), blocks (e.g., \cite{beschi2019capirci, huang2016design, huang2017code3, moros2019programming, ramouglu2017programming, weintrop2017blockly, weintrop2018evaluating}), and icons (e.g., \cite{ryokai2009children, sapounidis2013tangible, stenmark2017simplified}) as their graphical representations. A primary weakness of the visual programming method is its limited representational power, since it can be intractable to represent every possible robot command visually. Thus, visual programming is often combined with alternate methods to enhance its expressivity, such as textual programming (e.g., \cite{barakova2013end}) or kinesthetic demonstrations (e.g., \cite{huang2017code3, stenmark2017simplified}). Visual programming requires users to program within a 2-D interface where the programming commands are disconnected from the task environment. This limitation has led to new approaches to end-user robot programming that combine visual programming concepts with spatial context (i.e., spatial-visual programming \cite{huang2020vipo}).

For some visual programming systems (e.g., \cite{datta2011end, leonardi2019trigger, liang2018simultaneous, liang2019end, matthaiakis2017flexible, pedersen2015gesture, racca2020interactive, schou2018skill}), end-users specify robot programs using Graphical User Interfaces (GUIs) that makes use of common interactive visual components for authoring programs, such as windows, buttons, input forms, menus, and sliders. GUIs generally provide pre-specified programming options rather than allowing open-ended user input. Some systems also include simulations of the robot in their GUI for visualization and programming purposes (e.g., \cite{matthaiakis2017flexible, liang2019end}). GUI-based end-user robot programming emphasizes familiar, intuitive interaction styles and visualization of program information, but it may still result in high cognitive burden for end-users because of the difficulty of programming robot capabilities meant for 3-D task environments using 2-D screens \cite{gadre2019end, huang2020vipo, ramouglu2017programming, weintrop2017blockly, weintrop2018evaluating}. This drawback of GUIs for end-user robot programming has contributed to increased interest in augmented and mixed reality-based robot programming in recent years.

\emph{Augmented/Mixed Reality.} Augmented Reality (AR), in which the real-world environment is supplemented with virtual information, and Mixed Reality (MR), in which physical and virtual objects can interact, are increasingly leveraged as mediums for end-user robot programming. AR- and MR-based robotics applications have been commonly used in industrial environments \cite{gattullo2019towards, fang2014novel}, where the need for  safety awareness may limit the use of potentially distracting interfaces that are disjoint from the working environment, and most AR- and MR-based programming systems are intended for programming industrial robots and manufacturing tasks (e.g., \cite{gao2019pati, kapinus2019spatially, ong2020augmented, quintero2018robot}). Augmented and mixed reality interfaces for end-user robot programming enhance the physical environment with virtual overlays through the use of head-mounted displays (e.g., \cite{bambuvssek2019combining, gadre2019end, ong2020augmented, quintero2018robot}), projectors (e.g., \cite{bambuvssek2019combining, gao2019pati}), or mobile devices (e.g., \cite{kapinus2019spatially}). End-users are able to interact with virtual items in the augmented environment using gestures (e.g., \cite{bambuvssek2019combining, gadre2019end, gao2019pati, quintero2018robot}), touch (e.g., \cite{bambuvssek2019combining}), speech (e.g., \cite{quintero2018robot}), direct interaction (e.g., \cite{bambuvssek2019combining, gadre2019end, kapinus2019spatially}), or using devices, such as pointers (e.g., \cite{ong2020augmented}) or gesture control bands (e.g., \cite{quintero2018robot}). End-user program specification is performed by indicating task objects and locations for manipulation (e.g., \cite{bambuvssek2019combining, gao2019pati}), motion trajectories (e.g., \cite{gadre2019end, ong2020augmented, quintero2018robot}) or the high-level, situated flow of the program (e.g., \cite{kapinus2019spatially}). AR- and MR-based end-user robot programming overcomes the limitations of other programming methods that may require end-users to be collocated with the robots being programmed, to translate 3-D environments onto 2-D screens, and to perform context switches between the robot and the programming interface. Furthermore, they may also better enable non-anthropomorphic robots to convey intent during interactive programming \cite{hedayati2018improving, coovert2014spatial}. However, augmented reality interfaces may be susceptible to perceptual issues during end-user robot programming because of issues such as occluded or constrained viewpoints and difficult depth perception \cite{kruijff2010perceptual}. Designers of AR-based programming systems may need to consider these issues carefully, as perceptual issues caused by hardware limitations or overdraw during object rendering may result in higher user cognitive load \cite{bambuvssek2019combining}.  

\emph{Demonstration.} A common and intuitive technique end-users use to program robots is demonstration of the capabilities to be programmed (e.g., \cite{jha2015application, young2014design}). A common form of demonstration used for program specification is kinesthetic demonstration, which is provided by end-users using \emph{kinesthetic teaching}, also known as lead-through programming. During kinesthetic teaching, users physically maneuver the robot through the desired robot motion trajectory. Kinesthetic teaching requires low cognitive load for users \cite{bambuvssek2019combining, quintero2018robot}; allows programmers to have better control of a robot's movements compared to alternative methods, such as direct manipulation of a virtual image of the robot \cite{manohar2014programming}; and can be faster and more intuitive for end-users because it works within the robot's configuration space \cite{ong2020augmented}, making it the predominant end-user robot programming method in robotics \cite{weintrop2017blockly} and a popular programming method in scientific literature (e.g., \cite{cakmak2014teaching, liang2019end, riedl2019fast, schou2018skill, stenmark2017simplified}). Despite its advantages and widespread use, kinesthetic teaching has limitations, including the requirement for the robot to be physically present during programming \cite{bambuvssek2019combining, manohar2014programming}, which may lessen workplace productivity \cite{ong2020augmented}; the physical demands placed on the user in physically guiding the robot \cite{bambuvssek2019combining, quintero2018robot, ajaykumar2020user}, which may make it infeasible to program heavy industrial robots using this method \cite{quintero2018robot}; and the need for the robot being programmed to be equipped with sensors and motors that enable kinesthetic demonstrations \cite{manohar2014programming}.

\emph{Natural Language.} Natural language is commonly employed for end-user robot programming in the form of speech (e.g., \cite{alexandrova2014robot, beschi2019capirci, cakmak2014teaching, gorostiza2011end, huang2017code3, quintero2018robot}) and text (e.g., \cite{beschi2019capirci, buchina2016design, buchina2019natural}). Speech-based programming in particular has the  advantage of easy accessibility for most users because of its basis in innate human communication, including for users with physical limitations or who have difficulties spelling \cite{kubota2020jessie}. However, speech-based programming is subject to vulnerabilities, such as speech misrecognition in noisy environments, which may be common in industrial workplaces where end-user robot programming systems may be deployed \cite{beschi2019capirci}, and is constrained to work with programming commands that are easy to describe verbally \cite{gadre2019end}. Perhaps due to the potentially unreliable nature of natural language recognition, end-user robot programming systems that use natural language also make use of additional programming methods, such as kinesthetic teaching (e.g., \cite{alexandrova2014robot, cakmak2014teaching, huang2017code3}), GUI visualizations (e.g., \citep{alexandrova2014robot}), augmented reality (e.g., \cite{quintero2018robot}), and visual programming (e.g., \cite{beschi2019capirci, buchina2016design, buchina2019natural, huang2017code3}), or additional input modalities, such as gestures (e.g., \cite{gorostiza2011end, quintero2018robot}).

\emph{Tangible.} Tangible programming is a programming method where end-users can author a program by indicating a program's structure and instructions using physical objects situated in the real-world environment. Common objects used for tangible programming include cards (e.g., \cite{kubota2020jessie, ryokai2009children}) and blocks (e.g., \cite{sapounidis2013tangible, sefidgar2017situated, sefidgar2018robotist}). Tangible programming often involves high-level commands, which can include task-level concepts (e.g., \cite{kubota2020jessie}), robot behaviors (e.g., \cite{ryokai2009children}), or directional or goal-based aspects of motions (e.g., \cite{sefidgar2017situated, sefidgar2018robotist, sapounidis2013tangible}). Due to its simplicity and ease of use, tangible programming systems are often designed for use by children (e.g., \cite{ryokai2009children, sapounidis2013tangible}), though recent systems have expanded their use to general end-users (e.g., \cite{sefidgar2017situated, sefidgar2018robotist}) and domain specialists (e.g., \cite{kubota2020jessie}). Unlike other programming systems that combine multiple programming methods, 
tangible robot programming systems generally rely on interaction with physical objects alone for program specification, making it difficult for tangible robot programming to be used in specifying and modifying longer and more complex programs. 
\subsubsection{Programming Features}
End-user robot programming systems include various programming features common to computer programming, such as conditional instructions 
and loops. 
Since robotics often involves the simultaneous processing of multiple sensor streams and actuation of various motors, end-user robot programming systems may also allow end-users to specify elements of parallelization (e.g., \cite{barakova2013end, gorostiza2011end, guerin2015framework, kapinus2019spatially, leonardi2019trigger}) and event-driven programming (e.g., \cite{beschi2019capirci, guerin2015framework}). 
Depending on the background of the intended end-users, designers may also choose to add more advanced programming features into end-user robot programming systems, such as nesting (e.g., \cite{alexandrova2015roboflow}) or functions and recursion (e.g., \cite{huang2020vipo}). To reduce the complexity of working with programming features for users without coding experience, some systems embed concepts such as conditions (e.g., \citep{alexandrova2015roboflow, sefidgar2017situated, sefidgar2018robotist}) and looping (e.g., \cite{pedersen2015gesture, sefidgar2017situated, sefidgar2018robotist}) within commands so that users do not explicitly have to work with statements like \textit{if/else} or \textit{while}.

End-user robot programming systems often involve online (e.g., \cite{alexandrova2014robot, matthaiakis2017flexible, quintero2018robot}) or offline (e.g., \cite{alexandrova2014robot, beschi2019capirci, manohar2014programming, matthaiakis2017flexible, weintrop2017blockly, weintrop2018evaluating}) programming features. Online programming features, such as those that use kinesthetic teaching, require the use of a real, physical robot. In contrast, offline programming features enable users to program using a virtual simulation of a robot. While interacting with a physical robot may give end-users a better idea of the robot's capabilities and may be less computationally intensive than simulation-based programming, offline programming features can help reduce user effort and program suboptimalities \cite{ong2020augmented}, cost \cite{pan2012recent}, difficulties in modifying and developing programs \cite{pan2012recent}, and idle time in waiting for a robot to become available \cite{pan2012recent}. To leverage the advantages of both online and offline features, end-user robot programming systems may provide a combination of both, such as by decoupling online program specification from offline parameter specification (e.g., \cite{schou2018skill}). Alternatively, some end-user robot programming systems may offer programming capabilities that require neither physical nor virtual robot presence (e.g., \cite{bambuvssek2019combining}).

\subsection{Editing and Debugging}
End-user robot programming systems typically provide program editing tools to allow users to iteratively refine the program and remove any errors or suboptimalities. 
At the simplest level, end-users can often delete, move, copy, or clear individual commands or the entire program (e.g., \cite{cakmak2014teaching, gao2019pati, alexandrova2015roboflow, guerin2015framework, ryokai2009children, stenmark2017simplified}) or modify parameters for saved programs or templates (e.g., \cite{bambuvssek2019combining, datta2011end, gadre2019end, liang2018simultaneous, liang2019end}). 
In addition, systems focused on editing may provide additional editing functionality, including modification of the configuration or reference frame of individual points in manipulation programs (e.g., \cite{alexandrova2014robot, forbes2014robot, gadre2019end}). End-users may also use editing tools to correct system inferences for systems using techniques such as intent recognition or automated planning to automatically specify portions of the program (e.g., \cite{liang2017framework, liang2018simultaneous, liang2019end}). 
Program modification may also be achieved using more indirect feedback, such as via simple reinforcement learning (e.g., \cite{young2014design}).

While most systems support pre- and post-execution editing, some systems support real-time editing while the program is running, such as by dynamically changing a robot motion path while it is being previewed in simulation (e.g., \cite{quintero2018robot}) or executed (e.g., \cite{matthaiakis2017flexible, young2014design}). To support debugging, end-user robot programming systems may include features such as stepwise execution and program reversion (e.g., \cite{racca2020interactive, stenmark2017simplified, pedersen2015gesture}). These systems may enable editing of visualizations of the robot and the program via GUIs (e.g., \cite{alexandrova2014robot, forbes2014robot, gadre2019end, riedl2019fast}) or augmented reality (e.g., \cite{liang2019end}), which allows end-users to avoid maneuvering the physical robot to modify portions of the program. To make program editing more intuitive, end-user robot programming systems may use common concepts from other domains, such as filmstrip views or playback from video editing (e.g., \cite{riedl2019fast}). For systems focusing on editing to help end-users in providing more diverse, optimal demonstrations for subsequent robot learning, editing may be accomplished through crowdsourcing (e.g., \cite{forbes2014robot}). 

\subsection{Verification}
Verifying that a program will work correctly when executed on the robot is an important part of the programming process, especially for safety-critical environments like clinical settings where users may need to confirm that the program is safe to run around patients \cite{buchina2016design}. A program may fail to run correctly because of errors introduced during program authoring or because of errors due to hardware limitations or probabilistic system components, such as robot perception (e.g., \cite{sefidgar2018robotist}). Verification of program correctness is often the responsibility of the user, even when the program is automatically generated by the system (e.g., \cite{liang2017framework, liang2019end}), but systems may offer error feedback (e.g., \cite{weintrop2017blockly, sefidgar2018robotist}) and prevention mechanisms (e.g., \cite{barivsic2018leveraging, beschi2019capirci, weintrop2017blockly}) to ease or automate verification. Some systems check for errors during program authoring (e.g., \cite{huang2020vipo, leonardi2019trigger}), while others also perform error checking during program execution (e.g., \cite{huang2020vipo, sefidgar2018robotist, alexandrova2014robot}). Furthermore, systems using techniques like control synthesis may be able to guarantee users that their program is automatically correct-by-construction, elimating the need for manual verification (e.g., \cite{kubota2020jessie}). 
End-user robot programming systems may also include a compilation phase prior to execution (e.g., \cite{sapounidis2013tangible, sefidgar2017situated, huang2017code3, huang2020vipo}), similar to computer programming systems.

\section{Design of End-User Robot Programming Systems} \label{sec:design}
The design of end-user robot programming systems is often inspired by observed user difficulties and needs in robot programming, and designers focus on common design aspects to meet these user needs and make robot programming more accessible. We summarize common design inspirations and aspects and how they are implemented in end-user robot programming systems.

\subsection{Design Inspirations}
\label{designins}
End-user robot programming systems are often developed with user needs and domain requirements in mind. For example, end-user robot programming systems are commonly developed using a Human-Centered Design (HCD) process, where the system design is iteratively updated based on user evaluations and feedback. 
The HCD design process often involves observation of or interviews with potential end-users at the domain the system is intended for (e.g., \cite{huang2016design, huang2020vipo}), which may help designers understand design requirements and goals for their system \cite{huang2020vipo, stenmark2017simplified}. Furthermore, designers may follow a bottom-up design approach where user characteristics and needs are used to inspire design (e.g., \cite{beschi2019capirci}). The HCD process may be applied for both the design of the programming system as well as of any supplementary instructional materials that will accompany it \cite{cakmak2014teaching}. 
End-user robot programming systems may also take inspiration from interfaces from other fields and domains, such as projected interfaces in human-computer interaction (e.g., \cite{gao2019pati}), programming environments in education (e.g., \cite{weintrop2017blockly}), or editors in multimedia (e.g., \cite{riedl2019fast}). 

\subsection{Design Aspects}
End-user robot programming systems prioritize some common design aspects to help provide a more flexible, robust, and easy robot programming experience for users. We detail the aspects commonly considered in the design of end-user robot programming systems.

\subsubsection{Reusability}
\label{generalizability}
End-user robot programming systems commonly enable end-users to reuse elements, or the entirety, of their programs. Reusability can be enabled at multiple levels within an end-user robot programming scenario, from reusing robot primitives or modules of a program to reusing a program in a different domain \cite{barakova2013end}. Furthermore, program parameters, such as locations, may also be reused (e.g., \cite{weintrop2018evaluating}). Reusability can be a valuable design aspect for end-user robot programming systems in improving programming efficiency and simplicity \cite{stenmark2017simplified}.

To enhance reusability, end-user robot programming systems may follow a modular design where the program is broken into independent modules that can be reused within or shared among programs (e.g., \cite{barakova2013end, huang2016design}). In many cases, end-users may develop their own custom modules that group together specific portions of the program that they would like to reuse (e.g., \cite{barivsic2018leveraging, gadre2019end, guerin2015framework, stenmark2017simplified}). End-user robot programming systems may also enable reusability using traditional modular programming constructs like functions and recursion (e.g., \cite{huang2020vipo}). 
Program templates are another popular mechanism to support reusability in end-user robot programming. Program templates enable end-users to program robot actions by quickly and easily parametrizing prespecified program sequences. Program templates can be created by end-users \cite{bambuvssek2019combining, gadre2019end}, or they may be predefined in the programming system for common tasks, such as screwing (e.g., \cite{matthaiakis2017flexible}) or pick-and-place (e.g., \cite{weintrop2017blockly}).
Although end-user robot programming systems for program specification do not focus on robot learning, they may still include components in their systems to help increase the generalizability, and thus reusability, of user programs to new scenarios without relying on the robot to learn how to adapt the program. Program generalizability is often emphasized in robot manipulation programming systems, as manipulation programs are often specified in terms of specific objects and positions instead of higher-level concepts that can apply to different situations. 

For robot manipulation programs, generalizability is often achieved by representing the waypoints in a motion trajectory relative to reference frames, often in reference to a landmark in the environment (e.g., \cite{alexandrova2014robot, alexandrova2015roboflow, paxton2018evaluating, stenmark2017simplified, huang2017code3}), rather than as an absolute point. Using this approach, program generalizability may be the responsibility of the user, as users can generalize a program to new environments by manually editing the reference frames for waypoints as needed when a program needs to be run for a new task or in a new setting (e.g., \cite{alexandrova2014robot, stenmark2017simplified}). Alternatively, the programming system may be responsible for helping make the manipulation program generalizable by automatically adapting the program to work with task changes, such as the number, types, configurations, or environments of task objects (e.g., \cite{alexandrova2015roboflow, liang2019end, paxton2018evaluating, matthaiakis2017flexible}), allowing flexibility in how and when programmed actions can be performed \cite{liang2019end}. Another common approach to make end-user robot programming systems and programs generalizable to different domains, tasks, and environments is to include parametric programming components that can be easily reparametrized according to the requirements of a new scenario (e.g., \cite{beschi2019capirci, guerin2015framework, liang2018simultaneous, liang2019end, pedersen2015gesture, stenmark2017simplified}). To further enhance program generalizability, end-user robot programming systems may support parameter ranges, rather than single values, for parametric components so that the system can adapt its parameters' values to new scenarios (e.g., \cite{racca2020interactive}). As a whole, more generalizable end-user robot programming helps drive robot programs towards being more reusable for a diversity of tasks \cite{liang2018simultaneous, liang2019end}.

\subsubsection{Context}
As robot programs are situated within the real, physical environment, end-user robot programming systems need to represent the task environment throughout the programming process. While some end-user robot programming systems are inherently situated in the task environment through the use of online, AR, natural language, or tangible programming features, other robot programming systems may need to include additional components to help users contextualize their programs. In addition to displaying a representation of the program structure, these systems may include visualizations of the robot (e.g., \cite{alexandrova2014robot, weintrop2017blockly}),
task objects (e.g., \cite{huang2017code3, liang2019end}), and surfaces in the environment (e.g., \cite{alexandrova2014robot}). For manipulation programs, programmed motion paths and robot poses (e.g., \cite{ong2020augmented, alexandrova2014robot, forbes2014robot, quintero2018robot, gadre2019end, liang2019end, riedl2019fast}) or the high-level program sequence (e.g., \cite{kapinus2019spatially}) may also be visualized within the environment. For navigation programs, where a 2-D representation is sufficient for conveying spatial information, visualizations of robots and relevant objects may be represented at a lower fidelity (e.g., \cite{huang2020vipo}). Some systems assume that the task environment will remain static during the programming process, with objects and landmarks being predefined (e.g., \cite{liang2019end}), while others assume dynamic, flexible task environments (e.g., \cite{gao2019pati, matthaiakis2017flexible}), where real-time changes to the environment are supported (e.g., \cite{alexandrova2015roboflow}). 
For programs involving interactive robot behaviors, end-users may benefit from social or interaction context. Social context is especially critical when end-users demonstrate programs by acting out robot behaviors, as contextual cues can support effective acting \cite{schmidt2002long}. Social context can take the form of nonverbal or verbal behaviors (e.g., \cite{manohar2014programming}) or collaborative scenarios (e.g., \cite{young2014design}). 

\subsubsection{Transparency}
End-user robot programming systems help increase the transparency of robot programming by helping end-users in developing accurate mental models of robots and programming systems. Accurate mental models are essential in helping users effectively learn to operate devices \cite{kieras1984role} and use programming commands properly \cite{sefidgar2017situated}, helping novice programmers avoid programming errors \cite{huang2016design} and better align their perceived and actual task performance \cite{weintrop2018evaluating}, and helping users better understand the robot behavior resulting from complex programs \cite{leonardi2019trigger} and make use of advanced robotic capabilities \cite{paxton2018evaluating}. Supporting accurate mental models may be especially important for systems that include probabilistic components, such as speech recognition, that could lead to unintended changes to the program that the user may fail to notice \cite{huang2017code3}. 

To help users form clearer mental models during the programming process, end-user robot programming systems may contextualize the program within the physical environment so that end-users can understand what is visible to the robot and visualize where commands will occur in the real world (e.g., \cite{alexandrova2014robot, sefidgar2018robotist}). In addition, end-user robot programming systems may use direct mapping techniques to make the process of robot programming easier and draw off of end-users' existing mental models, such as by directly mapping the programming environment to the spatial task environment (e.g., \cite{huang2020vipo}), directly mapping program primitives to human actions (e.g., \cite{pedersen2015gesture}), or directly mapping program building to jigsaw puzzle building to help users understand what commands can be connected sequentially (e.g., \cite{huang2017code3}). Furthermore, systems may display the real-time status of the system and the robot as it is programmed so that users can better understand the effects of their programming actions as well as causes of errors or failures (e.g., \cite{huang2020vipo, sefidgar2018robotist}). Besides displaying the state and status of robot programs, systems may better enhance transparency of robot programming by displaying the ``thought process'' of the robot as it executes programmed operations, such as by displaying its current state in terms of inputs, outputs, and commands (e.g., \cite{ryokai2009children}) or its interpretation of the current programming command (e.g., \cite{sefidgar2018robotist, liang2019end}), to help users understand their programs' effects in real-time.

Effective robot programming also requires end-users to have a good understanding of what robots are and are not capable of. End-users who do not have prior experience with robots may be susceptible to misperceptions and incorrect mental models of robot capabilities (e.g., \cite{liang2017framework}) due to robots' physical \cite{fussell2008people, powers2006advisor}, speech \cite{cha2014effects}, movement \cite{cha2015perceived}, and behavioral \cite{lee2005can} characteristics, as well as due to overblown representations of robots in the media \cite{sandoval2014human}. This can lead to novice robot programmers having incorrect expectations on what a robot and programming system can accomplish during robot programming. Thus, end-user robot programming systems often include elements in the system to better convey robot and system capabilities. Robot capabilities may be communicated by representing the reachability of different points in space by the robot for manipulation tasks (e.g., \cite{ong2020augmented}), providing all programmable robot capabilities as programming commands (e.g., \cite{kubota2020jessie}) and providing a status bar indicating their real-time feasibility (e.g., \cite{huang2020vipo, sefidgar2018robotist}), specifying which task and programming objects are detected by the robot (e.g., \cite{sefidgar2018robotist}), and indicating the workspace in which the robot can operate (e.g., \cite{sefidgar2018robotist}). For end-user robot programming methods that may appear to have an unlimited space of programming options, such as natural language-based programming, it is especially important for systems to communicate the constrained set of programming commands that the robot is capable of executing \cite{buchina2016design}.

\subsubsection{Abstraction}
Abstraction is a common design aspect of end-users robot programming systems that helps simplify the complexity involved in robot programming. Abstraction is inherently part of end-user robot programming systems, which abstract away processes like motion planning or localization. However, some programming systems may include more abstraction than others, depending on whether they choose to abstract away computer programming concepts such as conditions (e.g., \cite{alexandrova2015roboflow, sefidgar2017situated}) or loops (e.g., \cite{pedersen2015gesture, sefidgar2017situated}), represent programming commands at a primitive or semantic level (e.g., \cite{huang2020vipo, kubota2020jessie, paxton2018evaluating, schou2018skill}), or enable the creation of reusable modules (e.g., \cite{barakova2013end, liang2019end}). Though abstraction is a critical aspect of end-user robot programming systems, it may be difficult for users to create and use abstractions themselves \cite{green1998cognitive}. Thus, end-user robot programming systems employing abstractions such as templates or functions may need to either pre-define the abstractions for end-users or provide assistance in their development, or they could risk reducing the accessibility of the system for some users \cite{green1998cognitive}. One approach to enhance user control and flexibility in working with abstractions is to provide users with access to multiple levels of abstraction during robot programming (e.g., \cite{buchina2016design, leonardi2019trigger, ong2020augmented}), which can allow them to use or ignore abstractions as desired.

\subsubsection{Assistance}
\label{assistance}
End-user robot programming systems may offer different levels of assistance to further decrease the complexity of robot programming. At the simplest level, end-user robot programming systems may help simplify the process of manual specification of programs. Error prevention, detection, and handling mechanisms are a common example of this type of assistance (e.g., \cite{alexandrova2015roboflow, huang2016design}). At a more complex level, programming systems may use interactive, data-driven approaches to automate aspects of program specification. For example, the system may use mixed-initiative features to actively prompt the end-user to provide information that may improve the quality of their program, rather than relying on the user to optimize the program themselves (e.g., \cite{gorostiza2011end, racca2020interactive}). In a similar vein, programming systems may use automated planning techniques to assist the user by automatically generating programs (e.g., \cite{liang2017framework, liang2019end}) and control synthesis to automatically generate control code for specified high-level robot behaviors (e.g., \cite{kubota2020jessie}). Recent works have explored the use of more data-driven and probabilistic approaches to providing assistance, allowing the use of techniques like intent recognition to infer task goals (e.g., \cite{liang2018simultaneous}) or active learning to help end-users narrow down the range of feasible values for continuous program parameters (e.g., \cite{racca2020interactive}).

\section{Evaluation of End-User Robot Programming Systems} \label{sec:eval}
In this section, we summarize common components and approaches in the design of user evaluations for end-user robot programming.

\subsection{Methods}
User evaluations of end-user robot programming systems often seek to evaluate system use for program authoring, with users being instructed to program a specific or open-ended task. In some cases, it may be desirable to evaluate different or more nuanced aspects of the system use. For this reason, in addition to tasks related to program authoring, users may be assigned different tasks testing their comprehension of existing programs or their ability to debug erroneous programs (e.g., \cite{alexandrova2015roboflow, erich2017visual, huang2020vipo, sefidgar2017situated}). User evaluations are often conducted using controlled experiments, but may also take form in case (e.g., \cite{huang2016design, huang2017code3, stenmark2017simplified}), field (e.g., \cite{datta2011end}), and observational studies (e.g., \cite{huang2017code3}). Observation tends to be favored over controlled experiments when the goal is to evaluate the expressivity of an end-user robot programming system for expert users (e.g., \cite{guerin2015framework, liang2019end}). Evaluations may be conducted not only to understand user experiences in using an end-user robot programming system, but also in the programming process as a whole. For example, user evaluations may seek to understand the effectiveness of supplementary materials or instructions for end-user robot programming systems in addition to the effectiveness of the programming systems (e.g., \cite{cakmak2014teaching, ramouglu2017programming}). User evaluations generally involve one participant at a time, but user evaluations involving children often allow pairs of users to program at a time (e.g., \cite{moros2019programming, ryokai2009children, sapounidis2013tangible}).

End-user robot programming systems may be evaluated alone or against comparative examples or baseline conditions. Comparative examples and evaluation baselines may involve robot programming using existing commercial (e.g., \cite{barivsic2018leveraging, buchina2016design, erich2017visual, manohar2014programming, ong2020augmented, stenmark2017simplified, weintrop2018evaluating}) or academic (e.g., \cite{barivsic2018leveraging, gao2019pati}) programming systems. 
For systems that are intended for use by expert programmers as well as novices, traditional textual programming languages may serve as the baseline programming method (e.g., \cite{stenmark2017simplified}). 
For user studies seeking to evaluate system learnability, the baseline condition can involve the use of the same version of the system without instructions (e.g., \cite{sefidgar2017situated}) to gauge the immediate understanding of users who do not have any training on how to use the system.

\subsection{Data}
Most evaluations collect data about users' demographics and backgrounds, which often include questions about their attitudes towards and prior experience with technology (e.g., \cite{cakmak2014teaching}), programming (e.g., \cite{leonardi2019trigger}), and robots (e.g., \cite{stenmark2017simplified}), to determine whether there is any correlation between user characteristics and various study measures (e.g., \cite{leonardi2019trigger, stenmark2017simplified}). We describe common categories of study measures used in user evaluations for end-user robot programming.

\subsubsection{Success and Quality of User-Authored Programs}
Many user evaluations for end-user robot programming systems evaluate the success and quality of programs developed using the end-user robot programming system. We summarize common measures related to task and program success and quality.

\emph{Task Success.} 
A common task measure used to evaluate whether a programming system enables effective robot programming is task success (e.g., \cite{alexandrova2014robot, cakmak2014teaching, gadre2019end, stenmark2017simplified}). Task success is measured by checking whether a user-authored program results in successful execution (e.g., \cite{huang2017code3, quintero2018robot}) or meets pre-defined standards of correctness (e.g., \cite{huang2020vipo, liang2018simultaneous, weintrop2018evaluating}). 

\emph{Subjective Evaluation of Program Quality.} When the programming task relates to developing stylistic or social robot behaviors, task success may be based on more subjective measures related to the user-authored program, such as discernability (e.g., \cite{manohar2014programming}). In this case, the user-authored program may need to undergo evaluation by other study participants to determine its quality (e.g., \cite{manohar2014programming, young2014design}). 

\emph{Similarity.} In some cases, a program's quality may  be determined by its similarity to other programs, such as programs from previously related work (e.g., \cite{manohar2014programming}) or experts' programs (e.g., \cite{racca2020interactive}). Similarity-related measures may also be used to evaluate the operation of the system. For instance, for systems that automatically generate a program by imitating a user demonstration, the evaluation may measure the dissimilarity of the generated program from the user's original demonstration to understand how faithfully the system managed to recreate the user's demonstration as a robot manipulation program (e.g., \cite{jha2015application}).

\emph{Generalizability.} 
Generalizability tests are often performed for programs created by experts, most frequently by the system developers themselves (e.g., \cite{liang2019end}). However, the generalizability of programs created by study participants may also be evaluated. This is commonly accomplished by having the participant develop the program and then having the experimenter test its generalizability in a new scenario after the study is over (e.g., \cite{alexandrova2014robot, forbes2014robot}). 

\emph{Complexity.} User-authored programs may be evaluated for their complexity. Depending on the goal of the end-user robot programming system, complexity may be either a desired or undesired quality of a program. For example, for systems meant for the development of complex programs, program complexity may be evaluated to indicate whether users are able to master the use of the system to create more involved programs (e.g., \cite{buchina2016design}). 
\subsubsection{Programming Time and Progress}
Time and progress are often the focus of user evaluations for end-user robot programming. Around 64\% of the surveyed papers included user evaluations that measure task and programming time, which can be used to get an idea of how efficient end-user robot programming is using the evaluated system.
Another task metric used to understand user efficiency while programming an end-user robot programming system is task progress, or the number of tasks that a user managed to complete during the study (e.g., \cite{buchina2016design, gadre2019end, huang2017code3, weintrop2018evaluating}). While metrics such as time and progress may provide an idea of the programming efficiency enabled by an end-user robot programming system, these measures are not infallible representations of efficiency since they may be affected by other factors. For instance, studies have suggested that expert programmers may have higher task times because they spend more time exploring a system beyond the requirements of a task \cite{liang2019end} and that some programming methods are inherently faster than others, which may affect task progress measures in studies comparing different programming methods (e.g., \cite{quintero2018robot}). Furthermore, high task times may be desirable in some circumstances, as spending more time on earlier tasks may be an effective strategy for programming by helping users master programming features \cite{alexandrova2014robot, liang2019end}.

In addition to efficiency, time metrics may highlight the user's experience and understanding of the system. For example, measuring the amount of time the user spends practicing before they begin programming using a system may suggest the immediate learnability of the system (e.g., \cite{gao2019pati}). While longer interaction times are generally considered negative when considering aspects such as system efficiency and learnability, they may be desirable when the end-user robot programming system is meant for edutainment. For example, interaction time may be measured as part of the user evaluation to assess how engaged a user is when programming using the system (e.g., \cite{gorostiza2011end}). 

\subsubsection{User Perceptions} 
User evaluations often seek to understand user perceptions of the end-user robot programming process, including their perceptions of the system, the robot, and the programming tasks. User perceptions of the system may be measured by having users give subjective evaluations of the system and its features in terms of their ease to learn (e.g., \cite{weintrop2018evaluating}) and use (e.g., \cite{huang2016design, sapounidis2013tangible, weintrop2018evaluating, moros2019programming}), functionality (e.g., \cite{pedersen2015gesture}), usefulness (e.g., \cite{alexandrova2014robot, liang2018simultaneous, moros2019programming, sefidgar2018robotist}), effectiveness (e.g., \cite{barivsic2018leveraging}), naturalness (e.g., \cite{gadre2019end}), and intuitiveness (e.g., \cite{pedersen2015gesture}). The perceived usability of an end-user robot programming system is widely measured in user evaluations. The most common tool for measuring system usability in user evaluations for end-user robot programming systems is the System Usability Scale (SUS) (e.g., \cite{bambuvssek2019combining, beschi2019capirci, gadre2019end, huang2017code3, kapinus2019spatially, kubota2020jessie, leonardi2019trigger, liang2019end, paxton2018evaluating, racca2020interactive, ramouglu2017programming}), an established 10-item questionnaire that produces a system usability score \cite{brooke1996sus}. User evaluations may also use variations of the Cognitive Dimensions Framework \cite{green1996usability} to evaluate the usability of the system along various cognitive dimensions (e.g., \cite{barakova2013end, buchina2016design, huang2020vipo}). Similarly, users may evaluate the user experience of the system using established questionnaires such as the User Experience Questionnaire (UEQ) \cite{laugwitz2008construction} (e.g., \cite{bambuvssek2019combining, kapinus2019spatially}). In addition, users' perceptions of the programming system may also be better understood by measuring their satisfaction with the system (e.g., \cite{barivsic2018leveraging, weintrop2018evaluating}), as well as by asking open-ended questions about their experience, opinions, and suggestions related to the system and its features. 
Open-ended comments may be subsequently grouped to represent common user perceptions using a Grounded Theory approach \cite{strauss1994grounded} (e.g., \cite{kubota2020jessie, weintrop2018evaluating, young2014design}). Furthermore, users may be asked about their willingness to use a system again (e.g., \cite{moros2019programming}) and about their preference for the system relative to alternate programming systems (e.g., \cite{sapounidis2013tangible, sefidgar2018robotist}).

In addition to users' perceptions and subjective evaluation of the end-user robot programming system, user evaluations may investigate users' perceptions of the robot being programmed to understand how using an end-user robot programming system affects a user's mental model of and relationship with the robot. User studies may include questions about the user's overall perception of the robot (e.g., \cite{barakova2013end}) or their evaluation of how the robot performed (e.g., \cite{alexandrova2014robot}). Users' perceptions of the robot being programmed may also be observed indirectly. For instance, the number of interactions with the robot may be measured to understand how engaging a user perceives the robot, and the programming system as a whole, to be (e.g., \cite{datta2011end}). 

User evaluations often aim to understand user perceptions of the programming tasks, since these tasks are representative of real-world programming scenarios the system is intended for. To understand how easy robot programming is using a system, user evaluations may include questions about the users' perceived difficulty (e.g., \cite{alexandrova2014robot, cakmak2014teaching, ramouglu2017programming}) or success (e.g., \cite{cakmak2014teaching}) for the programming tasks used in the study. For programming systems meant for edutainment, user evaluations may also ask users how fun or enjoyable they found the programming tasks to be (e.g., \cite{moros2019programming, sapounidis2013tangible}).

\subsubsection{User Effort}
User evaluations may include measures related to user effort to understand the extent to which an end-user robot programming system alleviates an end-user's burden during robot programming. User effort may be measured through the number of programming attempts the user makes to achieve a task (e.g., \cite{young2014design}). The most common tool for measuring user workload during end-user robot programming is the NASA Task Load Index (NASA-TLX) (e.g., \cite{alexandrova2014robot, bambuvssek2019combining, cakmak2014teaching, erich2017visual, forbes2014robot, gadre2019end, gao2019pati, kapinus2019spatially, quintero2018robot}), a subjective tool for assessing a user's workload while working with a system in terms of mental, physical and temporal demand; performance; effort; and frustration \cite{hart1988development}.

\subsubsection{User Understanding}
User evaluations for end-user robot programming systems may seek to measure the level of user understanding on how to effectively program using the system. Gaps in user understanding may be observed by measuring the number, rate, and type of errors users make while using the system (e.g., \cite{alexandrova2015roboflow, gao2019pati, weintrop2018evaluating})
and, in cases where the experimenter is available for assistance, the number of help requests they make to the researcher running the evaluation study (e.g., \cite{cakmak2014teaching, huang2017code3, schou2018skill}).
Lack of user understanding may also be identified by observing the challenges or difficulties encountered by users during programming (e.g., \cite{beschi2019capirci, kubota2020jessie, sefidgar2018robotist, stenmark2017simplified}). Users may also be directly tested on their understanding of different system features (e.g., \cite{sefidgar2017situated}), or they may be tested on their understanding of how to use the system before beginning the study to discover correlations between pre-study understanding with final task performances (e.g., \cite{liang2019end}).

\subsubsection{User Behavior}
In addition to the data collected from questionnaires and interviews, data from observations of users may provide insight into user behaviors while using an end-user robot programming system. Data from observing user behavior may provide information on how frequently different system features are used \cite{alexandrova2014robot, huang2017code3, liang2019end, ryokai2009children} and how different user groups approach programming tasks using the system \cite{alexandrova2015roboflow, pedersen2015gesture, sapounidis2013tangible, young2014design}. It may also help showcase unexpected participant behaviors that the system may not be designed for \cite{bambuvssek2019combining}. Evaluators may use techniques such as encouraging users to think aloud while using the system to understand common user behaviors and thought processes during programming (e.g., \cite{barakova2013end, ramouglu2017programming, young2014design}).

\subsection{Participants}
If possible, user evaluations will include participants from the intended user base of the programming system being evaluated. These may include specific populations, such as domain specialists (e.g., \cite{barakova2013end, buchina2016design, kubota2020jessie, riedl2019fast}) or children (e.g., \cite{moros2019programming, ryokai2009children, sapounidis2013tangible}) with or without programming experience. Because end-user robot programming is intended to be approachable for those without technical backgrounds, user evaluations may focus on the recruitment of participants without experience in robotics, 
in the programming method used by the system, 
and in programming in general. 
However, many evaluations also invite participation from experts in these technical areas. 
Conducting user evaluations with expert users, in addition to novice robot programmers, may be beneficial for gaining insights into how to improve a programming system \cite{alexandrova2014robot, young2014design} and serve a broader range of users who could benefit from the system \cite{young2014design} and into how well the system closes the performance gap between  novice robot programmers and experts \cite{schou2018skill}.

\subsection{Instruction and Training}
A common practice in user evaluations for end-user robot programming systems is to provide some form of instruction and training on how to use the systems to users before they begin performing the study tasks. This training may seek to expose users to not just the procedure for programming using the system, but also to special hardware needed to use the system (e.g., \cite{bambuvssek2019combining}). Common instruction materials include one-page reference sheets (e.g., \cite{alexandrova2014robot, weintrop2018evaluating}) 
and video (e.g., \cite{forbes2014robot, huang2020vipo})
or written (e.g., \cite{cakmak2014teaching, gao2019pati}) tutorials. The experimenter may also provide a demonstration (e.g., \cite{alexandrova2014robot, alexandrova2015roboflow}) or verbal explanation 
of how to use the programming system. Occasionally, tutorials and program examples may be embedded in the system interface itself (e.g., \cite{ramouglu2017programming}). 
Users may be given an opportunity to perform practice tasks (e.g., \cite{bambuvssek2019combining, huang2016design, kapinus2019spatially, liang2019end, paxton2018evaluating}) and to seek clarifications for any confusion they may still have (e.g., \cite{paxton2018evaluating}) before performing the experimental tasks. 
Furthermore, experimenters may choose not to provide any instruction or training on how to program using the system when they want to observe the immediate understanding of the users without any priming (e.g., \cite{sefidgar2017situated}).

\subsection{Tasks}
The study tasks used in user evaluations for end-user robot programming systems may represent common robot actions, domain-specific activities, or specific phases of end-user robot programming. Study tasks involving common robot actions include pick-and-place tasks (e.g., \cite{gao2019pati, paxton2018evaluating})
and human-robot or robot-robot interaction tasks, such as authoring scenarios commonly encountered in robot-facilitated therapy (e.g., \cite{buchina2016design}) or robot emotions and roles for multi-robot interaction (e.g., \cite{young2014design}). Study tasks representing domain-specific activities cover daily living tasks such as cleaning, retrieving and storing household items, and folding towels (e.g., \cite{alexandrova2014robot, alexandrova2015roboflow, cakmak2014teaching, huang2017code3, quintero2018robot, stenmark2017simplified}); manufacturing tasks such as machine tending, circuit board testing, and welding (e.g., \cite{guerin2015framework, kapinus2019spatially, ong2020augmented, pedersen2015gesture, quintero2018robot, riedl2019fast, schou2018skill}); and store shelving tasks (e.g., \cite{liang2018simultaneous}). While most study tasks focus on testing program authoring, they may also test other phases of end-user robot programming, such as parametrization (e.g., \cite{bambuvssek2019combining}), editing (e.g., \cite{bambuvssek2019combining, gadre2019end, liang2017framework}), and debugging (e.g., \cite{kapinus2019spatially}). Users may also go through an open-ended task where they freely interact with the programming system to develop their desired program (e.g., \cite{kubota2020jessie, leonardi2019trigger, young2014design}).

User evaluations for end-user robot programming systems generally involve multiple study tasks. Often, the number and type of tasks are chosen such that they fit within a prescribed time limit. Study tasks may be given to users in order of increasing complexity (e.g., \cite{beschi2019capirci, cakmak2014teaching}) or with decreasing instruction (e.g., \cite{barakova2013end}) to challenge users as they gain experience in using a system. Furthermore, different study tasks may be given to users of different levels of expertise (e.g., \cite{huang2017code3}).

\section{Discussion} \label{sec:discussion}
In this survey, we presented an overview of design and evaluation approaches used for end-user robot programming systems to provide users with the ability to program, edit and debug, and execute programs for various robots without requiring expertise in robot programming. While prior research has helped drive the movement towards connecting a wider range of users with the power of robot customization, there still remain unexplored directions in the field that could further highlight techniques for better understanding and meeting user needs in robot programming. Below, we discuss open challenges and areas of research in end-user robot programming.

\subsection{Supporting Representative User Evaluations}
User evaluations are a key component in the design of end-user robot programming systems, motivating the need to develop robust and representative evaluation methods. In this section, we describe open directions for improving evaluations of robot programming systems and to help researchers and developers further understand end-user experiences in robot programming.

\subsubsection{Longitudinal Studies} 
User evaluations in the end-user robot programming literature are largely short in scope, with studies often taking place within an hour and users performing pre-defined, timed programming tasks. Although user evaluations of this form may be useful in highlighting various aspects of system use, especially in terms of immediate user understanding and first-sight preferences, they may fail to capture the complete user experience in end-user robot programming. This can be observed in the prevalence of the learning effect in user evaluations for end-user robot programming systems (e.g., \cite{alexandrova2014robot, huang2017code3, huang2020vipo, liang2019end, racca2020interactive, ramouglu2017programming, stenmark2017simplified, young2014design}), where users' task performances improve as they progress with study tasks. With users often devoting substantial study time to understanding how to use a programming system, evaluations may miss out on user experiences outside of the learning phase and in programming using more complex features that require more time to learn than a short-term study can provide. Longitudinal studies may help researchers and developers obtain a better idea of how users would use systems in the long term, as well as help them understand the dynamics of how users' system usage may change over time as they become more familiar with an end-user robot programming system. 
Further work may be needed to enable system developers to easily track how long-term usage and user assessments evolve over time \cite{barivsic2018leveraging}.

\subsubsection{Benchmark Tasks and Metrics} A primary challenge found in user evaluations for end-user robot programming systems is the difficulty of preventing irrelevant factors from influencing measures in user evaluations. For example, programming tasks requiring more physical input from users (i.e., kinesthetic teaching) take inherently longer to complete, which may make it difficult to fairly compare the task times for programming tasks involving different levels of physical involvement to measure relative efficiency \cite{alexandrova2015roboflow}. Furthermore, subjective measures, such as system usability, may be susceptible to influence from factors unrelated to the programming methods and techniques being evaluated, such as hardware unreliability \cite{bambuvssek2019combining, pedersen2015gesture} or users' perceptions of the robot being programmed \cite{ramouglu2017programming}. Additionally, the novelty effect users may experience from programming or interacting with a robot for the first time may result in overly positive or negative evaluations of the system, especially for users with unrealistic expectations about a robot's abilities. End-users of robots are far from a homogenous group and vary greatly in their backgrounds and demographics, which can in turn result in a substantial variance in how different users program \cite{alexandrova2015roboflow, pedersen2015gesture, sapounidis2013tangible, young2014design}. This variance can be seen in the prevalence of unexpected user programming behaviors that emerge during user evaluations of end-user robot programming systems (e.g., \cite{bambuvssek2019combining}). While following a strict experimental protocol, including a variety of study measures, and collecting covariate data may offset some of the variability introduced by the influence of peripheral factors on study measures, it may be infeasible for experimenters to control for every factor in an evaluation. Moving towards deriving an established benchmark set of evaluation tasks and metrics, an approach common in fields such as computer vision and natural language processing, may highlight one way to make systematic progress in end-user robot programming by standardizing the evaluation process and simplifying the process of understanding and comparing different robot programming works. 

\subsubsection{Evaluations in Real-World Settings} Recent years have seen the publication of several works in end-user robot programming enabled through collaboration between academics and companies (e.g., \cite{barivsic2018leveraging, huang2016design}), highlighting increasing commercial interests in making end-user robot programming accessible through real-world products and applications. However, academic work on end-user robot programming may present a limited view of user experiences in end-user robot programming due to a lack of deployments and user evaluations in non-laboratory settings. Laboratory settings are generally highly controlled and are set up to prioritize participant safety, which may cause participants to program less cautiously than they would in a less controlled setting (e.g., \cite{racca2020interactive}). While some systems have been deployed in real-world settings, such as on production floors (e.g., \cite{guerin2015framework, pedersen2015gesture, schou2018skill}), these deployments generally focus on evaluating the technical capabilities of programming systems in enabling programming of complex tasks rather than on how end-users use the systems as part of their typical workflows. Further exploration is therefore needed to understand the applicability of end-user robot programming systems in the research literature to real-world domains, which may have requirements and constraints that laboratory settings may fail to represent. Conducting more observational and field studies may further help in gaining insight into how users program for real-world use cases in uncontrolled lab environments, while including open-ended programming tasks where end-users freely interact with a system in user evaluations may reveal user programming behaviors outside of a timed task context. Overall, further work is needed in end-user robot programming to create experimental approaches and data measures and analyses that faithfully represent realistic end-user programming experiences.

\subsection{Supporting a Wider Range of Programming Capabilities}
Although end-user robot programming has made many aspects of robot programming accessible for end-users, there still remain key use cases of robots and aspects of robot behavior that should be available for end-user customization to further the applicability of end-user robot programming. 

\subsubsection{Programming Collaborative Tasks} 
While there have been several proposed approaches for end-user programming of human-robot interactions, end-user robot programming techniques for programming human-robot collaboration are less represented in literature. Most end-user robot programming systems today skew towards programming autonomous tasks, with any programmed human-robot interactions being simple or highly structured. Unlike social human-robot interactions, which lend themselves to highly structured programming since they can be composed of established social norms (e.g., \cite{porfirio2018authoring}), human-robot collaboration may not always take the form of rigid turn-taking interactions. It instead can involve open-ended structures that the sequential and templated constructs of conventional end-user robot programming systems may not be able to handle. Human-robot collaboration can involve an unbounded space of inputs from a human collaborative partner and abstract concepts such as intent recognition that may be difficult for end-users to program. Furthermore, human-robot collaboration is highly dependent on  external events performed by collaborative agents, making it a highly dynamic process, with a pace and form that may change over time. 

Because of the idiosyncrasies involved in human-robot collaboration compared to autonomous and structured robot actions, end-user robot programming systems for programming human-robot collaboration must provide new types of programming capabilities for end-users. For example, systems may need to move towards enabling users to specify not only \emph{what} a robot will do (i.e., task goal) but \emph{how} it will do it (i.e., style or collaborative role) (e.g., \cite{young2014design}). Since stylistic aspects and robot roles may need to change in real-time according to the collaborative context at any given moment, which end-users may not be able to anticipate during programming, end-user robot programming systems should also enable \emph{metaprogramming} whereby end-user programs can generate robot behavior in real-time as the need arises rather than requiring users to pre-author the robot behavior. End-user robot programming systems should also support concepts such as parallel programming and synchronization, which are typically difficult even for advanced programmers \cite{mckenney2017parallel},  in an intuitive manner for end-users, as collaborative techniques such as action understanding and intent recognition may need to run as a background thread while a collaborative program sequence unfolds and inputs from human collaborators may arrive simultaneously, such as co-temporal speech and gestures. In cases of collocated collaboration, end-user robot programming systems may need to prioritize meeting strict safety constraints, possibly by performing automatic verification and optimization checks on user-created programs similarly to how social human-robot interaction programs have been verified to meet social norms \cite{porfirio2018authoring}. 

Bridging the areas of end-user robot programming and human-robot collaboration may be beneficial, as end-users know their collaborative preferences and abilities best and may want to personalize the nature and timing of their collaboration with a robot to better align with these preferences and goals (e.g., \cite{wang2020see}). Similar to how robotics as a field has shifted from largely autonomous machines to interactive, collaborative technologies, end-user robot programming will need to move from enabling highly structured autonomous programming capabilities to robot capabilities that will interact with and support human actions in new task contexts such as collaborative manufacturing or co-design. 

\subsubsection{Programming Multiple Robots} 
The applicability of end-user robot programming systems to real-world scenarios may be enhanced by enabling the simultaneous programming of multiple types of robots. Due to variance in the programmable capabilities of different robots, end-user robot programming systems are often designed to work with a particular robot. Although the availability of resources such as ROS, which has helped standardize the programming process for different robot technologies, may help developers create open-source end-user robot programming systems that can be easily modified to work with different robots, most systems do not enable users to program different robots simultaneously. This may limit the use of end-user robot programming systems, as some tasks may be too complex or infeasible to perform without simultaneously programming multiple robots, such as when tasks are not collocated but require synchronization, as in a missile launch  \cite{dudek1996taxonomy}. In these task scenarios, multi-robot systems (MRSs), homogeneous or heterogeneous groups of robots, can be required. MRSs are increasingly adopted in real-world domains due to their benefits in terms of task efficiency \cite{burgard2000collaborative} and system reliability and scalability \cite{prorok2012low}. Enabling intuitive robot programming of MRSs while handling the increased user workload and diminished situation awareness associated with attending to multiple robots \cite{wong2017workload} is an open challenge that must be addressed to combine the benefits of end-user robot programming and MRSs. 

\subsubsection{Programming Social Human-Robot Interaction} 
As robots make their way into everyday environments, there is expected to be an increasing amount of social encounters between humans and robots. While there have been tools designed for easing the process of generating social behaviors for robots (e.g., \cite{huang2012robot}) and of authoring programs for social human-robot interaction (e.g., \cite{porfirio2019computational, porfirio2018authoring, porfirio2019bodystorming, porfirio2020transforming}), these tools are generally intended for expert human-robot interaction designers rather than for the end-users interacting with the robot. Introducing more programming tools for social HRI may be beneficial in allowing human interactants to customize interactions according to their preferences by, for example, specifying the robot's personality and modifying the robot's interactive behaviors to meet social norms and work appropriately in the current interaction context. End-user robot programming for social HRI may be especially beneficial in keeping end-users engaged with robots in long-term deployments by enabling them to customize robot operation to produce new behaviors over time, which may reduce habituation effects \cite{koay2007living, leite2013social}. End-user robot programming may also serve as a tool for helping populations such as older adults or people with disabilities retain their sense of independence when using the services of socially assistive robots.

Work on end-user robot programming for social human-robot interaction remains largely limited to programming of robots intended for activities such as education and therapy, where interactions involve a limited set of robot responses and are not necessarily representative of social human-robot interaction in general. Enabling end-user robot programming of a wider space of social HRI requires further investigation to determine the appropriate methods and interfaces for enabling real-time end-user customization during interaction scenarios. By investigating these open areas of research, end-user robot programming tools can be put to use to better empower end-users during social human-robot interactions and potentially reduce the possible negative consequences of social HRI, such as emotional deception, by exposing the programmable nature of robots to end-users.

\subsection{Supporting Various Degrees of Programmability}
End-user robot programming systems that emphasize expressivity are critical in making programming of complex programs accessible for a wider range of users, especially in domains such as manufacturing that may require programming of more intricate procedures such as welding. However, citizen-level programming systems that emphasize a constrained programming space may serve as an equally valid approach towards developing end-user robot programming systems for everyday environments, such as homes. For day-to-day tasks, citizen developers may not have the need or the desire to engage in the end-to-end programming processes typical of most end-user robot programming systems. They may instead want to engage in quick specification of aspects of the programs that are relevant to the need at hand, such as setting up the area to be cleaned for a vacuum cleaner robot. User interactions with everyday technologies can exemplify how robot programming should be structured for everyday needs. For example, moving towards standard interaction conventions, such as the drag-and-drop and swipe interactions end-users use to operate touchscreen interfaces, for instructing robots may help make end-user robot programming accessible when the goal is quick and easy modification and personalization of robot behaviors, rather than detailed specification of a task procedure, and may also minimize user difficulties in learning new programming conventions and methods for every new robot that they encounter (e.g., \cite{han2020structuring}).

The varying needs of different users and domains with respect to the expressivity of end-user robot programs centers around the tradeoff between expressivity and simplicity. End-user robot programming systems often seek to strike a balance between these concepts, including enough features to enable programming of non-trivial tasks while constraining system features to simplify programming for untrained users and to minimize the chance of user errors (e.g., \cite{alexandrova2015roboflow}). End-user robot programming systems that prioritize expressivity may be able to better handle a variety of use cases but may be less accessible for users without programming experience \cite{huang2016design, huang2020vipo}, who may favor a limited set of programming capabilities \cite{leonardi2019trigger}. On the other hand, end-user robot programming systems that oversimplify programming may be less favored by more advanced users, who may desire a similar level of expressivity to the amount found in traditional textual languages \cite{sefidgar2018robotist}. Discovering the ideal balance between simplicity and expressivity and the limits to which end-users can effectively make use of programming expressivity is an open challenge that may determine how end-user robot programming systems will take form in the future.

\subsection{Supporting End-Users}
End-user robot programming centralizes around the goal of making end-user robot customization attainable for those without robotics experience. We highlight open research directions for further supporting end-users in meeting their programming goals without undue difficulties.

\subsubsection{Collaborative Programming} Work on end-user robot programming has largely focused on supporting singular users in easily programming robots. While some end-user robot programming systems support robot programming by multiple users with different roles, such as clinicians and caregivers, (e.g., \cite{datta2011end}), the programming interactions are generally limited to a single user at a time. However, in real world domains, there are situations where collaborative or multi-user programming may be desired. For example, clinicians may want to collaboratively develop robot programs for interactive therapies with patients to provide patients with flexibility in choosing robot actions that meet their goals \cite{kubota2020jessie}. Furthermore, multi-user tandem programming can help ease the workload of users when developing programs involving multiple robots \cite{young2014design}. Further investigation is needed to determine methods to simultaneously support multiple end-users during robot programming while avoiding interference between users. Prior work on virtual pair programming and multi-user visual programming may highlight design and implementation tools that can be used to make end-user robot programming more collaborative and social (e.g., \cite{campbell2002multi, repenning2011collective}). Special considerations should be taken for end-user robot programming systems meant for children, where it may be necessary to provide equal opportunities and control for each user to help reduce antagonism between children in collaborative robot programming scenarios \cite{sapounidis2013tangible}. 

\subsubsection{Assisted End-User Robot Programming}
End-user robot programming has helped transform robot programming from an occupation reserved for robotics engineers to an act that everyday users and domain specialists can take to meet their personal needs. While the increased accessibility of robot programming for users without robotics experience is beneficial in empowering more users to customize robots to meet their needs, it is not without its risks, as software developed by end-users without programming experience tend to contain substantial amount of errors \cite{burnett2006next}. Without the training undergone by developers, end-users may be more likely to include workarounds and shortcuts in their programs and to misjudge their own programming abilities \cite{harrison2004editor}. Errors in robotics software developed by end-users can be especially dangerous since robot programs often include interactions with the physical world and people, including vulnerable populations such as children and patients. Therefore, there is a need for end-user robot programming systems to \emph{assist} users in identifying errors and developing more optimal, correct programs.

Assistance is commonly found in computer programming, with integrated development environments increasingly providing syntax-level assistance, such as autocorrect and autocomplete, and more personalized, intelligent assistance, such as code completion and smart suggestion. Similarly, end-user robot programming systems should offer both syntax-level and personalized  assistance to users. While end-user robot programming systems in the research literature have included assistance at the syntax-level (Section \ref{assistance}), research on personalized assistance in end-user robot programming is still incipient. To meet the needs of diverse groups of end-users while handling the idiosyncrasies of individuals' programming approaches and behaviors, end-user robot programming systems should incorporate personalized data-driven assistance as a tool, introducing intelligent tools that can identify an individual's programming intent and challenges in real-time to provide just-in-time assistance that can help end-users avoid introducing errors and suboptimalities into their programs. For example, assistance may be introduced to kinesthetic teaching to reduce the tradeoff between users' mental and physical workloads. Programming by kinesthetic demonstration may be favored by end-users for the low cognitive effort involved in kinesthetic teaching, but the method may also be tiring and difficult for users because of the high physical workload required to move robot manipulators \cite{quintero2018robot}. Introducing assistance into the kinesthetic programming workflow, such as by offering support positioning the robot when the end-user is experiencing difficulties in maneuvering a robot to a specific configuration or when they are approaching a joint limit, may reduce the physical difficulties associated with kinesthetic teaching and in turn reduce suboptimal and redundant movements in the resulting robot program. Alternatively, assistance can be provided by predicting the user's intent during programming and proactively completing the user's program (e.g., \cite{liang2018simultaneous}) or suggesting program commands that can be used to meet the user's programming goal. Different programming methods may warrant different types of assistance, such as assistance in perceiving relevant aspects of a program within a limited field of view for AR-based programming.

Including assistance in end-user robot programming systems will involve shifting more of the control over the programming process from the user to the system. Providing less control to the user during robot programming can be beneficial in reducing user workload in dealing with tedious aspects of robot programming unrelated to the programming task such as motion planning \cite{ong2020augmented, weintrop2017blockly}, in reducing imperfections in user programs such as noise in kinesthetic demonstrations due to natural hand tremors \cite{antonelli2013training}, and in increasing the flexibility of programs in working in new contexts \cite{young2014design}. However, although prior work has suggested that participants like automatic assistance provided by end-user robot programming systems \cite{liang2019end}, end-users may be unwilling to accept assistance if it means relinquishing control to automated assistance in specifying programs \cite{racca2020interactive} or if it means interference to their programming workflow \cite{robertson2004impact}. Furthermore, reduced user control could lead to disinterest in programming for younger populations \cite{sapounidis2013tangible}. Therefore, although assistance has potential to further reduce barriers that prevent everyday users from effectively programming robots, more work is needed to understand the level of assistance end-users are willing to accept and to develop assistance that supports users in programming while still empowering them in leveraging robots according to their individual needs.

\section{Conclusion} \label{sec:conclusion}
Similarly to how the pervasiveness of computers have necessitated methods for end-users to program computing technologies, the increasing number of robots in a diversity of domains, from the home to factory floors, has warranted tools to achieve end-user robot programming. The challenges associated with end-user robot programming in terms of the prerequisite programming, engineering, and robotics expertise required to effectively program robots have inspired many different techniques and methods to lower the barriers involved in customizing robot behavior. The present work presents a survey of the research literature covering end-user program specification, with a structured review of the application domains, programming phases, and design techniques and aspects encompassed by the end-user robot programming space. We highlight both past trends and future directions in better supporting users, conducting more representative user evaluations, and providing wider programming spaces and ranges of programmability. Our survey not only provides a comprehensive review of the end-user robot programming space, but also seeks to highlight the need for continuous improvement on end-user robot programming systems to truly democratize the customization of robotic technologies.

\bibliographystyle{ACM-Reference-Format}
\bibliography{2020-csur-ajaykumar-preprint}


\begin{thebibliography}{121}


\ifx \showCODEN    \undefined \def \showCODEN     #1{\unskip}     \fi
\ifx \showDOI      \undefined \def \showDOI       #1{#1}\fi
\ifx \showISBNx    \undefined \def \showISBNx     #1{\unskip}     \fi
\ifx \showISBNxiii \undefined \def \showISBNxiii  #1{\unskip}     \fi
\ifx \showISSN     \undefined \def \showISSN      #1{\unskip}     \fi
\ifx \showLCCN     \undefined \def \showLCCN      #1{\unskip}     \fi
\ifx \shownote     \undefined \def \shownote      #1{#1}          \fi
\ifx \showarticletitle \undefined \def \showarticletitle #1{#1}   \fi
\ifx \showURL      \undefined \def \showURL       {\relax}        \fi
\providecommand\bibfield[2]{#2}
\providecommand\bibinfo[2]{#2}
\providecommand\natexlab[1]{#1}
\providecommand\showeprint[2][]{arXiv:#2}

\bibitem[\protect\citeauthoryear{Ajaykumar and Huang}{Ajaykumar and
  Huang}{2020}]%
        {ajaykumar2020user}
\bibfield{author}{\bibinfo{person}{Gopika Ajaykumar} {and}
  \bibinfo{person}{Chien-Ming Huang}.} \bibinfo{year}{2020}\natexlab{}.
\newblock \showarticletitle{User needs and design opportunities in end-user
  robot programming}. In \bibinfo{booktitle}{\emph{Companion of the 2020
  ACM/IEEE International Conference on Human-Robot Interaction}}.
  \bibinfo{pages}{93--95}.
\newblock


\bibitem[\protect\citeauthoryear{Alexandrova, Cakmak, Hsiao, and
  Takayama}{Alexandrova et~al\mbox{.}}{2014}]%
        {alexandrova2014robot}
\bibfield{author}{\bibinfo{person}{Sonya Alexandrova}, \bibinfo{person}{Maya
  Cakmak}, \bibinfo{person}{Kaijen Hsiao}, {and} \bibinfo{person}{Leila
  Takayama}.} \bibinfo{year}{2014}\natexlab{}.
\newblock \showarticletitle{Robot programming by demonstration with interactive
  action visualizations.}. In \bibinfo{booktitle}{\emph{Robotics: science and
  systems}}. Citeseer.
\newblock


\bibitem[\protect\citeauthoryear{Alexandrova, Tatlock, and Cakmak}{Alexandrova
  et~al\mbox{.}}{2015}]%
        {alexandrova2015roboflow}
\bibfield{author}{\bibinfo{person}{Sonya Alexandrova}, \bibinfo{person}{Zachary
  Tatlock}, {and} \bibinfo{person}{Maya Cakmak}.}
  \bibinfo{year}{2015}\natexlab{}.
\newblock \showarticletitle{RoboFlow: A flow-based visual programming language
  for mobile manipulation tasks}. In \bibinfo{booktitle}{\emph{2015 IEEE
  International Conference on Robotics and Automation (ICRA)}}. IEEE,
  \bibinfo{pages}{5537--5544}.
\newblock


\bibitem[\protect\citeauthoryear{Antonelli, Astanin, Galetto, and
  Mastrogiacomo}{Antonelli et~al\mbox{.}}{2013}]%
        {antonelli2013training}
\bibfield{author}{\bibinfo{person}{Dario Antonelli}, \bibinfo{person}{S
  Astanin}, \bibinfo{person}{M Galetto}, {and} \bibinfo{person}{L
  Mastrogiacomo}.} \bibinfo{year}{2013}\natexlab{}.
\newblock \showarticletitle{Training by demonstration for welding robots by
  optical trajectory tracking}.
\newblock \bibinfo{journal}{\emph{Procedia Cirp}}  \bibinfo{volume}{12}
  (\bibinfo{year}{2013}), \bibinfo{pages}{145--150}.
\newblock


\bibitem[\protect\citeauthoryear{Argall, Chernova, Veloso, and Browning}{Argall
  et~al\mbox{.}}{2009}]%
        {argall2009survey}
\bibfield{author}{\bibinfo{person}{Brenna~D Argall}, \bibinfo{person}{Sonia
  Chernova}, \bibinfo{person}{Manuela Veloso}, {and} \bibinfo{person}{Brett
  Browning}.} \bibinfo{year}{2009}\natexlab{}.
\newblock \showarticletitle{A survey of robot learning from demonstration}.
\newblock \bibinfo{journal}{\emph{Robotics and autonomous systems}}
  \bibinfo{volume}{57}, \bibinfo{number}{5} (\bibinfo{year}{2009}),
  \bibinfo{pages}{469--483}.
\newblock


\bibitem[\protect\citeauthoryear{Bambu{\v{s}}sek, Materna, Kapinus, Beran, and
  Smr{\v{z}}}{Bambu{\v{s}}sek et~al\mbox{.}}{2019}]%
        {bambuvssek2019combining}
\bibfield{author}{\bibinfo{person}{Daniel Bambu{\v{s}}sek},
  \bibinfo{person}{Zden{\v{e}}k Materna}, \bibinfo{person}{Michal Kapinus},
  \bibinfo{person}{V{\'\i}t{\v{e}}zslav Beran}, {and} \bibinfo{person}{Pavel
  Smr{\v{z}}}.} \bibinfo{year}{2019}\natexlab{}.
\newblock \showarticletitle{Combining Interactive Spatial Augmented Reality
  with Head-Mounted Display for End-User Collaborative Robot Programming}. In
  \bibinfo{booktitle}{\emph{2019 28th IEEE International Conference on Robot
  and Human Interactive Communication (RO-MAN)}}. IEEE, \bibinfo{pages}{1--8}.
\newblock


\bibitem[\protect\citeauthoryear{Barakova, Gillesen, Huskens, and
  Lourens}{Barakova et~al\mbox{.}}{2013}]%
        {barakova2013end}
\bibfield{author}{\bibinfo{person}{Emilia~I Barakova}, \bibinfo{person}{Jan~CC
  Gillesen}, \bibinfo{person}{Bibi~EBM Huskens}, {and} \bibinfo{person}{Tino
  Lourens}.} \bibinfo{year}{2013}\natexlab{}.
\newblock \showarticletitle{End-user programming architecture facilitates the
  uptake of robots in social therapies}.
\newblock \bibinfo{journal}{\emph{Robotics and Autonomous Systems}}
  \bibinfo{volume}{61}, \bibinfo{number}{7} (\bibinfo{year}{2013}),
  \bibinfo{pages}{704--713}.
\newblock


\bibitem[\protect\citeauthoryear{Bari{\v{s}}i{\'c}, Cambeiro, Amaral,
  Goul{\~a}o, and Mota}{Bari{\v{s}}i{\'c} et~al\mbox{.}}{2018}]%
        {barivsic2018leveraging}
\bibfield{author}{\bibinfo{person}{Ankica Bari{\v{s}}i{\'c}},
  \bibinfo{person}{Jo{\~a}o Cambeiro}, \bibinfo{person}{Vasco Amaral},
  \bibinfo{person}{Miguel Goul{\~a}o}, {and} \bibinfo{person}{Tarqu{\'\i}nio
  Mota}.} \bibinfo{year}{2018}\natexlab{}.
\newblock \showarticletitle{Leveraging teenagers feedback in the development of
  a domain-specific language: the case of programming low-cost robots}. In
  \bibinfo{booktitle}{\emph{Proceedings of the 33rd Annual ACM Symposium on
  Applied Computing}}. \bibinfo{pages}{1221--1229}.
\newblock


\bibitem[\protect\citeauthoryear{Barricelli, Cassano, Fogli, and
  Piccinno}{Barricelli et~al\mbox{.}}{2019}]%
        {barricelli2019end}
\bibfield{author}{\bibinfo{person}{Barbara~Rita Barricelli},
  \bibinfo{person}{Fabio Cassano}, \bibinfo{person}{Daniela Fogli}, {and}
  \bibinfo{person}{Antonio Piccinno}.} \bibinfo{year}{2019}\natexlab{}.
\newblock \showarticletitle{End-user development, end-user programming and
  end-user software engineering: A systematic mapping study}.
\newblock \bibinfo{journal}{\emph{Journal of Systems and Software}}
  \bibinfo{volume}{149} (\bibinfo{year}{2019}), \bibinfo{pages}{101--137}.
\newblock


\bibitem[\protect\citeauthoryear{Beschi, Fogli, and Tampalini}{Beschi
  et~al\mbox{.}}{2019}]%
        {beschi2019capirci}
\bibfield{author}{\bibinfo{person}{Sara Beschi}, \bibinfo{person}{Daniela
  Fogli}, {and} \bibinfo{person}{Fabio Tampalini}.}
  \bibinfo{year}{2019}\natexlab{}.
\newblock \showarticletitle{CAPIRCI: a multi-modal system for collaborative
  robot programming}. In \bibinfo{booktitle}{\emph{International Symposium on
  End User Development}}. Springer, \bibinfo{pages}{51--66}.
\newblock


\bibitem[\protect\citeauthoryear{Biggs and MacDonald}{Biggs and
  MacDonald}{2003}]%
        {biggs2003survey}
\bibfield{author}{\bibinfo{person}{Geoffrey Biggs} {and} \bibinfo{person}{Bruce
  MacDonald}.} \bibinfo{year}{2003}\natexlab{}.
\newblock \showarticletitle{A survey of robot programming systems}. In
  \bibinfo{booktitle}{\emph{Proceedings of the Australasian conference on
  robotics and automation}}. \bibinfo{pages}{1--3}.
\newblock


\bibitem[\protect\citeauthoryear{Billard, Calinon, Dillmann, and
  Schaal}{Billard et~al\mbox{.}}{2008}]%
        {billard2008survey}
\bibfield{author}{\bibinfo{person}{Aude Billard}, \bibinfo{person}{Sylvain
  Calinon}, \bibinfo{person}{Ruediger Dillmann}, {and} \bibinfo{person}{Stefan
  Schaal}.} \bibinfo{year}{2008}\natexlab{}.
\newblock \showarticletitle{Survey: Robot programming by demonstration}.
\newblock \bibinfo{journal}{\emph{Handbook of robotics}} \bibinfo{volume}{59},
  \bibinfo{number}{BOOK\_CHAP} (\bibinfo{year}{2008}).
\newblock


\bibitem[\protect\citeauthoryear{Brandt, Guo, Lewenstein, and Klemmer}{Brandt
  et~al\mbox{.}}{2008}]%
        {brandt2008opportunistic}
\bibfield{author}{\bibinfo{person}{Joel Brandt}, \bibinfo{person}{Philip~J
  Guo}, \bibinfo{person}{Joel Lewenstein}, {and} \bibinfo{person}{Scott~R
  Klemmer}.} \bibinfo{year}{2008}\natexlab{}.
\newblock \showarticletitle{Opportunistic programming: How rapid ideation and
  prototyping occur in practice}. In \bibinfo{booktitle}{\emph{Proceedings of
  the 4th international workshop on End-user software engineering}}.
  \bibinfo{pages}{1--5}.
\newblock


\bibitem[\protect\citeauthoryear{Bravo, Gonz{\'a}lez, and Gonz{\'a}lez}{Bravo
  et~al\mbox{.}}{2017}]%
        {bravo2017review}
\bibfield{author}{\bibinfo{person}{Flor~A Bravo}, \bibinfo{person}{Alejandra~M
  Gonz{\'a}lez}, {and} \bibinfo{person}{Enrique Gonz{\'a}lez}.}
  \bibinfo{year}{2017}\natexlab{}.
\newblock \showarticletitle{A review of intuitive robot programming
  environments for educational purposes}. In \bibinfo{booktitle}{\emph{2017
  IEEE 3rd Colombian Conference on Automatic Control (CCAC)}}. IEEE,
  \bibinfo{pages}{1--6}.
\newblock


\bibitem[\protect\citeauthoryear{Breazeal and Scassellati}{Breazeal and
  Scassellati}{2002}]%
        {breazeal2002robots}
\bibfield{author}{\bibinfo{person}{Cynthia Breazeal} {and}
  \bibinfo{person}{Brian Scassellati}.} \bibinfo{year}{2002}\natexlab{}.
\newblock \showarticletitle{Robots that imitate humans}.
\newblock \bibinfo{journal}{\emph{Trends in cognitive sciences}}
  \bibinfo{volume}{6}, \bibinfo{number}{11} (\bibinfo{year}{2002}),
  \bibinfo{pages}{481--487}.
\newblock


\bibitem[\protect\citeauthoryear{Brooke}{Brooke}{1996}]%
        {brooke1996sus}
\bibfield{author}{\bibinfo{person}{John Brooke}.}
  \bibinfo{year}{1996}\natexlab{}.
\newblock \showarticletitle{SUS: a “quick and dirty'usability}.
\newblock \bibinfo{journal}{\emph{Usability evaluation in industry}}
  (\bibinfo{year}{1996}), \bibinfo{pages}{189}.
\newblock


\bibitem[\protect\citeauthoryear{Buchina, Kamel, and Barakova}{Buchina
  et~al\mbox{.}}{2016}]%
        {buchina2016design}
\bibfield{author}{\bibinfo{person}{Nina Buchina}, \bibinfo{person}{Sherin
  Kamel}, {and} \bibinfo{person}{Emilia Barakova}.}
  \bibinfo{year}{2016}\natexlab{}.
\newblock \showarticletitle{Design and evaluation of an end-user friendly tool
  for robot programming}. In \bibinfo{booktitle}{\emph{2016 25th IEEE
  International Symposium on Robot and Human Interactive Communication
  (RO-MAN)}}. IEEE, \bibinfo{pages}{185--191}.
\newblock


\bibitem[\protect\citeauthoryear{Buchina, Sterkenburg, Lourens, and
  Barakova}{Buchina et~al\mbox{.}}{2019}]%
        {buchina2019natural}
\bibfield{author}{\bibinfo{person}{Nina~G Buchina}, \bibinfo{person}{Paula
  Sterkenburg}, \bibinfo{person}{Tino Lourens}, {and} \bibinfo{person}{Emilia~I
  Barakova}.} \bibinfo{year}{2019}\natexlab{}.
\newblock \showarticletitle{Natural language interface for programming
  sensory-enabled scenarios for human-robot interaction}. In
  \bibinfo{booktitle}{\emph{2019 28th IEEE International Conference on Robot
  and Human Interactive Communication (RO-MAN)}}. IEEE, \bibinfo{pages}{1--8}.
\newblock


\bibitem[\protect\citeauthoryear{Burgard, Moors, Fox, Simmons, and
  Thrun}{Burgard et~al\mbox{.}}{2000}]%
        {burgard2000collaborative}
\bibfield{author}{\bibinfo{person}{Wolfram Burgard}, \bibinfo{person}{Mark
  Moors}, \bibinfo{person}{Dieter Fox}, \bibinfo{person}{Reid Simmons}, {and}
  \bibinfo{person}{Sebastian Thrun}.} \bibinfo{year}{2000}\natexlab{}.
\newblock \showarticletitle{Collaborative multi-robot exploration}. In
  \bibinfo{booktitle}{\emph{Proceedings 2000 ICRA. Millennium Conference. IEEE
  International Conference on Robotics and Automation. Symposia Proceedings
  (Cat. No. 00CH37065)}}, Vol.~\bibinfo{volume}{1}. IEEE,
  \bibinfo{pages}{476--481}.
\newblock


\bibitem[\protect\citeauthoryear{Burnett, Myers, Rosson, and
  Wiedenbeck}{Burnett et~al\mbox{.}}{2006}]%
        {burnett2006next}
\bibfield{author}{\bibinfo{person}{Margaret Burnett}, \bibinfo{person}{Brad
  Myers}, \bibinfo{person}{Mary~Beth Rosson}, {and} \bibinfo{person}{Susan
  Wiedenbeck}.} \bibinfo{year}{2006}\natexlab{}.
\newblock \showarticletitle{The next step: from end-user programming to
  end-user software engineering}. In \bibinfo{booktitle}{\emph{CHI'06 Extended
  Abstracts on Human Factors in Computing Systems}}.
  \bibinfo{pages}{1699--1702}.
\newblock


\bibitem[\protect\citeauthoryear{Cakmak and Takayama}{Cakmak and
  Takayama}{2014}]%
        {cakmak2014teaching}
\bibfield{author}{\bibinfo{person}{Maya Cakmak} {and} \bibinfo{person}{Leila
  Takayama}.} \bibinfo{year}{2014}\natexlab{}.
\newblock \showarticletitle{Teaching people how to teach robots: The effect of
  instructional materials and dialog design}. In
  \bibinfo{booktitle}{\emph{Proceedings of the 2014 ACM/IEEE international
  conference on Human-robot interaction}}. \bibinfo{pages}{431--438}.
\newblock


\bibitem[\protect\citeauthoryear{Calinon}{Calinon}{2009}]%
        {calinon2009robot}
\bibfield{author}{\bibinfo{person}{Sylvain Calinon}.}
  \bibinfo{year}{2009}\natexlab{}.
\newblock \bibinfo{booktitle}{\emph{Robot programming by demonstration}}.
\newblock \bibinfo{publisher}{EPFL Press}.
\newblock


\bibitem[\protect\citeauthoryear{Campbell}{Campbell}{2002}]%
        {campbell2002multi}
\bibfield{author}{\bibinfo{person}{Jeffrey~D Campbell}.}
  \bibinfo{year}{2002}\natexlab{}.
\newblock \showarticletitle{Multi-user collaborative visual program
  development}. In \bibinfo{booktitle}{\emph{Proceedings IEEE 2002 Symposia on
  Human Centric Computing Languages and Environments}}. IEEE,
  \bibinfo{pages}{122--130}.
\newblock


\bibitem[\protect\citeauthoryear{Cha, Dragan, Forlizzi, and Srinivasa}{Cha
  et~al\mbox{.}}{2014}]%
        {cha2014effects}
\bibfield{author}{\bibinfo{person}{Elizabeth Cha}, \bibinfo{person}{Anca
  Dragan}, \bibinfo{person}{Jodi Forlizzi}, {and} \bibinfo{person}{Siddhartha~S
  Srinivasa}.} \bibinfo{year}{2014}\natexlab{}.
\newblock \showarticletitle{Effects of speech on perceived capability}. In
  \bibinfo{booktitle}{\emph{2014 9th ACM/IEEE International Conference on
  Human-Robot Interaction (HRI)}}. IEEE, \bibinfo{pages}{134--135}.
\newblock


\bibitem[\protect\citeauthoryear{Cha, Dragan, and Srinivasa}{Cha
  et~al\mbox{.}}{2015}]%
        {cha2015perceived}
\bibfield{author}{\bibinfo{person}{Elizabeth Cha}, \bibinfo{person}{Anca~D
  Dragan}, {and} \bibinfo{person}{Siddhartha~S Srinivasa}.}
  \bibinfo{year}{2015}\natexlab{}.
\newblock \showarticletitle{Perceived robot capability}. In
  \bibinfo{booktitle}{\emph{2015 24th IEEE International Symposium on Robot and
  Human Interactive Communication (RO-MAN)}}. IEEE, \bibinfo{pages}{541--548}.
\newblock


\bibitem[\protect\citeauthoryear{Chernova and Thomaz}{Chernova and
  Thomaz}{2014}]%
        {chernova2014robot}
\bibfield{author}{\bibinfo{person}{Sonia Chernova} {and}
  \bibinfo{person}{Andrea~L Thomaz}.} \bibinfo{year}{2014}\natexlab{}.
\newblock \showarticletitle{Robot learning from human teachers}.
\newblock \bibinfo{journal}{\emph{Synthesis Lectures on Artificial Intelligence
  and Machine Learning}} \bibinfo{volume}{8}, \bibinfo{number}{3}
  (\bibinfo{year}{2014}), \bibinfo{pages}{1--121}.
\newblock


\bibitem[\protect\citeauthoryear{Christensen, Batzinger, Bekris, Bohringer,
  Bordogna, Bradski, Brock, Burnstein, Fuhlbrigge, Eastman,
  et~al\mbox{.}}{Christensen et~al\mbox{.}}{2009}]%
        {christensen2009roadmap}
\bibfield{author}{\bibinfo{person}{Henrik~I Christensen}, \bibinfo{person}{T
  Batzinger}, \bibinfo{person}{K Bekris}, \bibinfo{person}{K Bohringer},
  \bibinfo{person}{J Bordogna}, \bibinfo{person}{G Bradski}, \bibinfo{person}{O
  Brock}, \bibinfo{person}{J Burnstein}, \bibinfo{person}{T Fuhlbrigge},
  \bibinfo{person}{R Eastman}, {et~al\mbox{.}}}
  \bibinfo{year}{2009}\natexlab{}.
\newblock \showarticletitle{A roadmap for us robotics: from internet to
  robotics}.
\newblock \bibinfo{journal}{\emph{Computing Community Consortium}}
  \bibinfo{volume}{44} (\bibinfo{year}{2009}).
\newblock


\bibitem[\protect\citeauthoryear{Coovert, Lee, Shindev, and Sun}{Coovert
  et~al\mbox{.}}{2014}]%
        {coovert2014spatial}
\bibfield{author}{\bibinfo{person}{Michael~D Coovert}, \bibinfo{person}{Tiffany
  Lee}, \bibinfo{person}{Ivan Shindev}, {and} \bibinfo{person}{Yu Sun}.}
  \bibinfo{year}{2014}\natexlab{}.
\newblock \showarticletitle{Spatial augmented reality as a method for a mobile
  robot to communicate intended movement}.
\newblock \bibinfo{journal}{\emph{Computers in Human Behavior}}
  \bibinfo{volume}{34} (\bibinfo{year}{2014}), \bibinfo{pages}{241--248}.
\newblock


\bibitem[\protect\citeauthoryear{Coronado, Mastrogiovanni, Indurkhya, and
  Venture}{Coronado et~al\mbox{.}}{2020}]%
        {coronado2020visual}
\bibfield{author}{\bibinfo{person}{Enrique Coronado}, \bibinfo{person}{Fulvio
  Mastrogiovanni}, \bibinfo{person}{Bipin Indurkhya}, {and}
  \bibinfo{person}{Gentiane Venture}.} \bibinfo{year}{2020}\natexlab{}.
\newblock \showarticletitle{Visual Programming Environments for End-User
  Development of Intelligent and Social Robots, a Systematic Review}.
\newblock \bibinfo{journal}{\emph{Journal of Computer Languages}}
  (\bibinfo{year}{2020}), \bibinfo{pages}{100970}.
\newblock


\bibitem[\protect\citeauthoryear{Cypher and Smith}{Cypher and Smith}{1995}]%
        {cypher1995kidsim}
\bibfield{author}{\bibinfo{person}{Allen Cypher} {and}
  \bibinfo{person}{David~Canfield Smith}.} \bibinfo{year}{1995}\natexlab{}.
\newblock \showarticletitle{KidSim: end user programming of simulations}. In
  \bibinfo{booktitle}{\emph{Proceedings of the SIGCHI conference on Human
  factors in computing systems}}. \bibinfo{pages}{27--34}.
\newblock


\bibitem[\protect\citeauthoryear{Datta, Yang, Tiwari, Kuo, and MacDonald}{Datta
  et~al\mbox{.}}{2011}]%
        {datta2011end}
\bibfield{author}{\bibinfo{person}{Chandan Datta}, \bibinfo{person}{H~Yul
  Yang}, \bibinfo{person}{Priyesh Tiwari}, \bibinfo{person}{I~Han Kuo}, {and}
  \bibinfo{person}{Bruce~A MacDonald}.} \bibinfo{year}{2011}\natexlab{}.
\newblock \showarticletitle{End user programming to enable closed-loop
  medication management using a healthcare robot}.
\newblock \bibinfo{journal}{\emph{Social Science}} (\bibinfo{year}{2011}).
\newblock


\bibitem[\protect\citeauthoryear{Diprose, MacDonald, Hosking, and
  Plimmer}{Diprose et~al\mbox{.}}{2017}]%
        {diprose2017designing}
\bibfield{author}{\bibinfo{person}{James Diprose}, \bibinfo{person}{Bruce
  MacDonald}, \bibinfo{person}{John Hosking}, {and} \bibinfo{person}{Beryl
  Plimmer}.} \bibinfo{year}{2017}\natexlab{}.
\newblock \showarticletitle{Designing an API at an appropriate abstraction
  level for programming social robot applications}.
\newblock \bibinfo{journal}{\emph{Journal of Visual Languages \& Computing}}
  \bibinfo{volume}{39} (\bibinfo{year}{2017}), \bibinfo{pages}{22--40}.
\newblock


\bibitem[\protect\citeauthoryear{Dudek, Jenkin, Milios, and Wilkes}{Dudek
  et~al\mbox{.}}{1996}]%
        {dudek1996taxonomy}
\bibfield{author}{\bibinfo{person}{Gregory Dudek}, \bibinfo{person}{Michael~RM
  Jenkin}, \bibinfo{person}{Evangelos Milios}, {and} \bibinfo{person}{David
  Wilkes}.} \bibinfo{year}{1996}\natexlab{}.
\newblock \showarticletitle{A taxonomy for multi-agent robotics}.
\newblock \bibinfo{journal}{\emph{Autonomous Robots}} \bibinfo{volume}{3},
  \bibinfo{number}{4} (\bibinfo{year}{1996}), \bibinfo{pages}{375--397}.
\newblock


\bibitem[\protect\citeauthoryear{Erich, Hirokawa, and Suzuki}{Erich
  et~al\mbox{.}}{2017}]%
        {erich2017visual}
\bibfield{author}{\bibinfo{person}{Floris Erich}, \bibinfo{person}{Masakazu
  Hirokawa}, {and} \bibinfo{person}{Kenji Suzuki}.}
  \bibinfo{year}{2017}\natexlab{}.
\newblock \showarticletitle{A Visual Environment for Reactive Robot Programming
  of Macro-level Behaviors}. In \bibinfo{booktitle}{\emph{International
  Conference on Social Robotics}}. Springer, \bibinfo{pages}{577--586}.
\newblock


\bibitem[\protect\citeauthoryear{Fang, Ong, and Nee}{Fang
  et~al\mbox{.}}{2014}]%
        {fang2014novel}
\bibfield{author}{\bibinfo{person}{HC Fang}, \bibinfo{person}{SK Ong}, {and}
  \bibinfo{person}{AYC Nee}.} \bibinfo{year}{2014}\natexlab{}.
\newblock \showarticletitle{A novel augmented reality-based interface for robot
  path planning}.
\newblock \bibinfo{journal}{\emph{International Journal on Interactive Design
  and Manufacturing (IJIDeM)}} \bibinfo{volume}{8}, \bibinfo{number}{1}
  (\bibinfo{year}{2014}), \bibinfo{pages}{33--42}.
\newblock


\bibitem[\protect\citeauthoryear{Fernaeus and Jacobsson}{Fernaeus and
  Jacobsson}{2009}]%
        {fernaeus2009comics}
\bibfield{author}{\bibinfo{person}{Ylva Fernaeus} {and}
  \bibinfo{person}{Mattias Jacobsson}.} \bibinfo{year}{2009}\natexlab{}.
\newblock \showarticletitle{Comics, Robots, Fashion and Programming: outlining
  the concept of actDresses}. In \bibinfo{booktitle}{\emph{Proceedings of the
  3rd International Conference on Tangible and Embedded Interaction}}.
  \bibinfo{pages}{3--8}.
\newblock


\bibitem[\protect\citeauthoryear{Forbes, Chung, Cakmak, and Rao}{Forbes
  et~al\mbox{.}}{2014}]%
        {forbes2014robot}
\bibfield{author}{\bibinfo{person}{Maxwell Forbes}, \bibinfo{person}{Michael
  Jae-Yoon Chung}, \bibinfo{person}{Maya Cakmak}, {and}
  \bibinfo{person}{Rajesh~PN Rao}.} \bibinfo{year}{2014}\natexlab{}.
\newblock \showarticletitle{Robot Programming by Demonstration with
  Crowdsourced Action Fixes.}. In \bibinfo{booktitle}{\emph{HCOMP}}.
\newblock


\bibitem[\protect\citeauthoryear{Franklin, Hill, Dwyer, Hansen, Iveland, and
  Harlow}{Franklin et~al\mbox{.}}{2016}]%
        {franklin2016initialization}
\bibfield{author}{\bibinfo{person}{Diana Franklin}, \bibinfo{person}{Charlotte
  Hill}, \bibinfo{person}{Hilary~A Dwyer}, \bibinfo{person}{Alexandria~K
  Hansen}, \bibinfo{person}{Ashley Iveland}, {and} \bibinfo{person}{Danielle~B
  Harlow}.} \bibinfo{year}{2016}\natexlab{}.
\newblock \showarticletitle{Initialization in scratch: Seeking knowledge
  transfer}. In \bibinfo{booktitle}{\emph{Proceedings of the 47th ACM Technical
  Symposium on Computing Science Education}}. \bibinfo{pages}{217--222}.
\newblock


\bibitem[\protect\citeauthoryear{Fussell, Kiesler, Setlock, and Yew}{Fussell
  et~al\mbox{.}}{2008}]%
        {fussell2008people}
\bibfield{author}{\bibinfo{person}{Susan~R Fussell}, \bibinfo{person}{Sara
  Kiesler}, \bibinfo{person}{Leslie~D Setlock}, {and} \bibinfo{person}{Victoria
  Yew}.} \bibinfo{year}{2008}\natexlab{}.
\newblock \showarticletitle{How people anthropomorphize robots}. In
  \bibinfo{booktitle}{\emph{2008 3rd ACM/IEEE International Conference on
  Human-Robot Interaction (HRI)}}. IEEE, \bibinfo{pages}{145--152}.
\newblock


\bibitem[\protect\citeauthoryear{Gadre, Rosen, Chien, Phillips, Tellex, and
  Konidaris}{Gadre et~al\mbox{.}}{2019}]%
        {gadre2019end}
\bibfield{author}{\bibinfo{person}{Samir~Yitzhak Gadre}, \bibinfo{person}{Eric
  Rosen}, \bibinfo{person}{Gary Chien}, \bibinfo{person}{Elizabeth Phillips},
  \bibinfo{person}{Stefanie Tellex}, {and} \bibinfo{person}{George Konidaris}.}
  \bibinfo{year}{2019}\natexlab{}.
\newblock \showarticletitle{End-user robot programming using mixed reality}. In
  \bibinfo{booktitle}{\emph{2019 International Conference on Robotics and
  Automation (ICRA)}}. IEEE, \bibinfo{pages}{2707--2713}.
\newblock


\bibitem[\protect\citeauthoryear{Gao and Huang}{Gao and Huang}{2019}]%
        {gao2019pati}
\bibfield{author}{\bibinfo{person}{Yuxiang Gao} {and}
  \bibinfo{person}{Chien-Ming Huang}.} \bibinfo{year}{2019}\natexlab{}.
\newblock \showarticletitle{PATI: a projection-based augmented table-top
  interface for robot programming}. In \bibinfo{booktitle}{\emph{Proceedings of
  the 24th International Conference on Intelligent User Interfaces}}.
  \bibinfo{pages}{345--355}.
\newblock


\bibitem[\protect\citeauthoryear{Gattullo, Scurati, Fiorentino, Uva, Ferrise,
  and Bordegoni}{Gattullo et~al\mbox{.}}{2019}]%
        {gattullo2019towards}
\bibfield{author}{\bibinfo{person}{Michele Gattullo},
  \bibinfo{person}{Giulia~Wally Scurati}, \bibinfo{person}{Michele Fiorentino},
  \bibinfo{person}{Antonio~Emmanuele Uva}, \bibinfo{person}{Francesco Ferrise},
  {and} \bibinfo{person}{Monica Bordegoni}.} \bibinfo{year}{2019}\natexlab{}.
\newblock \showarticletitle{Towards augmented reality manuals for industry 4.0:
  A methodology}.
\newblock \bibinfo{journal}{\emph{Robotics and Computer-Integrated
  Manufacturing}}  \bibinfo{volume}{56} (\bibinfo{year}{2019}),
  \bibinfo{pages}{276--286}.
\newblock


\bibitem[\protect\citeauthoryear{Gorostiza and Salichs}{Gorostiza and
  Salichs}{2010}]%
        {gorostiza2010natural}
\bibfield{author}{\bibinfo{person}{Javier~F Gorostiza} {and}
  \bibinfo{person}{Miguel~Angel Salichs}.} \bibinfo{year}{2010}\natexlab{}.
\newblock \showarticletitle{Natural Programming of a Social Robot by Dialogs.}.
  In \bibinfo{booktitle}{\emph{AAAI Fall Symposium: Dialog with Robots}}.
\newblock


\bibitem[\protect\citeauthoryear{Gorostiza and Salichs}{Gorostiza and
  Salichs}{2011}]%
        {gorostiza2011end}
\bibfield{author}{\bibinfo{person}{Javi~F Gorostiza} {and}
  \bibinfo{person}{Miguel~A Salichs}.} \bibinfo{year}{2011}\natexlab{}.
\newblock \showarticletitle{End-user programming of a social robot by dialog}.
\newblock \bibinfo{journal}{\emph{Robotics and Autonomous Systems}}
  \bibinfo{volume}{59}, \bibinfo{number}{12} (\bibinfo{year}{2011}),
  \bibinfo{pages}{1102--1114}.
\newblock


\bibitem[\protect\citeauthoryear{Green and Blackwell}{Green and
  Blackwell}{1998}]%
        {green1998cognitive}
\bibfield{author}{\bibinfo{person}{Thomas Green} {and} \bibinfo{person}{Alan
  Blackwell}.} \bibinfo{year}{1998}\natexlab{}.
\newblock \showarticletitle{Cognitive dimensions of information artefacts: a
  tutorial}. In \bibinfo{booktitle}{\emph{BCS HCI Conference}},
  Vol.~\bibinfo{volume}{98}. \bibinfo{pages}{1--75}.
\newblock


\bibitem[\protect\citeauthoryear{Green and Petre}{Green and Petre}{1996}]%
        {green1996usability}
\bibfield{author}{\bibinfo{person}{Thomas R.~G. Green} {and}
  \bibinfo{person}{Marian Petre}.} \bibinfo{year}{1996}\natexlab{}.
\newblock \showarticletitle{Usability analysis of visual programming
  environments: a ‘cognitive dimensions’ framework}.
\newblock \bibinfo{journal}{\emph{Journal of Visual Languages \& Computing}}
  \bibinfo{volume}{7}, \bibinfo{number}{2} (\bibinfo{year}{1996}),
  \bibinfo{pages}{131--174}.
\newblock


\bibitem[\protect\citeauthoryear{Guerin, Lea, Paxton, and Hager}{Guerin
  et~al\mbox{.}}{2015}]%
        {guerin2015framework}
\bibfield{author}{\bibinfo{person}{Kelleher~R Guerin}, \bibinfo{person}{Colin
  Lea}, \bibinfo{person}{Chris Paxton}, {and} \bibinfo{person}{Gregory~D
  Hager}.} \bibinfo{year}{2015}\natexlab{}.
\newblock \showarticletitle{A framework for end-user instruction of a robot
  assistant for manufacturing}. In \bibinfo{booktitle}{\emph{2015 IEEE
  international conference on robotics and automation (ICRA)}}. IEEE,
  \bibinfo{pages}{6167--6174}.
\newblock


\bibitem[\protect\citeauthoryear{Han, Ajaykumar, Li, and Huang}{Han
  et~al\mbox{.}}{2020}]%
        {han2020structuring}
\bibfield{author}{\bibinfo{person}{Ji Han}, \bibinfo{person}{Gopika Ajaykumar},
  \bibinfo{person}{Ze Li}, {and} \bibinfo{person}{Chien-Ming Huang}.}
  \bibinfo{year}{2020}\natexlab{}.
\newblock \showarticletitle{Structuring Human-Robot Interactions via
  Interaction Conventions}. In \bibinfo{booktitle}{\emph{2020 29th IEEE
  International Conference on Robot and Human Interactive Communication
  (RO-MAN)}}. IEEE, \bibinfo{pages}{341--348}.
\newblock


\bibitem[\protect\citeauthoryear{Harrison}{Harrison}{2004}]%
        {harrison2004editor}
\bibfield{author}{\bibinfo{person}{Warren Harrison}.}
  \bibinfo{year}{2004}\natexlab{}.
\newblock \showarticletitle{From the editor: The dangers of end-user
  programming}.
\newblock \bibinfo{journal}{\emph{IEEE software}} \bibinfo{volume}{21},
  \bibinfo{number}{4} (\bibinfo{year}{2004}), \bibinfo{pages}{5--7}.
\newblock


\bibitem[\protect\citeauthoryear{Hart and Staveland}{Hart and
  Staveland}{1988}]%
        {hart1988development}
\bibfield{author}{\bibinfo{person}{Sandra~G Hart} {and}
  \bibinfo{person}{Lowell~E Staveland}.} \bibinfo{year}{1988}\natexlab{}.
\newblock \showarticletitle{Development of NASA-TLX (Task Load Index): Results
  of empirical and theoretical research}.
\newblock In \bibinfo{booktitle}{\emph{Advances in psychology}}.
  Vol.~\bibinfo{volume}{52}. \bibinfo{publisher}{Elsevier},
  \bibinfo{pages}{139--183}.
\newblock


\bibitem[\protect\citeauthoryear{Hedayati, Walker, and Szafir}{Hedayati
  et~al\mbox{.}}{2018}]%
        {hedayati2018improving}
\bibfield{author}{\bibinfo{person}{Hooman Hedayati}, \bibinfo{person}{Michael
  Walker}, {and} \bibinfo{person}{Daniel Szafir}.}
  \bibinfo{year}{2018}\natexlab{}.
\newblock \showarticletitle{Improving collocated robot teleoperation with
  augmented reality}. In \bibinfo{booktitle}{\emph{Proceedings of the 2018
  ACM/IEEE International Conference on Human-Robot Interaction}}.
  \bibinfo{pages}{78--86}.
\newblock


\bibitem[\protect\citeauthoryear{Huang and Mutlu}{Huang and Mutlu}{2012}]%
        {huang2012robot}
\bibfield{author}{\bibinfo{person}{Chien-Ming Huang} {and}
  \bibinfo{person}{Bilge Mutlu}.} \bibinfo{year}{2012}\natexlab{}.
\newblock \showarticletitle{Robot behavior toolkit: generating effective social
  behaviors for robots}. In \bibinfo{booktitle}{\emph{2012 7th ACM/IEEE
  International Conference on Human-Robot Interaction (HRI)}}. IEEE,
  \bibinfo{pages}{25--32}.
\newblock


\bibitem[\protect\citeauthoryear{Huang, Rao, Wu, Qian, Nof, Ramani, and
  Quinn}{Huang et~al\mbox{.}}{2020}]%
        {huang2020vipo}
\bibfield{author}{\bibinfo{person}{Gaoping Huang}, \bibinfo{person}{Pawan~S
  Rao}, \bibinfo{person}{Meng-Han Wu}, \bibinfo{person}{Xun Qian},
  \bibinfo{person}{Shimon~Y Nof}, \bibinfo{person}{Karthik Ramani}, {and}
  \bibinfo{person}{Alexander~J Quinn}.} \bibinfo{year}{2020}\natexlab{}.
\newblock \showarticletitle{Vipo: Spatial-Visual Programming with Functions for
  Robot-IoT Workflows}. In \bibinfo{booktitle}{\emph{Proceedings of the 2020
  CHI Conference on Human Factors in Computing Systems}}.
  \bibinfo{pages}{1--13}.
\newblock


\bibitem[\protect\citeauthoryear{Huang and Cakmak}{Huang and Cakmak}{2017}]%
        {huang2017code3}
\bibfield{author}{\bibinfo{person}{Justin Huang} {and} \bibinfo{person}{Maya
  Cakmak}.} \bibinfo{year}{2017}\natexlab{}.
\newblock \showarticletitle{Code3: A system for end-to-end programming of
  mobile manipulator robots for novices and experts}. In
  \bibinfo{booktitle}{\emph{2017 12th ACM/IEEE International Conference on
  Human-Robot Interaction (HRI}}. IEEE, \bibinfo{pages}{453--462}.
\newblock


\bibitem[\protect\citeauthoryear{Huang, Lau, and Cakmak}{Huang
  et~al\mbox{.}}{2016}]%
        {huang2016design}
\bibfield{author}{\bibinfo{person}{Justin Huang}, \bibinfo{person}{Tessa Lau},
  {and} \bibinfo{person}{Maya Cakmak}.} \bibinfo{year}{2016}\natexlab{}.
\newblock \showarticletitle{Design and evaluation of a rapid programming system
  for service robots}. In \bibinfo{booktitle}{\emph{2016 11th ACM/IEEE
  International Conference on Human-Robot Interaction (HRI)}}. IEEE,
  \bibinfo{pages}{295--302}.
\newblock


\bibitem[\protect\citeauthoryear{Hussein, Gaber, Elyan, and Jayne}{Hussein
  et~al\mbox{.}}{2017}]%
        {hussein2017imitation}
\bibfield{author}{\bibinfo{person}{Ahmed Hussein},
  \bibinfo{person}{Mohamed~Medhat Gaber}, \bibinfo{person}{Eyad Elyan}, {and}
  \bibinfo{person}{Chrisina Jayne}.} \bibinfo{year}{2017}\natexlab{}.
\newblock \showarticletitle{Imitation learning: A survey of learning methods}.
\newblock \bibinfo{journal}{\emph{ACM Computing Surveys (CSUR)}}
  \bibinfo{volume}{50}, \bibinfo{number}{2} (\bibinfo{year}{2017}),
  \bibinfo{pages}{1--35}.
\newblock


\bibitem[\protect\citeauthoryear{Jha, Chiddarwar, and Andulkar}{Jha
  et~al\mbox{.}}{2015}]%
        {jha2015application}
\bibfield{author}{\bibinfo{person}{Abhishek Jha}, \bibinfo{person}{Shital~S
  Chiddarwar}, {and} \bibinfo{person}{Mayur Andulkar}.}
  \bibinfo{year}{2015}\natexlab{}.
\newblock \showarticletitle{Application of human arm kinematics for robot path
  programming using imitation}. In \bibinfo{booktitle}{\emph{Proceedings of the
  2015 Conference on Advances In Robotics}}. \bibinfo{pages}{1--6}.
\newblock


\bibitem[\protect\citeauthoryear{Kapinus, Beran, Materna, and
  Bambu{\v{s}}ek}{Kapinus et~al\mbox{.}}{2019}]%
        {kapinus2019spatially}
\bibfield{author}{\bibinfo{person}{Michal Kapinus},
  \bibinfo{person}{V{\'\i}t{\v{e}}zslav Beran}, \bibinfo{person}{Zden{\v{e}}k
  Materna}, {and} \bibinfo{person}{Daniel Bambu{\v{s}}ek}.}
  \bibinfo{year}{2019}\natexlab{}.
\newblock \showarticletitle{Spatially Situated End-User Robot Programming in
  Augmented Reality}. In \bibinfo{booktitle}{\emph{2019 28th IEEE International
  Conference on Robot and Human Interactive Communication (RO-MAN)}}. IEEE,
  \bibinfo{pages}{1--8}.
\newblock


\bibitem[\protect\citeauthoryear{Kieras and Bovair}{Kieras and Bovair}{1984}]%
        {kieras1984role}
\bibfield{author}{\bibinfo{person}{David~E Kieras} {and} \bibinfo{person}{Susan
  Bovair}.} \bibinfo{year}{1984}\natexlab{}.
\newblock \showarticletitle{The role of a mental model in learning to operate a
  device}.
\newblock \bibinfo{journal}{\emph{Cognitive science}} \bibinfo{volume}{8},
  \bibinfo{number}{3} (\bibinfo{year}{1984}), \bibinfo{pages}{255--273}.
\newblock


\bibitem[\protect\citeauthoryear{Ko, Abraham, Beckwith, Blackwell, Burnett,
  Erwig, Scaffidi, Lawrance, Lieberman, Myers, et~al\mbox{.}}{Ko
  et~al\mbox{.}}{2011}]%
        {ko2011state}
\bibfield{author}{\bibinfo{person}{Andrew~J Ko}, \bibinfo{person}{Robin
  Abraham}, \bibinfo{person}{Laura Beckwith}, \bibinfo{person}{Alan Blackwell},
  \bibinfo{person}{Margaret Burnett}, \bibinfo{person}{Martin Erwig},
  \bibinfo{person}{Chris Scaffidi}, \bibinfo{person}{Joseph Lawrance},
  \bibinfo{person}{Henry Lieberman}, \bibinfo{person}{Brad Myers},
  {et~al\mbox{.}}} \bibinfo{year}{2011}\natexlab{}.
\newblock \showarticletitle{The state of the art in end-user software
  engineering}.
\newblock \bibinfo{journal}{\emph{ACM Computing Surveys (CSUR)}}
  \bibinfo{volume}{43}, \bibinfo{number}{3} (\bibinfo{year}{2011}),
  \bibinfo{pages}{1--44}.
\newblock


\bibitem[\protect\citeauthoryear{Koay, Syrdal, Walters, and Dautenhahn}{Koay
  et~al\mbox{.}}{2007}]%
        {koay2007living}
\bibfield{author}{\bibinfo{person}{Kheng~Lee Koay}, \bibinfo{person}{Dag~Sverre
  Syrdal}, \bibinfo{person}{Michael~L Walters}, {and} \bibinfo{person}{Kerstin
  Dautenhahn}.} \bibinfo{year}{2007}\natexlab{}.
\newblock \showarticletitle{Living with robots: Investigating the habituation
  effect in participants' preferences during a longitudinal human-robot
  interaction study}. In \bibinfo{booktitle}{\emph{RO-MAN 2007-The 16th IEEE
  International Symposium on Robot and Human Interactive Communication}}. IEEE,
  \bibinfo{pages}{564--569}.
\newblock


\bibitem[\protect\citeauthoryear{Kruijff, Swan, and Feiner}{Kruijff
  et~al\mbox{.}}{2010}]%
        {kruijff2010perceptual}
\bibfield{author}{\bibinfo{person}{Ernst Kruijff}, \bibinfo{person}{J~Edward
  Swan}, {and} \bibinfo{person}{Steven Feiner}.}
  \bibinfo{year}{2010}\natexlab{}.
\newblock \showarticletitle{Perceptual issues in augmented reality revisited}.
  In \bibinfo{booktitle}{\emph{2010 IEEE International Symposium on Mixed and
  Augmented Reality}}. IEEE, \bibinfo{pages}{3--12}.
\newblock


\bibitem[\protect\citeauthoryear{Kubota, Peterson, Rajendren, Kress-Gazit, and
  Riek}{Kubota et~al\mbox{.}}{2020}]%
        {kubota2020jessie}
\bibfield{author}{\bibinfo{person}{Alyssa Kubota}, \bibinfo{person}{Emma~IC
  Peterson}, \bibinfo{person}{Vaishali Rajendren}, \bibinfo{person}{Hadas
  Kress-Gazit}, {and} \bibinfo{person}{Laurel~D Riek}.}
  \bibinfo{year}{2020}\natexlab{}.
\newblock \showarticletitle{JESSIE: Synthesizing Social Robot Behaviors for
  Personalized Neurorehabilitation and Beyond}. In
  \bibinfo{booktitle}{\emph{Proceedings of the 2020 ACM/IEEE International
  Conference on Human-Robot Interaction}}. \bibinfo{pages}{121--130}.
\newblock


\bibitem[\protect\citeauthoryear{Laugwitz, Held, and Schrepp}{Laugwitz
  et~al\mbox{.}}{2008}]%
        {laugwitz2008construction}
\bibfield{author}{\bibinfo{person}{Bettina Laugwitz}, \bibinfo{person}{Theo
  Held}, {and} \bibinfo{person}{Martin Schrepp}.}
  \bibinfo{year}{2008}\natexlab{}.
\newblock \showarticletitle{Construction and evaluation of a user experience
  questionnaire}. In \bibinfo{booktitle}{\emph{Symposium of the Austrian HCI
  and usability engineering group}}. Springer, \bibinfo{pages}{63--76}.
\newblock


\bibitem[\protect\citeauthoryear{Lee, Park, and Song}{Lee
  et~al\mbox{.}}{2005}]%
        {lee2005can}
\bibfield{author}{\bibinfo{person}{Kwan~Min Lee}, \bibinfo{person}{Namkee
  Park}, {and} \bibinfo{person}{Hayeon Song}.} \bibinfo{year}{2005}\natexlab{}.
\newblock \showarticletitle{Can a Robot Be Perceived as a Developing Creature?
  Effects of a Robot's Long-Term Cognitive Developments on Its Social Presence
  and People's Social Responses Toward It}.
\newblock \bibinfo{journal}{\emph{Human communication research}}
  \bibinfo{volume}{31}, \bibinfo{number}{4} (\bibinfo{year}{2005}),
  \bibinfo{pages}{538--563}.
\newblock


\bibitem[\protect\citeauthoryear{Leite, Martinho, and Paiva}{Leite
  et~al\mbox{.}}{2013}]%
        {leite2013social}
\bibfield{author}{\bibinfo{person}{Iolanda Leite}, \bibinfo{person}{Carlos
  Martinho}, {and} \bibinfo{person}{Ana Paiva}.}
  \bibinfo{year}{2013}\natexlab{}.
\newblock \showarticletitle{Social robots for long-term interaction: a survey}.
\newblock \bibinfo{journal}{\emph{International Journal of Social Robotics}}
  \bibinfo{volume}{5}, \bibinfo{number}{2} (\bibinfo{year}{2013}),
  \bibinfo{pages}{291--308}.
\newblock


\bibitem[\protect\citeauthoryear{Leonardi, Manca, Patern{\`o}, and
  Santoro}{Leonardi et~al\mbox{.}}{2019}]%
        {leonardi2019trigger}
\bibfield{author}{\bibinfo{person}{Nicola Leonardi}, \bibinfo{person}{Marco
  Manca}, \bibinfo{person}{Fabio Patern{\`o}}, {and} \bibinfo{person}{Carmen
  Santoro}.} \bibinfo{year}{2019}\natexlab{}.
\newblock \showarticletitle{Trigger-action programming for personalising
  humanoid robot behaviour}. In \bibinfo{booktitle}{\emph{Proceedings of the
  2019 CHI Conference on Human Factors in Computing Systems}}.
  \bibinfo{pages}{1--13}.
\newblock


\bibitem[\protect\citeauthoryear{Liang, Pellier, Fiorino, and Pesty}{Liang
  et~al\mbox{.}}{2017}]%
        {liang2017framework}
\bibfield{author}{\bibinfo{person}{Ying~Siu Liang}, \bibinfo{person}{Damien
  Pellier}, \bibinfo{person}{Humbert Fiorino}, {and} \bibinfo{person}{Sylvie
  Pesty}.} \bibinfo{year}{2017}\natexlab{}.
\newblock \showarticletitle{A framework for robot programming in cobotic
  environments: first user experiments}. In
  \bibinfo{booktitle}{\emph{Proceedings of the 3rd International Conference on
  Mechatronics and Robotics Engineering}}. \bibinfo{pages}{30--35}.
\newblock


\bibitem[\protect\citeauthoryear{Liang, Pellier, Fiorino, and Pesty}{Liang
  et~al\mbox{.}}{2019}]%
        {liang2019end}
\bibfield{author}{\bibinfo{person}{Ying~Siu Liang}, \bibinfo{person}{Damien
  Pellier}, \bibinfo{person}{Humbert Fiorino}, {and} \bibinfo{person}{Sylvie
  Pesty}.} \bibinfo{year}{2019}\natexlab{}.
\newblock \showarticletitle{End-User Programming of Low-and High-Level Actions
  for Robotic Task Planning}. In \bibinfo{booktitle}{\emph{2019 28th IEEE
  International Conference on Robot and Human Interactive Communication
  (RO-MAN)}}. IEEE, \bibinfo{pages}{1--8}.
\newblock


\bibitem[\protect\citeauthoryear{Liang, Pellier, Fiorino, Pesty, and
  Cakmak}{Liang et~al\mbox{.}}{2018}]%
        {liang2018simultaneous}
\bibfield{author}{\bibinfo{person}{Ying~Siu Liang}, \bibinfo{person}{Damien
  Pellier}, \bibinfo{person}{Humbert Fiorino}, \bibinfo{person}{Sylvie Pesty},
  {and} \bibinfo{person}{Maya Cakmak}.} \bibinfo{year}{2018}\natexlab{}.
\newblock \showarticletitle{Simultaneous end-user programming of goals and
  actions for robotic shelf organization}. In \bibinfo{booktitle}{\emph{2018
  IEEE/RSJ International Conference on Intelligent Robots and Systems (IROS)}}.
  IEEE, \bibinfo{pages}{6566--6573}.
\newblock


\bibitem[\protect\citeauthoryear{Liu, Li, Liu, and Kan}{Liu
  et~al\mbox{.}}{2020}]%
        {liu2020skill}
\bibfield{author}{\bibinfo{person}{Yueyue Liu}, \bibinfo{person}{Zhijun Li},
  \bibinfo{person}{Huaping Liu}, {and} \bibinfo{person}{Zhen Kan}.}
  \bibinfo{year}{2020}\natexlab{}.
\newblock \showarticletitle{Skill transfer learning for autonomous robots and
  human-robot cooperation: A survey}.
\newblock \bibinfo{journal}{\emph{Robotics and Autonomous Systems}}
  (\bibinfo{year}{2020}), \bibinfo{pages}{103515}.
\newblock


\bibitem[\protect\citeauthoryear{Lozano-Perez}{Lozano-Perez}{1983}]%
        {lozano1983robot}
\bibfield{author}{\bibinfo{person}{Tomas Lozano-Perez}.}
  \bibinfo{year}{1983}\natexlab{}.
\newblock \showarticletitle{Robot programming}.
\newblock \bibinfo{journal}{\emph{Proc. IEEE}} \bibinfo{volume}{71},
  \bibinfo{number}{7} (\bibinfo{year}{1983}), \bibinfo{pages}{821--841}.
\newblock


\bibitem[\protect\citeauthoryear{Maceli}{Maceli}{2017}]%
        {maceli2017tools}
\bibfield{author}{\bibinfo{person}{Monica~G Maceli}.}
  \bibinfo{year}{2017}\natexlab{}.
\newblock \showarticletitle{Tools of the trade: a survey of technologies in
  end-user development literature}. In \bibinfo{booktitle}{\emph{International
  symposium on end user development}}. Springer, \bibinfo{pages}{49--65}.
\newblock


\bibitem[\protect\citeauthoryear{Manohar and Crandall}{Manohar and
  Crandall}{2014}]%
        {manohar2014programming}
\bibfield{author}{\bibinfo{person}{Vimitha Manohar} {and}
  \bibinfo{person}{Jacob~W Crandall}.} \bibinfo{year}{2014}\natexlab{}.
\newblock \showarticletitle{Programming robots to express emotions: interaction
  paradigms, communication modalities, and context}.
\newblock \bibinfo{journal}{\emph{IEEE transactions on human-machine systems}}
  \bibinfo{volume}{44}, \bibinfo{number}{3} (\bibinfo{year}{2014}),
  \bibinfo{pages}{362--373}.
\newblock


\bibitem[\protect\citeauthoryear{Matthaiakis, Dimoulas, Athanasatos, Mparis,
  Dimitrakopoulos, Gkournelos, Papavasileiou, Fousekis, Papanastasiou,
  Michalos, et~al\mbox{.}}{Matthaiakis et~al\mbox{.}}{2017}]%
        {matthaiakis2017flexible}
\bibfield{author}{\bibinfo{person}{Stereos~Alexandros Matthaiakis},
  \bibinfo{person}{Konstantinos Dimoulas}, \bibinfo{person}{Athanasios
  Athanasatos}, \bibinfo{person}{Konstantinos Mparis}, \bibinfo{person}{George
  Dimitrakopoulos}, \bibinfo{person}{Christos Gkournelos},
  \bibinfo{person}{Apostolis Papavasileiou}, \bibinfo{person}{Nikos Fousekis},
  \bibinfo{person}{Stergios Papanastasiou}, \bibinfo{person}{George Michalos},
  {et~al\mbox{.}}} \bibinfo{year}{2017}\natexlab{}.
\newblock \showarticletitle{Flexible programming tool enabling synergy between
  human and robot}.
\newblock \bibinfo{journal}{\emph{Procedia Manufacturing}}
  \bibinfo{volume}{11} (\bibinfo{year}{2017}), \bibinfo{pages}{431--440}.
\newblock


\bibitem[\protect\citeauthoryear{McKenney}{McKenney}{2017}]%
        {mckenney2017parallel}
\bibfield{author}{\bibinfo{person}{Paul~E McKenney}.}
  \bibinfo{year}{2017}\natexlab{}.
\newblock \showarticletitle{Is parallel programming hard, and, if so, what can
  you do about it?(v2017. 01.02 a)}.
\newblock \bibinfo{journal}{\emph{arXiv preprint arXiv:1701.00854}}
  (\bibinfo{year}{2017}).
\newblock


\bibitem[\protect\citeauthoryear{Moros, Wood, Robins, Dautenhahn, and
  Castro-Gonz{\'a}lez}{Moros et~al\mbox{.}}{2019}]%
        {moros2019programming}
\bibfield{author}{\bibinfo{person}{S{\'\i}lvia Moros}, \bibinfo{person}{Luke
  Wood}, \bibinfo{person}{Ben Robins}, \bibinfo{person}{Kerstin Dautenhahn},
  {and} \bibinfo{person}{{\'A}lvaro Castro-Gonz{\'a}lez}.}
  \bibinfo{year}{2019}\natexlab{}.
\newblock \showarticletitle{Programming a humanoid robot with the scratch
  language}. In \bibinfo{booktitle}{\emph{International Conference on Robotics
  and Education RiE 2017}}. Springer, \bibinfo{pages}{222--233}.
\newblock


\bibitem[\protect\citeauthoryear{Neto and Mendes}{Neto and Mendes}{2013}]%
        {neto2013direct}
\bibfield{author}{\bibinfo{person}{Pedro Neto} {and} \bibinfo{person}{Nuno
  Mendes}.} \bibinfo{year}{2013}\natexlab{}.
\newblock \showarticletitle{Direct off-line robot programming via a common CAD
  package}.
\newblock \bibinfo{journal}{\emph{Robotics and Autonomous Systems}}
  \bibinfo{volume}{61}, \bibinfo{number}{8} (\bibinfo{year}{2013}),
  \bibinfo{pages}{896--910}.
\newblock


\bibitem[\protect\citeauthoryear{of~Robotics}{of~Robotics}{2020}]%
        {international2020world}
\bibfield{author}{\bibinfo{person}{International~Federation of Robotics}.}
  \bibinfo{year}{2020}\natexlab{}.
\newblock \bibinfo{title}{World Robotics Report 2020}.
\newblock
\newblock


\bibitem[\protect\citeauthoryear{Oishi, Kanda, Kanbara, Satake, and
  Hagita}{Oishi et~al\mbox{.}}{2017}]%
        {oishi2017toward}
\bibfield{author}{\bibinfo{person}{Yoha Oishi}, \bibinfo{person}{Takayuki
  Kanda}, \bibinfo{person}{Masayuki Kanbara}, \bibinfo{person}{Satoru Satake},
  {and} \bibinfo{person}{Norihiro Hagita}.} \bibinfo{year}{2017}\natexlab{}.
\newblock \showarticletitle{Toward end-user programming for robots in stores}.
  In \bibinfo{booktitle}{\emph{Proceedings of the Companion of the 2017
  ACM/IEEE International Conference on Human-Robot Interaction}}.
  \bibinfo{pages}{233--234}.
\newblock


\bibitem[\protect\citeauthoryear{Ong, Yew, Thanigaivel, and Nee}{Ong
  et~al\mbox{.}}{2020}]%
        {ong2020augmented}
\bibfield{author}{\bibinfo{person}{SK Ong}, \bibinfo{person}{AWW Yew},
  \bibinfo{person}{NK Thanigaivel}, {and} \bibinfo{person}{AYC Nee}.}
  \bibinfo{year}{2020}\natexlab{}.
\newblock \showarticletitle{Augmented reality-assisted robot programming system
  for industrial applications}.
\newblock \bibinfo{journal}{\emph{Robotics and Computer-Integrated
  Manufacturing}}  \bibinfo{volume}{61} (\bibinfo{year}{2020}),
  \bibinfo{pages}{101820}.
\newblock


\bibitem[\protect\citeauthoryear{Pan, Polden, Larkin, Van~Duin, and
  Norrish}{Pan et~al\mbox{.}}{2012}]%
        {pan2012recent}
\bibfield{author}{\bibinfo{person}{Zengxi Pan}, \bibinfo{person}{Joseph
  Polden}, \bibinfo{person}{Nathan Larkin}, \bibinfo{person}{Stephen Van~Duin},
  {and} \bibinfo{person}{John Norrish}.} \bibinfo{year}{2012}\natexlab{}.
\newblock \showarticletitle{Recent progress on programming methods for
  industrial robots}.
\newblock \bibinfo{journal}{\emph{Robotics and Computer-Integrated
  Manufacturing}} \bibinfo{volume}{28}, \bibinfo{number}{2}
  (\bibinfo{year}{2012}), \bibinfo{pages}{87--94}.
\newblock


\bibitem[\protect\citeauthoryear{Patern{\`o}}{Patern{\`o}}{2013}]%
        {paterno2013end}
\bibfield{author}{\bibinfo{person}{Fabio Patern{\`o}}.}
  \bibinfo{year}{2013}\natexlab{}.
\newblock \showarticletitle{End user development: Survey of an emerging field
  for empowering people}.
\newblock \bibinfo{journal}{\emph{ISRN Software Engineering}}
  \bibinfo{volume}{2013} (\bibinfo{year}{2013}).
\newblock


\bibitem[\protect\citeauthoryear{Patern{\`o} and Santoro}{Patern{\`o} and
  Santoro}{2019}]%
        {paterno2019end}
\bibfield{author}{\bibinfo{person}{Fabio Patern{\`o}} {and}
  \bibinfo{person}{Carmen Santoro}.} \bibinfo{year}{2019}\natexlab{}.
\newblock \showarticletitle{End-user development for personalizing
  applications, things, and robots}.
\newblock \bibinfo{journal}{\emph{International Journal of Human-Computer
  Studies}}  \bibinfo{volume}{131} (\bibinfo{year}{2019}),
  \bibinfo{pages}{120--130}.
\newblock


\bibitem[\protect\citeauthoryear{Paxton, Jonathan, Hundt, Mutlu, and
  Hager}{Paxton et~al\mbox{.}}{2018}]%
        {paxton2018evaluating}
\bibfield{author}{\bibinfo{person}{Chris Paxton}, \bibinfo{person}{Felix
  Jonathan}, \bibinfo{person}{Andrew Hundt}, \bibinfo{person}{Bilge Mutlu},
  {and} \bibinfo{person}{Gregory~D Hager}.} \bibinfo{year}{2018}\natexlab{}.
\newblock \showarticletitle{Evaluating methods for end-user creation of robot
  task plans}. In \bibinfo{booktitle}{\emph{2018 IEEE/RSJ International
  Conference on Intelligent Robots and Systems (IROS)}}. IEEE,
  \bibinfo{pages}{6086--6092}.
\newblock


\bibitem[\protect\citeauthoryear{Pedersen and Kr{\"u}ger}{Pedersen and
  Kr{\"u}ger}{2015}]%
        {pedersen2015gesture}
\bibfield{author}{\bibinfo{person}{Mikkel~Rath Pedersen} {and}
  \bibinfo{person}{Volker Kr{\"u}ger}.} \bibinfo{year}{2015}\natexlab{}.
\newblock \showarticletitle{Gesture-based extraction of robot skill parameters
  for intuitive robot programming}.
\newblock \bibinfo{journal}{\emph{Journal of Intelligent \& Robotic Systems}}
  \bibinfo{volume}{80}, \bibinfo{number}{1} (\bibinfo{year}{2015}),
  \bibinfo{pages}{149--163}.
\newblock


\bibitem[\protect\citeauthoryear{Porfirio, Fisher, Saupp{\'e}, Albarghouthi,
  and Mutlu}{Porfirio et~al\mbox{.}}{2019a}]%
        {porfirio2019bodystorming}
\bibfield{author}{\bibinfo{person}{David Porfirio}, \bibinfo{person}{Evan
  Fisher}, \bibinfo{person}{Allison Saupp{\'e}}, \bibinfo{person}{Aws
  Albarghouthi}, {and} \bibinfo{person}{Bilge Mutlu}.}
  \bibinfo{year}{2019}\natexlab{a}.
\newblock \showarticletitle{Bodystorming human-robot interactions}. In
  \bibinfo{booktitle}{\emph{Proceedings of the 32nd Annual ACM Symposium on
  User Interface Software and Technology}}. \bibinfo{pages}{479--491}.
\newblock


\bibitem[\protect\citeauthoryear{Porfirio, Saupp{\'e}, Albarghouthi, and
  Mutlu}{Porfirio et~al\mbox{.}}{2018}]%
        {porfirio2018authoring}
\bibfield{author}{\bibinfo{person}{David Porfirio}, \bibinfo{person}{Allison
  Saupp{\'e}}, \bibinfo{person}{Aws Albarghouthi}, {and} \bibinfo{person}{Bilge
  Mutlu}.} \bibinfo{year}{2018}\natexlab{}.
\newblock \showarticletitle{Authoring and verifying human-robot interactions}.
  In \bibinfo{booktitle}{\emph{Proceedings of the 31st Annual ACM Symposium on
  User Interface Software and Technology}}. \bibinfo{pages}{75--86}.
\newblock


\bibitem[\protect\citeauthoryear{Porfirio, Saupp{\'e}, Albarghouthi, and
  Mutlu}{Porfirio et~al\mbox{.}}{2019b}]%
        {porfirio2019computational}
\bibfield{author}{\bibinfo{person}{David Porfirio}, \bibinfo{person}{Allison
  Saupp{\'e}}, \bibinfo{person}{Aws Albarghouthi}, {and} \bibinfo{person}{Bilge
  Mutlu}.} \bibinfo{year}{2019}\natexlab{b}.
\newblock \showarticletitle{Computational tools for human-robot interaction
  design}. In \bibinfo{booktitle}{\emph{2019 14th ACM/IEEE International
  Conference on Human-Robot Interaction (HRI)}}. IEEE,
  \bibinfo{pages}{733--735}.
\newblock


\bibitem[\protect\citeauthoryear{Porfirio, Saupp{\'e}, Albarghouthi, and
  Mutlu}{Porfirio et~al\mbox{.}}{2020}]%
        {porfirio2020transforming}
\bibfield{author}{\bibinfo{person}{David Porfirio}, \bibinfo{person}{Allison
  Saupp{\'e}}, \bibinfo{person}{Aws Albarghouthi}, {and} \bibinfo{person}{Bilge
  Mutlu}.} \bibinfo{year}{2020}\natexlab{}.
\newblock \showarticletitle{Transforming Robot Programs Based on Social
  Context}. In \bibinfo{booktitle}{\emph{Proceedings of the 2020 CHI Conference
  on Human Factors in Computing Systems}}. \bibinfo{pages}{1--12}.
\newblock


\bibitem[\protect\citeauthoryear{Powers and Kiesler}{Powers and
  Kiesler}{2006}]%
        {powers2006advisor}
\bibfield{author}{\bibinfo{person}{Aaron Powers} {and} \bibinfo{person}{Sara
  Kiesler}.} \bibinfo{year}{2006}\natexlab{}.
\newblock \showarticletitle{The advisor robot: tracing people's mental model
  from a robot's physical attributes}. In \bibinfo{booktitle}{\emph{Proceedings
  of the 1st ACM SIGCHI/SIGART conference on Human-robot interaction}}.
  \bibinfo{pages}{218--225}.
\newblock


\bibitem[\protect\citeauthoryear{Prorok, Bahr, and Martinoli}{Prorok
  et~al\mbox{.}}{2012}]%
        {prorok2012low}
\bibfield{author}{\bibinfo{person}{Amanda Prorok}, \bibinfo{person}{Alexander
  Bahr}, {and} \bibinfo{person}{Alcherio Martinoli}.}
  \bibinfo{year}{2012}\natexlab{}.
\newblock \showarticletitle{Low-cost collaborative localization for large-scale
  multi-robot systems}. In \bibinfo{booktitle}{\emph{2012 IEEE International
  Conference on Robotics and Automation}}. Ieee, \bibinfo{pages}{4236--4241}.
\newblock


\bibitem[\protect\citeauthoryear{Quigley, Conley, Gerkey, Faust, Foote, Leibs,
  Wheeler, and Ng}{Quigley et~al\mbox{.}}{2009}]%
        {quigley2009ros}
\bibfield{author}{\bibinfo{person}{Morgan Quigley}, \bibinfo{person}{Ken
  Conley}, \bibinfo{person}{Brian Gerkey}, \bibinfo{person}{Josh Faust},
  \bibinfo{person}{Tully Foote}, \bibinfo{person}{Jeremy Leibs},
  \bibinfo{person}{Rob Wheeler}, {and} \bibinfo{person}{Andrew~Y Ng}.}
  \bibinfo{year}{2009}\natexlab{}.
\newblock \showarticletitle{ROS: an open-source Robot Operating System}. In
  \bibinfo{booktitle}{\emph{ICRA workshop on open source software}},
  Vol.~\bibinfo{volume}{3}. Kobe, Japan, \bibinfo{pages}{5}.
\newblock


\bibitem[\protect\citeauthoryear{Quintero, Li, Pan, Chan, Van~der Loos, and
  Croft}{Quintero et~al\mbox{.}}{2018}]%
        {quintero2018robot}
\bibfield{author}{\bibinfo{person}{Camilo~Perez Quintero},
  \bibinfo{person}{Sarah Li}, \bibinfo{person}{Matthew~KXJ Pan},
  \bibinfo{person}{Wesley~P Chan}, \bibinfo{person}{HF~Machiel Van~der Loos},
  {and} \bibinfo{person}{Elizabeth Croft}.} \bibinfo{year}{2018}\natexlab{}.
\newblock \showarticletitle{Robot programming through augmented trajectories in
  augmented reality}. In \bibinfo{booktitle}{\emph{2018 IEEE/RSJ International
  Conference on Intelligent Robots and Systems (IROS)}}. IEEE,
  \bibinfo{pages}{1838--1844}.
\newblock


\bibitem[\protect\citeauthoryear{Racca, Kyrki, and Cakmak}{Racca
  et~al\mbox{.}}{2020}]%
        {racca2020interactive}
\bibfield{author}{\bibinfo{person}{Mattia Racca}, \bibinfo{person}{Ville
  Kyrki}, {and} \bibinfo{person}{Maya Cakmak}.}
  \bibinfo{year}{2020}\natexlab{}.
\newblock \showarticletitle{Interactive Tuning of Robot Program Parameters via
  Expected Divergence Maximization}. In \bibinfo{booktitle}{\emph{Proceedings
  of the 2020 ACM/IEEE International Conference on Human-Robot Interaction}}.
  \bibinfo{pages}{629--638}.
\newblock


\bibitem[\protect\citeauthoryear{Ramo{\u{g}}lu, Gen{\c{c}}, and
  R{\i}zvano{\u{g}}lu}{Ramo{\u{g}}lu et~al\mbox{.}}{2017}]%
        {ramouglu2017programming}
\bibfield{author}{\bibinfo{person}{Muhammet Ramo{\u{g}}lu},
  \bibinfo{person}{{\c{C}}a{\u{g}}lar Gen{\c{c}}}, {and} \bibinfo{person}{Kerem
  R{\i}zvano{\u{g}}lu}.} \bibinfo{year}{2017}\natexlab{}.
\newblock \showarticletitle{Programming a robotic toy with a block coding
  application: A usability study with non-programmer adults}. In
  \bibinfo{booktitle}{\emph{International Conference of Design, User
  Experience, and Usability}}. Springer, \bibinfo{pages}{652--666}.
\newblock


\bibitem[\protect\citeauthoryear{Repenning, Ahmadi, Repenning, Ioannidou, Webb,
  and Marshall}{Repenning et~al\mbox{.}}{2011}]%
        {repenning2011collective}
\bibfield{author}{\bibinfo{person}{Alexander Repenning}, \bibinfo{person}{Navid
  Ahmadi}, \bibinfo{person}{Nadia Repenning}, \bibinfo{person}{Andri
  Ioannidou}, \bibinfo{person}{David Webb}, {and} \bibinfo{person}{Krista
  Marshall}.} \bibinfo{year}{2011}\natexlab{}.
\newblock \showarticletitle{Collective programming: Making end-user programming
  (more) social}. In \bibinfo{booktitle}{\emph{International Symposium on End
  User Development}}. Springer, \bibinfo{pages}{325--330}.
\newblock


\bibitem[\protect\citeauthoryear{Resnick, Maloney, Monroy-Hern{\'a}ndez, Rusk,
  Eastmond, Brennan, Millner, Rosenbaum, Silver, Silverman,
  et~al\mbox{.}}{Resnick et~al\mbox{.}}{2009}]%
        {resnick2009scratch}
\bibfield{author}{\bibinfo{person}{Mitchel Resnick}, \bibinfo{person}{John
  Maloney}, \bibinfo{person}{Andr{\'e}s Monroy-Hern{\'a}ndez},
  \bibinfo{person}{Natalie Rusk}, \bibinfo{person}{Evelyn Eastmond},
  \bibinfo{person}{Karen Brennan}, \bibinfo{person}{Amon Millner},
  \bibinfo{person}{Eric Rosenbaum}, \bibinfo{person}{Jay Silver},
  \bibinfo{person}{Brian Silverman}, {et~al\mbox{.}}}
  \bibinfo{year}{2009}\natexlab{}.
\newblock \showarticletitle{Scratch: programming for all}.
\newblock \bibinfo{journal}{\emph{Commun. ACM}} \bibinfo{volume}{52},
  \bibinfo{number}{11} (\bibinfo{year}{2009}), \bibinfo{pages}{60--67}.
\newblock


\bibitem[\protect\citeauthoryear{Riedl and Henrich}{Riedl and Henrich}{2019}]%
        {riedl2019fast}
\bibfield{author}{\bibinfo{person}{Michael Riedl} {and}
  \bibinfo{person}{Dominik Henrich}.} \bibinfo{year}{2019}\natexlab{}.
\newblock \showarticletitle{A Fast Robot Playback Programming System Using
  Video Editing Concepts}.
\newblock In \bibinfo{booktitle}{\emph{Tagungsband des 4. Kongresses Montage
  Handhabung Industrieroboter}}. \bibinfo{publisher}{Springer},
  \bibinfo{pages}{259--268}.
\newblock


\bibitem[\protect\citeauthoryear{Robertson, Prabhakararao, Burnett, Cook,
  Ruthruff, Beckwith, and Phalgune}{Robertson et~al\mbox{.}}{2004}]%
        {robertson2004impact}
\bibfield{author}{\bibinfo{person}{TJ Robertson}, \bibinfo{person}{Shrinu
  Prabhakararao}, \bibinfo{person}{Margaret Burnett}, \bibinfo{person}{Curtis
  Cook}, \bibinfo{person}{Joseph~R Ruthruff}, \bibinfo{person}{Laura Beckwith},
  {and} \bibinfo{person}{Amit Phalgune}.} \bibinfo{year}{2004}\natexlab{}.
\newblock \showarticletitle{Impact of interruption style on end-user
  debugging}. In \bibinfo{booktitle}{\emph{Proceedings of the SIGCHI conference
  on Human factors in computing systems}}. \bibinfo{pages}{287--294}.
\newblock


\bibitem[\protect\citeauthoryear{Ryokai, Lee, and Breitbart}{Ryokai
  et~al\mbox{.}}{2009}]%
        {ryokai2009children}
\bibfield{author}{\bibinfo{person}{Kimiko Ryokai},
  \bibinfo{person}{Michael~Jongseon Lee}, {and} \bibinfo{person}{Jonathan~Micah
  Breitbart}.} \bibinfo{year}{2009}\natexlab{}.
\newblock \showarticletitle{Children's storytelling and programming with
  robotic characters}. In \bibinfo{booktitle}{\emph{Proceedings of the seventh
  ACM conference on Creativity and cognition}}. \bibinfo{pages}{19--28}.
\newblock


\bibitem[\protect\citeauthoryear{Sandoval, Mubin, and Obaid}{Sandoval
  et~al\mbox{.}}{2014}]%
        {sandoval2014human}
\bibfield{author}{\bibinfo{person}{Eduardo~Benitez Sandoval},
  \bibinfo{person}{Omar Mubin}, {and} \bibinfo{person}{Mohammad Obaid}.}
  \bibinfo{year}{2014}\natexlab{}.
\newblock \showarticletitle{Human robot interaction and fiction: A
  contradiction}. In \bibinfo{booktitle}{\emph{International Conference on
  Social Robotics}}. Springer, \bibinfo{pages}{54--63}.
\newblock


\bibitem[\protect\citeauthoryear{Sapounidis and Demetriadis}{Sapounidis and
  Demetriadis}{2013}]%
        {sapounidis2013tangible}
\bibfield{author}{\bibinfo{person}{Theodosios Sapounidis} {and}
  \bibinfo{person}{Stavros Demetriadis}.} \bibinfo{year}{2013}\natexlab{}.
\newblock \showarticletitle{Tangible versus graphical user interfaces for robot
  programming: exploring cross-age children’s preferences}.
\newblock \bibinfo{journal}{\emph{Personal and ubiquitous computing}}
  \bibinfo{volume}{17}, \bibinfo{number}{8} (\bibinfo{year}{2013}),
  \bibinfo{pages}{1775--1786}.
\newblock


\bibitem[\protect\citeauthoryear{Scaffidi}{Scaffidi}{2016}]%
        {scaffidi2016potential}
\bibfield{author}{\bibinfo{person}{Chris Scaffidi}.}
  \bibinfo{year}{2016}\natexlab{}.
\newblock \showarticletitle{Potential financial motivations for end-user
  programming}. In \bibinfo{booktitle}{\emph{2016 IEEE Symposium on Visual
  Languages and Human-Centric Computing (VL/HCC)}}. IEEE,
  \bibinfo{pages}{180--184}.
\newblock


\bibitem[\protect\citeauthoryear{Scaffidi, Shaw, and Myers}{Scaffidi
  et~al\mbox{.}}{2005}]%
        {scaffidi2005estimating}
\bibfield{author}{\bibinfo{person}{Christopher Scaffidi}, \bibinfo{person}{Mary
  Shaw}, {and} \bibinfo{person}{Brad Myers}.} \bibinfo{year}{2005}\natexlab{}.
\newblock \showarticletitle{Estimating the numbers of end users and end user
  programmers}. In \bibinfo{booktitle}{\emph{2005 IEEE Symposium on Visual
  Languages and Human-Centric Computing (VL/HCC'05)}}. IEEE,
  \bibinfo{pages}{207--214}.
\newblock


\bibitem[\protect\citeauthoryear{Schaal, Ijspeert, and Billard}{Schaal
  et~al\mbox{.}}{2003}]%
        {schaal2003computational}
\bibfield{author}{\bibinfo{person}{Stefan Schaal}, \bibinfo{person}{Auke
  Ijspeert}, {and} \bibinfo{person}{Aude Billard}.}
  \bibinfo{year}{2003}\natexlab{}.
\newblock \showarticletitle{Computational approaches to motor learning by
  imitation}.
\newblock \bibinfo{journal}{\emph{Philosophical Transactions of the Royal
  Society of London. Series B: Biological Sciences}} \bibinfo{volume}{358},
  \bibinfo{number}{1431} (\bibinfo{year}{2003}), \bibinfo{pages}{537--547}.
\newblock


\bibitem[\protect\citeauthoryear{Schmidt, Boshuizen, and Breukelen}{Schmidt
  et~al\mbox{.}}{2002}]%
        {schmidt2002long}
\bibfield{author}{\bibinfo{person}{Henk~G Schmidt}, \bibinfo{person}{Henny~PA
  Boshuizen}, {and} \bibinfo{person}{Gerard JP~van Breukelen}.}
  \bibinfo{year}{2002}\natexlab{}.
\newblock \showarticletitle{Long-term retention of a theatrical script by
  repertory actors: The role of context}.
\newblock \bibinfo{journal}{\emph{Memory}} \bibinfo{volume}{10},
  \bibinfo{number}{1} (\bibinfo{year}{2002}), \bibinfo{pages}{21--28}.
\newblock


\bibitem[\protect\citeauthoryear{Schou, Andersen, Chrysostomou, B{\o}gh, and
  Madsen}{Schou et~al\mbox{.}}{2018}]%
        {schou2018skill}
\bibfield{author}{\bibinfo{person}{Casper Schou},
  \bibinfo{person}{Rasmus~Skovgaard Andersen}, \bibinfo{person}{Dimitrios
  Chrysostomou}, \bibinfo{person}{Simon B{\o}gh}, {and} \bibinfo{person}{Ole
  Madsen}.} \bibinfo{year}{2018}\natexlab{}.
\newblock \showarticletitle{Skill-based instruction of collaborative robots in
  industrial settings}.
\newblock \bibinfo{journal}{\emph{Robotics and Computer-Integrated
  Manufacturing}}  \bibinfo{volume}{53} (\bibinfo{year}{2018}),
  \bibinfo{pages}{72--80}.
\newblock


\bibitem[\protect\citeauthoryear{Sefidgar, Agarwal, and Cakmak}{Sefidgar
  et~al\mbox{.}}{2017}]%
        {sefidgar2017situated}
\bibfield{author}{\bibinfo{person}{Yasaman~S Sefidgar}, \bibinfo{person}{Prerna
  Agarwal}, {and} \bibinfo{person}{Maya Cakmak}.}
  \bibinfo{year}{2017}\natexlab{}.
\newblock \showarticletitle{Situated tangible robot programming}. In
  \bibinfo{booktitle}{\emph{2017 12th ACM/IEEE International Conference on
  Human-Robot Interaction (HRI}}. IEEE, \bibinfo{pages}{473--482}.
\newblock


\bibitem[\protect\citeauthoryear{Sefidgar, Weng, Harvey, Elliott, and
  Cakmak}{Sefidgar et~al\mbox{.}}{2018}]%
        {sefidgar2018robotist}
\bibfield{author}{\bibinfo{person}{Yasaman~S Sefidgar}, \bibinfo{person}{Thomas
  Weng}, \bibinfo{person}{Heather Harvey}, \bibinfo{person}{Sarah Elliott},
  {and} \bibinfo{person}{Maya Cakmak}.} \bibinfo{year}{2018}\natexlab{}.
\newblock \showarticletitle{RobotIST: Interactive Situated Tangible Robot
  Programming}. In \bibinfo{booktitle}{\emph{Proceedings of the Symposium on
  Spatial User Interaction}}. \bibinfo{pages}{141--149}.
\newblock


\bibitem[\protect\citeauthoryear{Spahn, Dorner, and Wulf}{Spahn
  et~al\mbox{.}}{2008}]%
        {spahn2008end}
\bibfield{author}{\bibinfo{person}{Michael Spahn}, \bibinfo{person}{Christian
  Dorner}, {and} \bibinfo{person}{Volker Wulf}.}
  \bibinfo{year}{2008}\natexlab{}.
\newblock \showarticletitle{End user development: approaches towards a flexible
  software design}.
\newblock  (\bibinfo{year}{2008}).
\newblock


\bibitem[\protect\citeauthoryear{Stenmark, Haage, and Topp}{Stenmark
  et~al\mbox{.}}{2017}]%
        {stenmark2017simplified}
\bibfield{author}{\bibinfo{person}{Maj Stenmark}, \bibinfo{person}{Mathias
  Haage}, {and} \bibinfo{person}{Elin~Anna Topp}.}
  \bibinfo{year}{2017}\natexlab{}.
\newblock \showarticletitle{Simplified programming of re-usable skills on a
  safe industrial robot: Prototype and evaluation}. In
  \bibinfo{booktitle}{\emph{Proceedings of the 2017 ACM/IEEE International
  Conference on Human-Robot Interaction}}. \bibinfo{pages}{463--472}.
\newblock


\bibitem[\protect\citeauthoryear{Strauss and Corbin}{Strauss and
  Corbin}{1994}]%
        {strauss1994grounded}
\bibfield{author}{\bibinfo{person}{Anselm Strauss} {and}
  \bibinfo{person}{Juliet Corbin}.} \bibinfo{year}{1994}\natexlab{}.
\newblock \showarticletitle{Grounded theory methodology}.
\newblock \bibinfo{journal}{\emph{Handbook of qualitative research}}
  \bibinfo{volume}{17}, \bibinfo{number}{1} (\bibinfo{year}{1994}),
  \bibinfo{pages}{273--285}.
\newblock


\bibitem[\protect\citeauthoryear{Villani, Pini, Leali, Secchi, and
  Fantuzzi}{Villani et~al\mbox{.}}{2018}]%
        {villani2018survey}
\bibfield{author}{\bibinfo{person}{Valeria Villani}, \bibinfo{person}{Fabio
  Pini}, \bibinfo{person}{Francesco Leali}, \bibinfo{person}{Cristian Secchi},
  {and} \bibinfo{person}{Cesare Fantuzzi}.} \bibinfo{year}{2018}\natexlab{}.
\newblock \showarticletitle{Survey on human-robot interaction for robot
  programming in industrial applications}.
\newblock \bibinfo{journal}{\emph{IFAC-PapersOnLine}} \bibinfo{volume}{51},
  \bibinfo{number}{11} (\bibinfo{year}{2018}), \bibinfo{pages}{66--71}.
\newblock


\bibitem[\protect\citeauthoryear{Wang, Ajaykumar, and Huang}{Wang
  et~al\mbox{.}}{2020}]%
        {wang2020see}
\bibfield{author}{\bibinfo{person}{Yeping Wang}, \bibinfo{person}{Gopika
  Ajaykumar}, {and} \bibinfo{person}{Chien-Ming Huang}.}
  \bibinfo{year}{2020}\natexlab{}.
\newblock \showarticletitle{See what i see: Enabling user-centric robotic
  assistance using first-person demonstrations}. In
  \bibinfo{booktitle}{\emph{Proceedings of the 2020 ACM/IEEE International
  Conference on Human-Robot Interaction}}. \bibinfo{pages}{639--648}.
\newblock


\bibitem[\protect\citeauthoryear{Weintrop, Afzal, Salac, Francis, Li, Shepherd,
  and Franklin}{Weintrop et~al\mbox{.}}{2018}]%
        {weintrop2018evaluating}
\bibfield{author}{\bibinfo{person}{David Weintrop}, \bibinfo{person}{Afsoon
  Afzal}, \bibinfo{person}{Jean Salac}, \bibinfo{person}{Patrick Francis},
  \bibinfo{person}{Boyang Li}, \bibinfo{person}{David~C Shepherd}, {and}
  \bibinfo{person}{Diana Franklin}.} \bibinfo{year}{2018}\natexlab{}.
\newblock \showarticletitle{Evaluating coblox: A comparative study of robotics
  programming environments for adult novices}. In
  \bibinfo{booktitle}{\emph{Proceedings of the 2018 CHI Conference on Human
  Factors in Computing Systems}}. \bibinfo{pages}{1--12}.
\newblock


\bibitem[\protect\citeauthoryear{Weintrop, Shepherd, Francis, and
  Franklin}{Weintrop et~al\mbox{.}}{2017}]%
        {weintrop2017blockly}
\bibfield{author}{\bibinfo{person}{David Weintrop}, \bibinfo{person}{David~C
  Shepherd}, \bibinfo{person}{Patrick Francis}, {and} \bibinfo{person}{Diana
  Franklin}.} \bibinfo{year}{2017}\natexlab{}.
\newblock \showarticletitle{Blockly goes to work: Block-based programming for
  industrial robots}. In \bibinfo{booktitle}{\emph{2017 IEEE Blocks and Beyond
  Workshop (B\&B)}}. IEEE, \bibinfo{pages}{29--36}.
\newblock


\bibitem[\protect\citeauthoryear{Wong and Seet}{Wong and Seet}{2017}]%
        {wong2017workload}
\bibfield{author}{\bibinfo{person}{Choon~Yue Wong} {and}
  \bibinfo{person}{Gerald Seet}.} \bibinfo{year}{2017}\natexlab{}.
\newblock \showarticletitle{Workload, awareness and automation in
  multiple-robot supervision}.
\newblock \bibinfo{journal}{\emph{International Journal of Advanced Robotic
  Systems}} \bibinfo{volume}{14}, \bibinfo{number}{3} (\bibinfo{year}{2017}),
  \bibinfo{pages}{1729881417710463}.
\newblock


\bibitem[\protect\citeauthoryear{Wulf and Jarke}{Wulf and Jarke}{2004}]%
        {wulf2004economics}
\bibfield{author}{\bibinfo{person}{Volker Wulf} {and} \bibinfo{person}{Matthias
  Jarke}.} \bibinfo{year}{2004}\natexlab{}.
\newblock \showarticletitle{The economics of end-user development}.
\newblock \bibinfo{journal}{\emph{Commun. ACM}} \bibinfo{volume}{47},
  \bibinfo{number}{9} (\bibinfo{year}{2004}), \bibinfo{pages}{41--42}.
\newblock


\bibitem[\protect\citeauthoryear{Young, Igarashi, Sharlin, Sakamoto, and
  Allen}{Young et~al\mbox{.}}{2014}]%
        {young2014design}
\bibfield{author}{\bibinfo{person}{James~E Young}, \bibinfo{person}{Takeo
  Igarashi}, \bibinfo{person}{Ehud Sharlin}, \bibinfo{person}{Daisuke
  Sakamoto}, {and} \bibinfo{person}{Jeffrey Allen}.}
  \bibinfo{year}{2014}\natexlab{}.
\newblock \showarticletitle{Design and evaluation techniques for authoring
  interactive and stylistic behaviors}.
\newblock \bibinfo{journal}{\emph{ACM Transactions on Interactive Intelligent
  Systems (TiiS)}} \bibinfo{volume}{3}, \bibinfo{number}{4}
  (\bibinfo{year}{2014}), \bibinfo{pages}{1--36}.
\newblock


\bibitem[\protect\citeauthoryear{Zhu and Hu}{Zhu and Hu}{2018}]%
        {zhu2018robot}
\bibfield{author}{\bibinfo{person}{Zuyuan Zhu} {and} \bibinfo{person}{Huosheng
  Hu}.} \bibinfo{year}{2018}\natexlab{}.
\newblock \showarticletitle{Robot learning from demonstration in robotic
  assembly: A survey}.
\newblock \bibinfo{journal}{\emph{Robotics}} \bibinfo{volume}{7},
  \bibinfo{number}{2} (\bibinfo{year}{2018}), \bibinfo{pages}{17}.
\newblock


\end{thebibliography}

\end{document}